\DocumentMetadata{
	lang        = en-US,
	pdfversion  = 1.7,
	pdfstandard = a-2b,
}

\documentclass[twoside]{mitthesis}

\usepackage{graphicx}
\usepackage{amsmath,amssymb,amsfonts}
\usepackage{microtype}
\usepackage{setspace}
\usepackage{hyperref}
\usepackage{xspace}
\usepackage{enumitem}
\usepackage{url}
\usepackage[normalem]{ulem}

\usepackage{booktabs}
\usepackage{array}
\usepackage{dcolumn}
\newcolumntype{d}[1]{D{.}{.}{#1}}
\usepackage{longtable}
\usepackage{multirow}
\usepackage{tabularx}
\usepackage{caption}
\DeclareCaptionLabelFormat{tablabel}{Tab.~#2}
\usepackage{sidecap}
\usepackage{wrapfig}
\usepackage[figuresleft]{rotating}

\usepackage{listings}
\usepackage{algorithm}
\usepackage{algpseudocode}
\lstdefinestyle{mystyle}{
    basicstyle=\tiny\ttfamily,
    breaklines=true,
    frame=single,
    numberstyle=\tiny,
    numbersep=5pt,
    tabsize=2,
    showstringspaces=false
}
\usepackage{tcolorbox}
\usepackage{epigraph}
\usepackage{pifont}
\usepackage{nicefrac}
\usepackage{dblfloatfix}

\usepackage[version=4]{mhchem}

\usepackage{tikz}
\usetikzlibrary{calc,decorations.pathreplacing}
\newcommand{\tikzmark}[1]{\tikz[overlay, remember picture] \node (#1) {};}

\usepackage{lipsum}
\IfPackageAtLeastTF{lipsum}{2021/09/20}{\setlipsum{auto-lang=false}}{}

\usepackage[style=ext-numeric-comp,giveninits=true,maxbibnames=10,sorting=none]{biblatex}
\hypersetup{
	colorlinks,
	allcolors=[rgb]{0,0.07843,0.45098},
	pdfsubject={Template for writing MIT theses with the mitthesis class},
	pdfkeywords={Massachusetts Institute of Technology, MIT},
	pdfurl={},
	pdfcontactemail={},
	pdfauthortitle={}
}

\DeclareMathOperator{\Erf}{erf}
\DeclareMathOperator{\Erfc}{erfc}
\DeclarePairedDelimiterXPP\erf[1]{\Erf\mkern1mu}(){}{#1}
\DeclarePairedDelimiterXPP\erfc[1]{\Erfc\mkern1mu}(){}{#1}

\usepackage{amsthm}
\newtheorem{condition}{Condition}
\newtheorem{theorem}{Theorem}
\newtheorem{lemma}{Lemma}

\newtheorem{assumption}{Assumption}

\definecolor{customred}{HTML}{A31F34}

\newcommand{\xhdr}[1]{\noindent\textbf{#1}~}
\newcommand{\cmark}{\ding{51}}
\newcommand{\xmark}{\ding{55}}

\newcommand{\GD}{\textit{GD}}
\newcommand{\BI}{\textit{BI}}
\newcommand{\RS}{\textit{RS}}
\newcommand{\OP}{\textit{OP}}
\newcommand{\SGS}{\textit{SS}}
\newcommand{\PR}{\textit{PR}}

\def\eqref#1{equation~\ref{#1}}

\def\1{\bm{1}}

\def\gL{{\mathcal{L}}}

\newcolumntype{L}[1]{>{\raggedright\let\newline\\\arraybackslash\hspace{0pt}}m{#1}}
\newcolumntype{C}[1]{>{\centering\let\newline\\\arraybackslash\hspace{0pt}}m{#1}}
\newcolumntype{R}[1]{>{\raggedleft\let\newline\\\arraybackslash\hspace{0pt}}m{#1}}

\newcommand{\sect}[1]{Section~\ref{#1}}

\newcommand{\fig}[1]{Fig.~\ref{#1}}
\newcommand{\tbl}[1]{Table~\ref{#1}}

\newcommand{\ignore}[1]{}

\makeatletter
\DeclareRobustCommand\onedot{\futurelet\@let@token\@onedot}
\def\@onedot{\ifx\@let@token.\else.\null\fi\xspace}

\def\eg{e.g\onedot}

\makeatother

\definecolor{MyDarkBlue}{rgb}{0,0.08,1}
\definecolor{MyDarkGreen}{rgb}{0.02,0.6,0.02}
\definecolor{MyDarkRed}{rgb}{0.8,0.02,0.02}
\definecolor{MyDarkOrange}{rgb}{0.40,0.2,0.02}
\definecolor{MyPurple}{RGB}{111,0,255}
\definecolor{MyRed}{rgb}{1.0,0.0,0.0}
\definecolor{MyGold}{rgb}{0.75,0.6,0.12}
\definecolor{MyDarkgray}{rgb}{0.66, 0.66, 0.66}
\definecolor{JiayuanColor}{rgb}{0.60,0.43,0.48}
\definecolor{FelixColor}{rgb}{0.086, 0.719 , 0.949}
\definecolor{MikeColor}{rgb}{0.7, 0.3 , 0.3}
\definecolor{JohnsonColor}{rgb}{0.66, 0.76, 0.66}
\definecolor{RebuttalColor}{rgb}{0.1,0.1,0.8}

\newcommand{\model}{GLiDE\xspace}
\newcommand{\modelfull}{\textbf{G}rounding \textbf{L}anguage \textbf{i}n \textbf{DE}monstrations\xspace}

\NewDocumentCommand{\prop}{m >{\SplitList{,}}m}{%
  \ensuremath{\textit{#1}(%
  \ProcessList{#2}{\propositionitem}%
  )}%
  \propositionfirstitemtrue
}

\newif\ifpropositionfirstitem
\propositionfirstitemtrue  
\newcommand\propositionitem[1]{%
  \ifpropositionfirstitem
    \propositionfirstitemfalse
  \else
    , %
  \fi
  \text{#1}}

\theoremstyle{plain}

\begin{document}

\title{Steering Robots with Inference-Time Interactions}

\Author{Yanwei Wang}{Department of Electrical Engineering and Computer Science}[B.A., Middlebury College, 2017][M.S., Northwestern University, 2018]
\Degree{Doctor of Philosophy}{Department of Electrical Engineering and Computer Science}
\Supervisor{Julie A. Shah}{H. N. Slater Professor of Aeronautics and Astronautics}
\Acceptor{Leslie A. Kolodziejski}{Professor of Electrical Engineering and Computer Science}{Chair, Department Committee for Graduate Students}
\DegreeDate{May}{2025}
\ThesisDate{April 25, 2025}

\maketitle

%
%


\NewDocumentCommand\CommitteePageTitle{m}{
	\vspace*{75pt}
	\IfPackageLoadedTF{microtype}
		{\textls*{\Large\textbf{\MakeUppercase{#1}}}}
		{{\Large\textbf{\MakeUppercase{#1}}}}%
	\pdfbookmark[0]{#1}{Committee}%
	\vspace*{10pt}%
}

\NewDocumentCommand\Role{m}{
	\vspace*{50pt}
	\IfPackageLoadedTF{microtype}
		{\textls*{\large{\textsc{#1}}}}
		{{\large\textsc{#1}}}%
	\vspace*{12pt}%
}


\begin{flushright}

\CommitteePageTitle{Thesis committee}

\Role{Thesis Supervisor}

 \textbf{Julie A. Shah} \\
 {\itshape
 H.N. Slater Professor of Aeronautics and Astronautics \\
 Massachusetts Institute of Technology \\
 }

\Role{Thesis Readers}

 \textbf{Leslie P. Kaelbling} \\
 {\itshape
   Panasonic Professor of Electrical Engineering and Computer Science \\
   Massachusetts Institute of Technology \\[18pt]
 }

 \textbf{Jacob Andreas}\\
 {\itshape
   Associate Professor of Electrical Engineering and Computer Science \\
   Massachusetts Institute of Technology \\[18pt]
 }

 \textbf{Dorsa Sadigh} \\
 {\itshape
   Associate Professor of the Computer Science Department \\
   Stanford University \\
 }

\end{flushright}

\cleardoublepage

\begin{abstract}
%
%

Imitation learning has driven the development of generalist policies capable of autonomously solving multiple tasks. However, when a pretrained policy makes errors during deployment, there are limited mechanisms for users to correct its behavior. While collecting additional data for finetuning can address such issues, doing so for each downstream use case is inefficient at deployment. My research proposes an alternative: keeping pretrained policies frozen as a fixed skill repertoire while allowing user interactions to guide behavior generation toward user preferences at inference time. By making pretrained policies steerable, users can help correct policy errors when the model struggles to generalize—without needing to finetune the policy. Specifically, I propose (1) inference-time steering, which leverages user interactions to switch between discrete skills, and (2) task and motion imitation, which enables user interactions to edit continuous motions while satisfying task constraints defined by discrete symbolic plans. These frameworks correct misaligned policy predictions without requiring additional training, maximizing the utility of pretrained models while achieving inference-time user objectives.

\end{abstract}

\tableofcontents

\chapter{Introduction}

\begin{flushright}
\textit{“You will never be happy if you continue to search for what happiness consists of. You will never live if you are looking for the meaning of life.” \\
— Albert Camus}
\end{flushright}

\begin{figure}[h!]
\centerline{\includegraphics[width=0.7\textwidth]{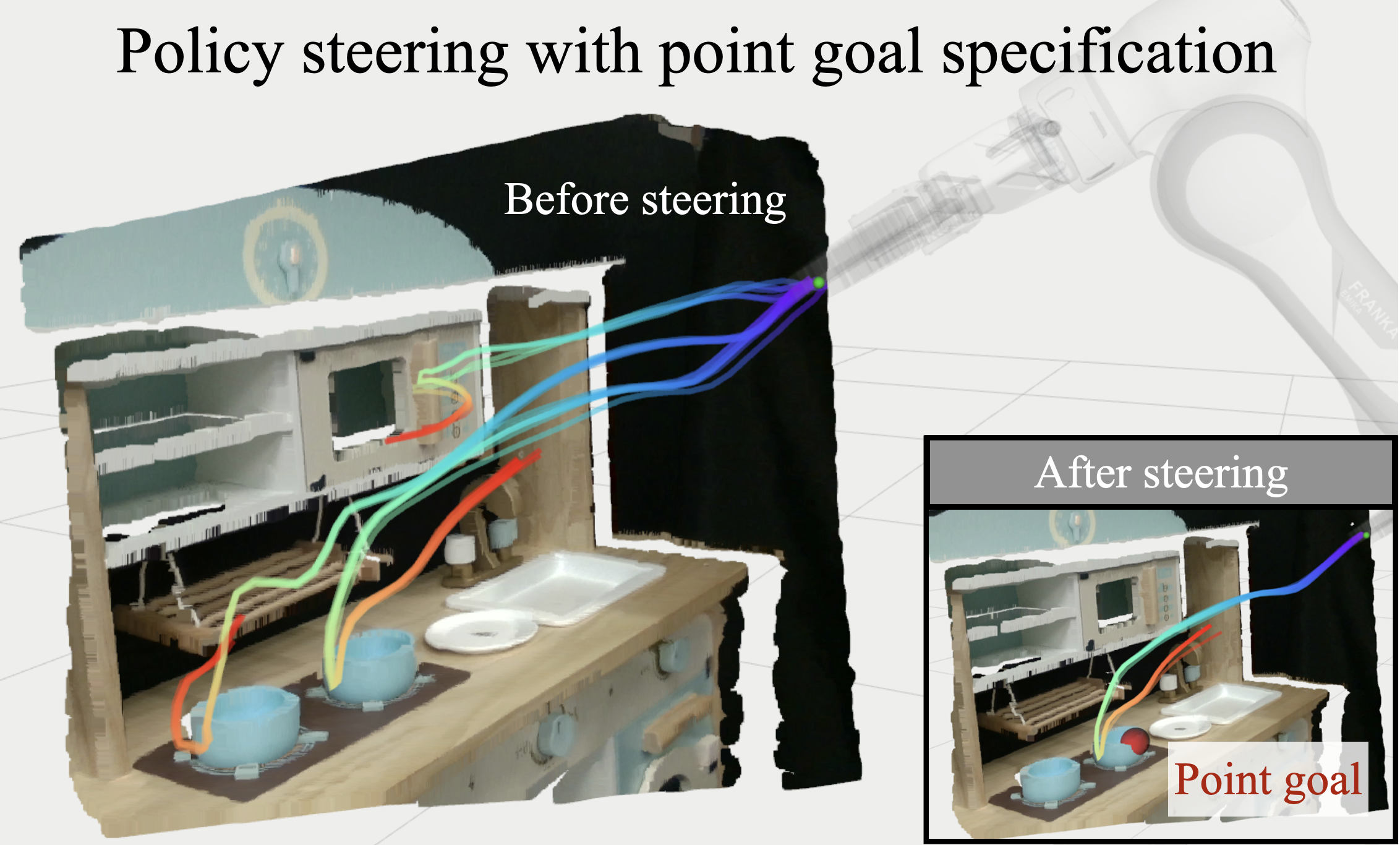}}
\caption{Steering robots with pointing inputs, where multimodal motion predictions collapse into the trajectory most aligned with user intent.}
\label{fig:intro_example}
\end{figure}

Imagine driving with Google Maps as your navigation assistant. You see multiple routes and can select one with a simple click. You can further customize your route by dragging your destination, and Google Maps instantly generates feasible routes to match your preferences. \textbf{My research aims to create a similar real-time interactive experience with pretrained imitation policies} \cite{argall2009survey}, allowing users to intuitively steer robots through physical interactions such as pointing or nudging, expecting immediate adjustments and feasible executions (Figure \ref{fig:intro_example}).

\begin{figure}[b]
\centerline{\includegraphics[width=1\textwidth]{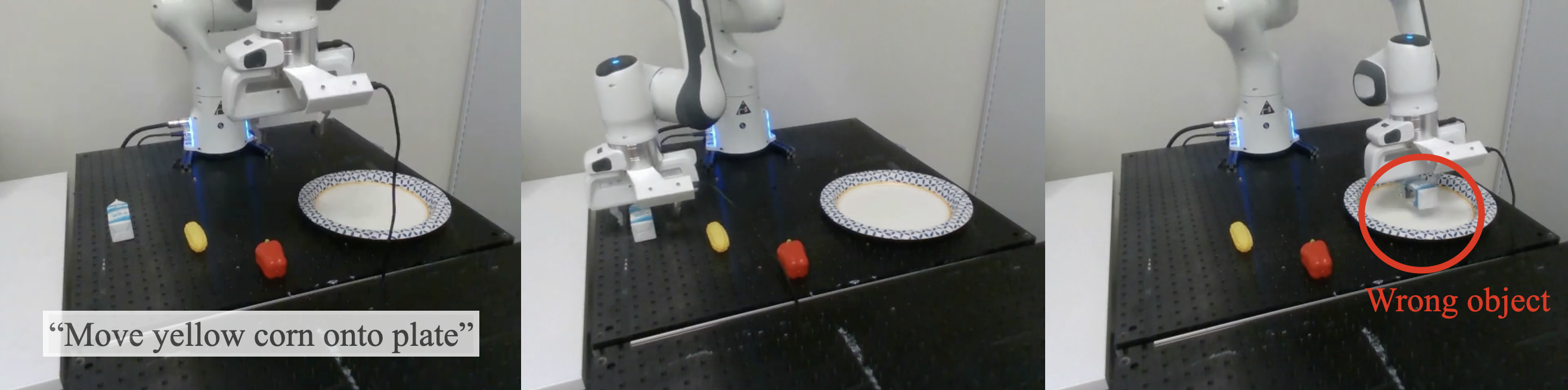}}
\caption{Vision language action model (VLA) failure case in a new environment \cite{kim2024openvla}.}
\label{fig:intro_failure}
\end{figure}

Recent successes in imitation learning systems \cite{chi2023diffusion, zhao2023learning} have spurred efforts to curate large datasets \cite{jang2022bc, o2023open} and develop vision-language-action (VLA) models for robot motion generation \cite{bommasani2021opportunities, brohan2023rt, kim2024openvla}. These pretrained VLAs are multitask policies capable of predicting multiple action sequences given visual observations. Typically, users guide these policies using high-level language instructions, after which actions are generated autonomously. While VLAs perform relatively well within their training domains, they struggle to generalize to new environments or language instructions during inference. For example, instructing the policy to pick up one object might inadvertently lead to it picking up another in an unfamiliar environment, as shown in Figure \ref{fig:intro_failure}. Furthermore, these policies typically cannot incorporate real-time user feedback to correct mistakes instantly during execution.

\begin{figure}[tp]
\centerline{\includegraphics[width=1\textwidth]{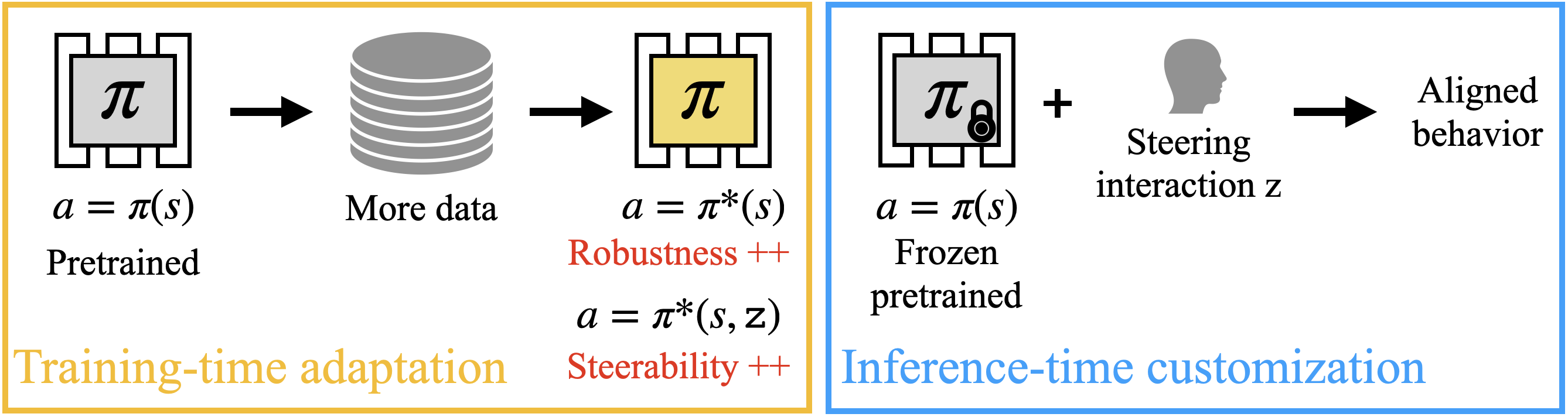}}
\caption{(Left) Prior work finetunes pretrained policies for robustness or steerability. (Right) Our work makes pretrained policies steerable without finetuning, leveraging online human interactions to correct mistakes and align behavior generation with user intent.}
\label{fig:intro_framework}
\end{figure}

When default language-based task specification mechanisms fail to produce desired behaviors \cite{karnik2024embodied}, existing approaches often collect additional data to finetune pretrained policies, as illustrated in Figure \ref{fig:intro_framework} (left). One line of work finetunes VLAs to improve robustness in novel environments or with novel language instructions \cite{kim2024openvla}. Others update model inputs to explicitly condition the policy on human corrective feedback during execution, making them steerable \cite{lynch2023interactive, shi2024yell}. We refer to these methods as \textit{training-time adaptation} approaches because they update policy weights using task-specific data collected downstream.

Although finetuning pretrained policies effectively corrects errors, end users often lack the necessary infrastructure to perform data collection and model training after deployment. Even when users have such infrastructure, finetuning during robot use in environments like homes, retail stores, or assistive settings \cite{dragan2013formalizing, gopinath2016human, woodworth2018preference} can disrupt workflow and human-robot collaboration. Thus, we investigate whether users can steer a non-interactive policy without modifying its weights, as illustrated in Figure \ref{fig:intro_framework} (right). We refer to this training-free adaptation as \textit{inference-time\footnote{We use "inference-time" \cite{li2023inference} synonymously with "test-time" \cite{peng2023diagnosis} or "deployment-time" \cite{hansen2020self}.} customization\footnote{The term "adaptation" implies training \cite{hu2021lora, lu2023inference}, so we use "customization" for training-free steering.}}.

To assess whether inference-time customization is viable, we first note that policy errors arise from two main sources: skill deficiency and task misspecification. While finetuning may be necessary for skill deficiencies, simpler tasks like pick-and-place (Figure \ref{fig:intro_failure}) often fail due to task misspecification rather than lacking necessary skills. In these cases, accessing the correct skill from a pretrained model via additional task specification can circumvent the need for finetuning. This thesis focuses on such scenarios, exploring inference-time methods that incorporate human interactions\footnote{We primarily focus on physical interactions like pointing, sketching, and physical corrections, as these directly ground user intent in task space, avoiding additional misspecifications introduced by mapping heterogeneous modalities like gaze \cite{saran2020efficiently} or haptic feedback \cite{cuan2024leveraging} to task space.}. Specifically, given a pretrained policy $a = \pi(s)$, where $s$ includes necessary policy inputs (e.g., robot state, image, language) and $a$ is a trajectory in $\mathrm{SE}(3)$, users provide steering interactions $z$ to clarify task specifications. Rather than modifying policy weights (i.e., $a=\pi^*(s)$) or updating input conditioning (i.e., $a=\pi^*(s,z)$), we seek to use $z$ to influence the trajectory generation process of the original policy $\pi$.

\begin{figure}[t] \centerline{\includegraphics[width=1\textwidth]{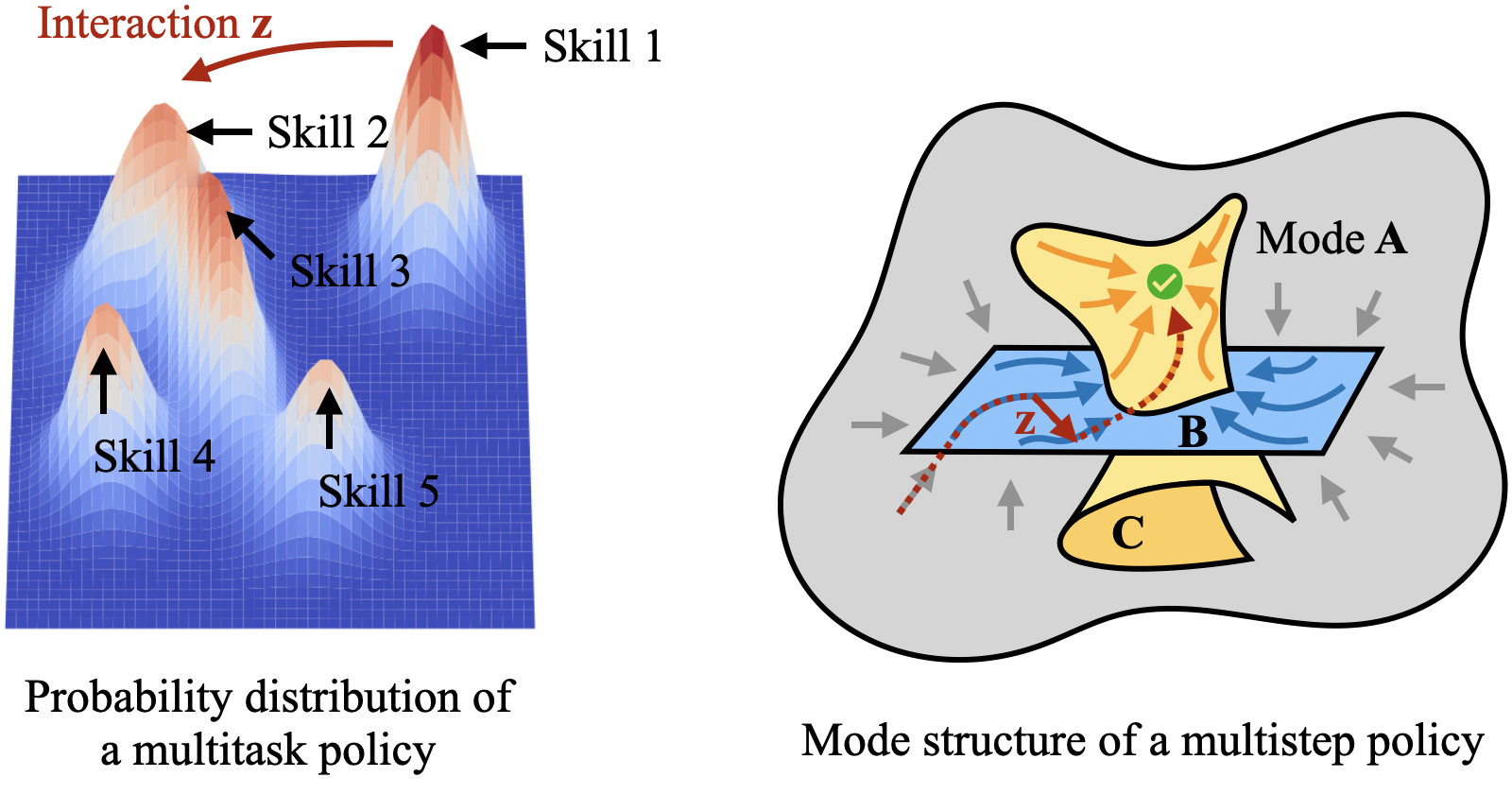}} \caption{(Left) \textbf{Inference-Time Policy Steering.} For a multitask policy, interaction $\mathbf{z}$ steers sampling from one probable region to another within a multimodal distribution. (Right) \textbf{Task and Motion Imitation.} For a multistep policy, interaction $\mathbf{z}$ steers the continuous motion trajectory, explicitly constraining it to remain within each corresponding mode in sequence (e.g., $A \rightarrow B \rightarrow C$).} \label{fig:intro_outline} \end{figure}

However, influencing a frozen policy presents significant challenges. Specifically, steering can exacerbate covariate shift \cite{ross2011reduction} due to compounding errors, particularly when introducing user interactions. These interactions may cause trajectories to drift from the training distribution, violate task constraints, and ultimately result in execution failures \cite{fu2024mobile}. To address this, we propose two frameworks to steer pretrained policies toward behaviors aligned with user intent, while minimizing deviations from the training distribution: (A) Inference-Time Policy Steering (ITPS, Figure \ref{fig:intro_outline}, left) and (B) Task and Motion Imitation (TAMI, Figure \ref{fig:intro_outline}, right). ITPS enables users to select discrete skills from multitask policies and implicitly mitigates distribution shift by probabilistically constraining behavior generation within the pretrained policy’s likelihood distribution. In contrast, TAMI allows steering of single-task, multistep policies toward preferred continuous motions. It explicitly ensures constraint satisfaction by first recovering a sequence of motion manifolds\footnote{We define a motion manifold as a smooth, low-dimensional subspace of the full configuration or trajectory space that encapsulates one coherent mode of feasible motion, often characterized by similar task objectives, constraints, or dynamic behaviors. More details can be found in the Task and Motion Imitation chapters, where we also call motion manifolds "modes."} (modes), and then bounding behavior generation sequentially within these modes.

\section{Inference-Time Policy Steering}

Consider a multitask policy $\pi(s)$ that predicts a multimodal\footnote{The term “multimodal” is used in various contexts: to describe a probability distribution with multiple modes \cite{hausman2017multi}, a task with inherent multimodal structure as in motion planning literature \cite{hauser2010multi}, or a system with multiple sensory modalities \cite{su2016learning}. In this thesis, we overload the term “multimodal” to refer both to distributions and task structures, depending on context.} distribution over action plans $a$, as illustrated in Figure \ref{fig:intro_outline} (left). Each mode corresponds to a different skill. Random sampling might produce a motion plan corresponding to skill $1$, while a user might intend skill $2$ instead. Given a steering interaction $z$ that unambiguously specifies skill $2$, how can we use $z$ to steer motion generation towards skill $2$? Prior methods have explored classifier guidance with diffusion models to steer image generation \cite{dhariwal2021diffusion}. However, classifier guidance introduces distribution shift, which, while less noticeable in the image domain, becomes significant in the trajectory domain, leading to execution failures.

To steer policies using interactions without exacerbating distribution shift, we propose an MCMC-inspired approach called \textit{stochastic sampling} for diffusion policies \cite{wang2024inference}. Stochastic sampling reduces the distribution shift caused by inference-time interactions by framing steering as conditional sampling from the likelihood distribution of a pretrained generative policy \cite{ajay2022conditional}. This likelihood distribution, learned during imitation, probabilistically constrains motion generation to the valid trajectory space, while conditional sampling ensures trajectories align closely with user preferences.

\section{Task and Motion Imitation}

Consider now a multistep policy designed for a single task, where task structure is represented as a sequence of intersecting modes, and task success requires traversing these modes sequentially, as illustrated in Figure \ref{fig:intro_outline} (right). Suppose a user wishes to edit the policy rollout with physical corrections. How can we ensure these interactions do not prematurely cause trajectories to exit a mode whose boundary encodes essential constraints (e.g., maintaining grasp during transport), thereby resulting in execution failures?

An ideal steerable policy should accept human interactions that reflect user preferences without compromising downstream tasks (e.g., allowing variability in inconsequential parts of reaching motions), yet also recover from interactions that might derail tasks (e.g., violating grasp constraints). Existing methods either treat user interactions as adversarial and impose strong priors for staying close to demonstrated trajectories \cite{ijspeert2013dynamical,laskey2017dart, haldar2022watch}, or allow compliance but focus solely on single-step reaching tasks \cite{khansari2011learning, bajcsy2017learning,figueroa2018physically}. Our key insight is that imitation of multistep tasks does not need to track demonstrations \textit{exactly}. Task execution succeeds as long as the continuous policy rollout conforms to a symbolic plan described by a mode sequence necessary for task completion. This view enhances steerability, as pretrained policies now only need to track a symbolic plan rather than a specific demonstration. For instance, in Figure \ref{fig:intro_outline} (right), a predicted trajectory $a = \pi(s)$ or its variant $a^*$ following human interaction will successfully execute a pick-and-place task if it can be mapped to the symbolic plan $A \rightarrow B \rightarrow C$, where modes $A$, $B$, and $C$ represent reaching, grasping, and transporting. Conversely, a trajectory mapped to $A \rightarrow C$ fails, as the pre-condition (mode $B$) of transporting is unmet before entering mode $C$.

Inspired by task and motion planning (TAMP) \cite{garrett2021integrated, kress2018synthesis}, we introduce Task and Motion Imitation (TAMI)—a framework that imitates (rather than plans) continuous motions satisfying discrete task constraints defined by user-specified symbolic plans. Unlike prior approaches that imitate TAMP solutions \cite{mcdonald2022guided, dalal2023imitating} but cannot guarantee success under user interactions, TAMI ensures both robustness and steerability within the learned constraint sets. Theoretically, TAMI allows deriving formal success guarantees for steering multistep policies. Practically, our method accommodates policies pretrained from \textit{a single demonstration} and a corresponding symbolic mode sequence, enabling flexible steering within mode boundaries and robust replanning if steering interactions cause deviations from the symbolic plan \cite{wang2022temporal}. Unlike the probabilistic constraints in ITPS, TAMI constrains motion generation with hard mode boundaries. Consequently, TAMI requires classifiers mapping continuous states into discrete modes to recover these boundaries, a topic elaborated in the next section.

\section{Outline}

In this thesis, we describe two frameworks that enhance inference-time steerability of pretrained policies. \textbf{Chapter 2} describes methods implementing the ITPS framework. \textbf{Chapters 3 and 4} address two key challenges within the TAMI framework: specifically, (\textbf{Chapter 3}) ensuring continuous motion generation satisfies the task constraints of a discrete symbolic plan given a mode classifier, and (\textbf{Chapter 4}) learning the mode classifier to map continuous motions to discrete modes.

\textbf{Chapter 2: Inference-Time Policy Steering with Diffusion.} In this chapter, we experiment with three types of interaction inputs grounded in robot task space, translating them into cost functions based on their L2 distance to behavior samples. We evaluate six sampling methods across two classes of generative models, identifying a fundamental alignment-constraint satisfaction tradeoff: steering that improves a frozen policy's alignment with user intent simultaneously increases constraint violations and task failures. Our proposed stochastic sampling procedure \cite{wang2024inference} for diffusion policies \cite{chi2023diffusion} achieves the best balance in this tradeoff, outperforming other sampling strategies.

\textbf{Chapter 3: Task and Motion Imitation with LTL Specification.} In this chapter, we introduce an instantiation of TAMI called Temporal Logic Imitation (TLI), where symbolic task plans are derived from user-specified linear temporal logic formulas. Given a classifier, TLI first identifies task constraints in the form of mode boundaries. When motion generation remains within constraints, TLI maximizes steerability using interaction-compliant policies \cite{figueroa2018physically}, allowing intuitive robot guidance by users. If generated motions risk violating task constraints, TLI applies inference-time corrections to edit trajectories that would otherwise leave mode boundaries. We demonstrate that, for human interactions to steer a pretrained multistep policy without causing constraint violations, the underlying imitation policy must satisfy both reachability and invariance properties \cite{wang2022temporal}.

\textbf{Chapter 4: Learning Grounding Classifiers for Task and Motion Imitation.} In this chapter, we introduce Grounding Language Plans in Demonstrations (GLiDE), a method for learning the grounding classifier required by the TAMI framework to connect continuous motions and discrete symbolic plans. Unlike prior work that relies on dense labeling to learn such classifiers \cite{migimatsu2022grounding}, we recognize that most pretrained policies lack dense constraint annotations. Moreover, manually labeling these constraints post hoc is tedious. To reduce human effort, our method first probes pretrained policies with physical perturbations, generating both successful and failed executions. By contrasting successful and failed outcomes, GLiDE learns a grounding classifier that partitions the configuration space into discrete modes whose boundaries encode task constraints. Across simulation and real-world experiments, we demonstrate that this approach accurately recovers sharp motion constraints with minimal labeling effort \cite{wang2024grounding}.

\textbf{Chapter 5: Conclusion.} In this chapter, we summarize the contributions and outline future research directions.

\chapter{Inference-Time Policy Steering with Diffusion}

\begin{tcolorbox}
\textsc{$\textbf{Inference-Time Policy Steering through Human Interactions}$ \\
\footnotesize Yanwei Wang, Lirui Wang, Yilun Du, Balakumar Sundaralingam, Xuning Yang, Yu-Wei Chao, Claudia Perez-D'Arpino, Dieter Fox, Julie Shah \\ \textbf{ICRA 2025}}
\end{tcolorbox}

\begin{flushright}
\vspace{1cm} 
\textit{"The only thing I can do is allow." \\
— Bert Hellinger}
\end{flushright}

\section{Introduction}

Behavior cloning \cite{osa2018algorithmic} has fueled a recent wave of generalist policies \cite{o2023open, team2024octo, kim2024openvla} capable of solving multiple tasks using a single deep generative model \cite{urain2024deep}. As these models acquire an increasing number of dexterous skills \cite{chi2023diffusion, zhao2023learning, fu2024mobile} from multimodal\footnote{In this work, multimodal refers to the data distribution, not interaction or sensor modalities.} human demonstrations, the natural next question arises: How can these skills be tailored to follow specific user objectives? Currently, there are few mechanisms to directly intervene and correct the behavior of these out-of-the-box policies at inference time, particularly when their actions misalign with user intent---often due to task under-specification or distribution shift during deployment.

\begin{figure}[t]
    \centering
    \includegraphics[width=0.8\linewidth]{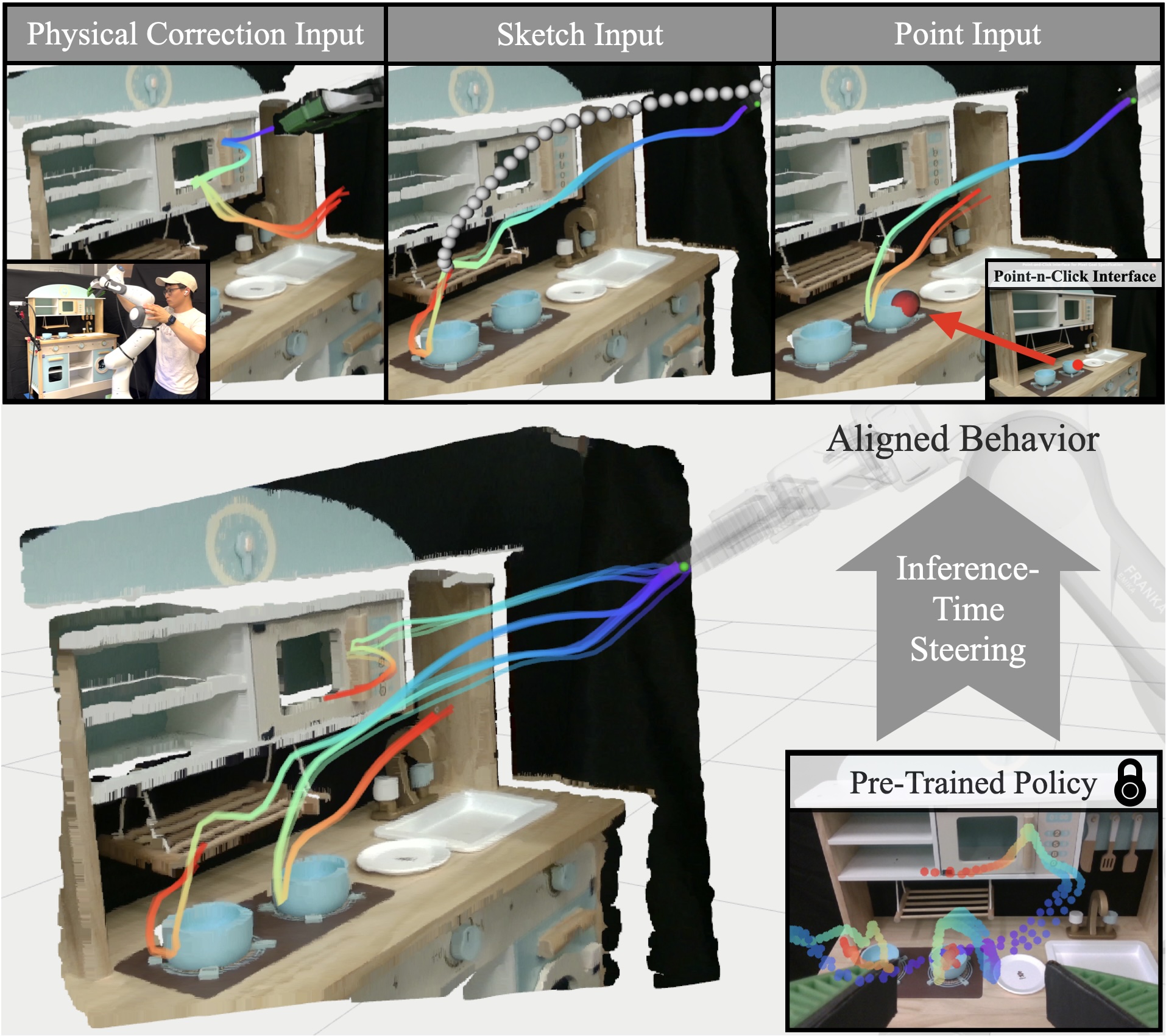}
    \captionsetup{width=0.8\linewidth}
    \captionof{figure}{\small \textbf{Inference-Time Policy Steering (ITPS).} We present a novel framework to unify various forms of human interactions to steer a frozen generative policy. User interactions ``prompt'' pre-trained policies to synthesize aligned behaviors at inference time. }
    \label{fig:itps_framework}
\end{figure}

One strategy for adapting policies designed for autonomous behavior generation to real-time human-robot interaction is to fine-tune them on interaction data, such as language corrections \cite{shi2024yell}. However, this approach requires additional data collection and training, and language may not always be the best modality for capturing low-level, continuous intent \cite{gu2023rt}. In this work, we explore whether a frozen pre-trained policy can be \textit{steered to generate behaviors aligned with user intent}---specified directly in the task space through point goals \cite{kemp2008point}, trajectory sketches \cite{gu2023rt}, and physical corrections \cite{wang2024grounding} (Figure \ref{fig:itps_framework})---without fine-tuning.

While inference-time interventions in the task space offer a direct way to guide behavior, they can inadvertently exacerbate distribution shift---a well-known issue in behavior cloning that often leads to execution failures \cite{ross2011reduction}.
Prior works addressing this issue \cite{losey2018review, losey2022physical, billard2022learning, wang2022temporal} largely focus on single-task settings, limiting their applicability to multi-task policies. To overcome this limitation, we leverage multimodal generative models to produce trajectories that respect likelihood constraints \cite{janner2022planning, ajay2022conditional, ye2024efficient}, ensuring the policy generates valid actions even after steering. Specifically, we frame policy steering as conditional sampling from the likelihood distribution of a learned generative policy. The likelihood constraints learned from successful demonstrations allow us to consistently synthesize valid trajectories, while conditional sampling ensures that these trajectories align with user objectives. By composing pre-trained policies with inference-time objectives, we can flexibly adapt generalist policies to each new downstream interaction modality, {\it without needing to modify the pre-trained policy in any way.}

To evaluate the effectiveness of inference-time steering, we formulate discrete and continuous alignment metrics to capture human preferences in discrete task execution and continuous motion shaping. We study a suite of six methods for converting interaction inputs into conditional sampling on generative models. We identify an alignment-constraint satisfaction trade-off: as these methods improve alignment, they tend to produce more constraint violations and task failures. To address this, we propose an MCMC procedure \cite{du2023reduce} for diffusion policy \cite{chi2023diffusion} that alleviates distribution shift during interaction-guided sampling, achieving the best alignment-constraint satisfaction trade-off across various combinations of generative policies and sampling strategies.

\textbf{Contributions} \textbf{(1)} We propose a novel inference-time framework (ITPS) that incorporates real-time user interactions to steer frozen imitation policies. \textbf{(2)} We introduce a set of alignment objectives, along with sampling methods for optimizing these objectives, and illustrate the alignment-constraint satisfaction trade-off. \textbf{(3)} We design a new inference algorithm for diffusion policy—stochastic sampling—which improves sample alignment with user intent while maintaining constraints within the data manifold.


\section{Method}

\subsection{Specification of User Intent}
In this work, we explore how to produce trajectories $\tau$ from frozen generative models that align with user intent specified either as discrete tasks (e.g. picking left or right bowl as shown in Figure 1) or continuous motions. For discrete preferences, we aim to maximize \textbf{T}ask \textbf{A}lignment (\texttt{TA}) as the percentage of predicted skills that execute intended tasks. For continuous preferences, we aim to maximize \textbf{M}otion \textbf{A}lignment (\texttt{MA}) as the negative $\mathcal{L}_2$ distance between generated trajectories and target trajectories. In addition to explicitly specified user objectives, we measure the percentage of generated plans that satisfy physical constraints---implicit user intents such as avoiding collisions or completing tasks---referred to as the \textbf{C}onstraint \textbf{S}atisfaction rate (\texttt{CS}). We define {\it steering towards user intent} as increasing \texttt{TA} or \texttt{MA} while maximizing \texttt{CS}. Specifically, maximizing \texttt{CS} is achieved through sampling in distribution of a pre-trained policy, while increasing \texttt{TA} or \texttt{MA} is achieved through minimizing an objective function $\xi(\tau, \mathbf{z})$, where user informs his intent through interactions $\mathbf{z}$ to score the space of trajectories $\tau$. 
We consider the following three interaction types and objective functions.

\textbf{Point Input.}  The first objective function $\xi$ has a user specify a point coordinate on an image we wish to have a robot trajectory reach. Given a generated trajectory $\tau = (\mathbf{s}_1, \mathbf{s}_2, \dots, \mathbf{s}_T) \in \mathbb{R}^3$, we map the specified pixel using the depth information in an RGB-D scene camera to a corresponding 3D state $\mathbf{z^{\text{point}}} \in \mathbb{R}^3$. The alignment to user intent is then defined as minimizing the objective function:
\begin{equation}
\xi(\tau, \mathbf{z}^{\text{point}}) = \sum_{t=1}^{T} \frac{1}{T} \|\mathbf{s}_t - \mathbf{z}^{\text{point}} \|_2, 
\end{equation}
which captures the average $\mathcal{L}_2$-distance between all states in the generated trajectory and the target 3D state $\mathbf{z}$\footnote{While $\min\limits_{s_1 \dots s_T} \|\mathbf{s}_t - \mathbf{z}\|_2$ is more accurate, gradients are not smooth.}. 
This objective function allows users to flexibly point goals in a scene, by specifying which objects to manipulate in a real-world kitchen environment (Figure \ref{fig:itps_framework}).

\textbf{Sketch Input.} The next objective function $\xi$ we consider allows a user to specify a more continuous intent, by generating a partial trajectory sketch $\mathbf{z^\text{sketch}} \in \mathbb{R}^{T \times 3}$ that we wish to have the robot follow. Given this sketch, we define $\xi$ as:
\begin{equation}
\xi(\tau, \mathbf{z}^\text{sketch}) = \sum_{t=1}^{T} \|\mathbf{s}_t - \mathbf{z}^\text{sketch}_t\|_2.
\end{equation}
When the sketch $\mathbf{z}$ has a different length than generated trajectories $\tau$, we uniformly resampled $\mathbf{z}^\text{sketch}$ to match the temporal dimension of generated samples\footnote{$\mathcal{L}_2$ used over DTW \cite{muller2007dynamic} for smooth gradients and linear time complexity.}.
In comparison to the point input, this objective function allows users to specify shape preferences of a trajectory through a directional path in a robot's workspace (Figure \ref{fig:itps_maze2d_qualitative}).

\textbf{Physical Correction Input.} Finally, we consider an objective $\xi$ which allows a user to specify intent through physical corrections $\mathbf{z}^{\text{nudge}}$ on the robot. Minimizing the objective
\begin{equation}
\xi(\tau, \mathbf{z}^{\text{nudge}}) = 
\begin{cases}
     0, & \mathbf{s}_t = \mathbf{z}^{\text{nudge}}_t \text{ for } t \le k \\
     \infty, & \text{otherwise}
\end{cases}
\end{equation}
corresponds to overwriting the beginning portion (e.g. first $k$ steps) of a trajectory $\tau$ with a user-specified $\mathbf{z^{\text{nudge}}}$:
\begin{equation}
\tau = [\mathbf{z}^{\text{nudge}}_1, \dots, \mathbf{z}^{\text{nudge}}_k,  {\textbf{s}}_{k+1}, \dots,  {\textbf{s}}_T].
\end{equation}
Compared to previous interaction types, physical corrections intervene directly in the robot's motion execution (Figure \ref{fig:itps_framework}).

\subsection{Policy Steering}

\begin{figure}
    \centering
    \includegraphics[width=\linewidth]{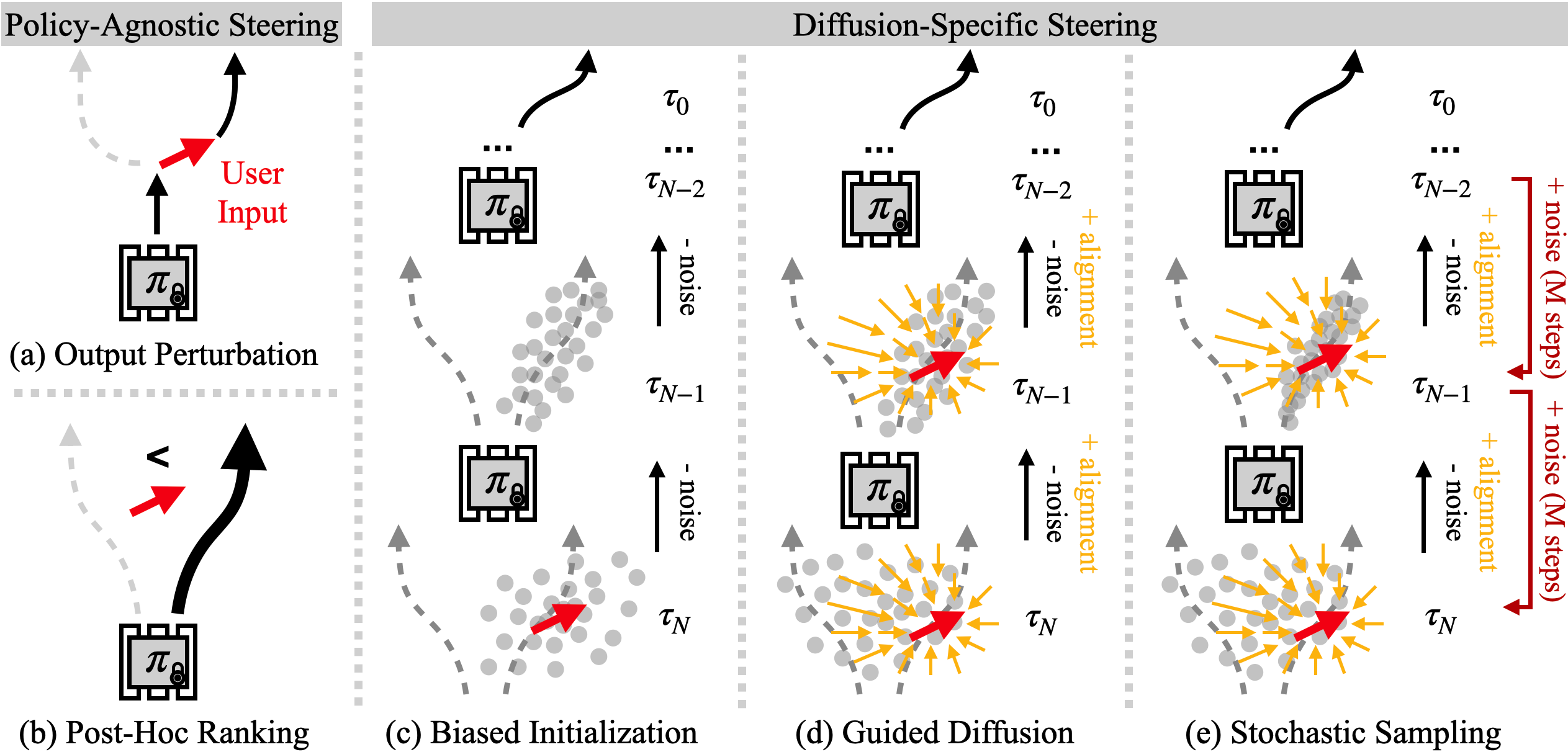} 
    \caption{\small \textbf{Policy Steering Methods.} Given user input, methods (a-c) incorporate the alignment objective either before or after inference via (a) perturbation, (b) ranking, or (c) initialization, whereas methods (d,e) integrate the objective directly during inference.}
    \label{fig:itps_method}
\end{figure}

Given an inference-time alignment objective $\xi(\tau, \mathbf{z})$ on trajectories $\tau$, we explore six methods for biasing trajectory generation to minimize this objective. The first three methods are applicable across generative models parameterized by $\theta$, while the latter three specifically leverage the implicit optimization procedure within the diffusion process. Figure~\ref{fig:itps_method} illustrates these optimization procedures.

\textbf{Random Sampling (RS).} In the \textit{Random Sampling} baseline, we sample a trajectory $\tau \sim \pi_\theta$ directly from the pre-trained model without any modification. This approach does not explicitly optimize any objective function $\xi$, but serves as a baseline for trajectory generation.

\textbf{Output Perturbation (\OP).} In \textit{Output Perturbation}, we first sample a trajectory $\tau$ from $\pi_\theta$ and apply a post-hoc perturbation to minimize the objective $\xi(\tau, \mathbf{z}^{\text{nudge}})$. We then resample from $\mathbf{z}_k^{\text{nudge}}$ to complete the remainder of trajectory $\tau$. If a user cannot provide direct physical correction, the first $k$ states of a sketch input can be used as $\mathbf{z}^{\text{nudge}}$. Although this sampling strategy maximizes alignment up to step $k$, it does not guarantee that synthesized trajectories from the perturbed state $\mathbf{z}_k^{\text{nudge}}$ will be constraint satisficing.

\textbf{Post-Hoc Ranking (PR).} In \textit{Post-Hoc Ranking}, we generate a batch of $N$ trajectories $\{\tau_j\}_{j=1}^N$ from $\pi_\theta$ and select $\tau^*$ that minimizes the objective $\xi(\tau, \mathbf{z}^{\text{point}})$ or $\xi(\tau, \mathbf{z}^{\text{sketch}})$.
This approach performs well when at least one generated sample closely aligns with the input $\mathbf{z}$, which may not hold if the robot is in a state without multimodal policy predictions.

\textbf{Biased Initialization (BI).}
In \textit{Biased Initialization}, inspired by \cite{yoneda2023noise}, we  modify the initialization of the reverse diffusion process. Instead of initializing with a noise trajectory $\tau_N$\footnote{Subscript denotes diffusion steps for $\tau_i$ and trajectory timesteps for $s_t$.} $\sim \mathcal{N}(0, I)$, we use a Gaussian-corrupted version of the user input $\mathbf{z^\text{point}}$ or $\mathbf{z^\text{sketch}}$ as $\tau_N$, bringing the process closer to the desired mode from the outset.
While this approach specifies user intent at initialization, the sampling process may still deviate from this input.

 \begin{figure}[t]
    \centering
    \begin{tabular}{p{0.55\linewidth}} 
        \rule{\linewidth}{1pt}  
        \noindent\textbf{\hspace{1em} Algorithm 1: Stochastic Sampling} \\[-0.3em]
        \rule{\linewidth}{0.5pt}  
        \noindent\textbf{\hspace{1em} Input:} diffusion $\pi_\theta$, interaction $\mathbf{z}$, alignment $\xi(\cdot)$ \\
        1: Initialize plan $\tau_N \sim \mathcal{N}(0, I)$ \\
        2: \textbf{for} $i = N, \dots, 1:$ \textcolor{gray}{\hspace{5em} // denoising steps} \\
        3: \hspace{1em} \textcolor{customred}{\textbf{for} $j = 1, \dots, M:$} \textcolor{customred}{\hspace{3.5em}// sampling steps} \\
        4: \hspace{2em} $\epsilon \gets \pi_\theta(\tau_i)$ \textcolor{gray}{\hspace{4em} // denoising gradient} \\
        5: \hspace{2.5em} $\delta \gets \nabla \xi(\tau_i, \mathbf{z})$ \textcolor{gray}{\hspace{2em} // alignment gradient} \\
        6: \hspace{2em} $\textcolor{customred}{\textbf{if } j < M}$: \\
        7: \hspace{3em} $\textcolor{customred}{\tau_{i} \gets \texttt{reverse}(\tau_i, \epsilon + \beta_i \delta, i)}$ \\
        8: \hspace{2em} \textcolor{customred}{\textbf{else}}: \\
        9: \hspace{3em} $\tau_{i-1} \gets \texttt{reverse}(\tau_i, \epsilon + \beta_i \delta, i-1)$ \\
        \rule{\linewidth}{0.5pt}  
    \end{tabular}
    \captionsetup{width=0.55\linewidth}
    \captionof{algorithm}{\small\textbf{Stochastic Sampling.} \textcolor{customred}{A four-line change} from a guided diffusion algorithm.}
    \label{alg:itps_psuedocode}
\end{figure}

\begin{figure}
    \centering
    \includegraphics[width=\linewidth]{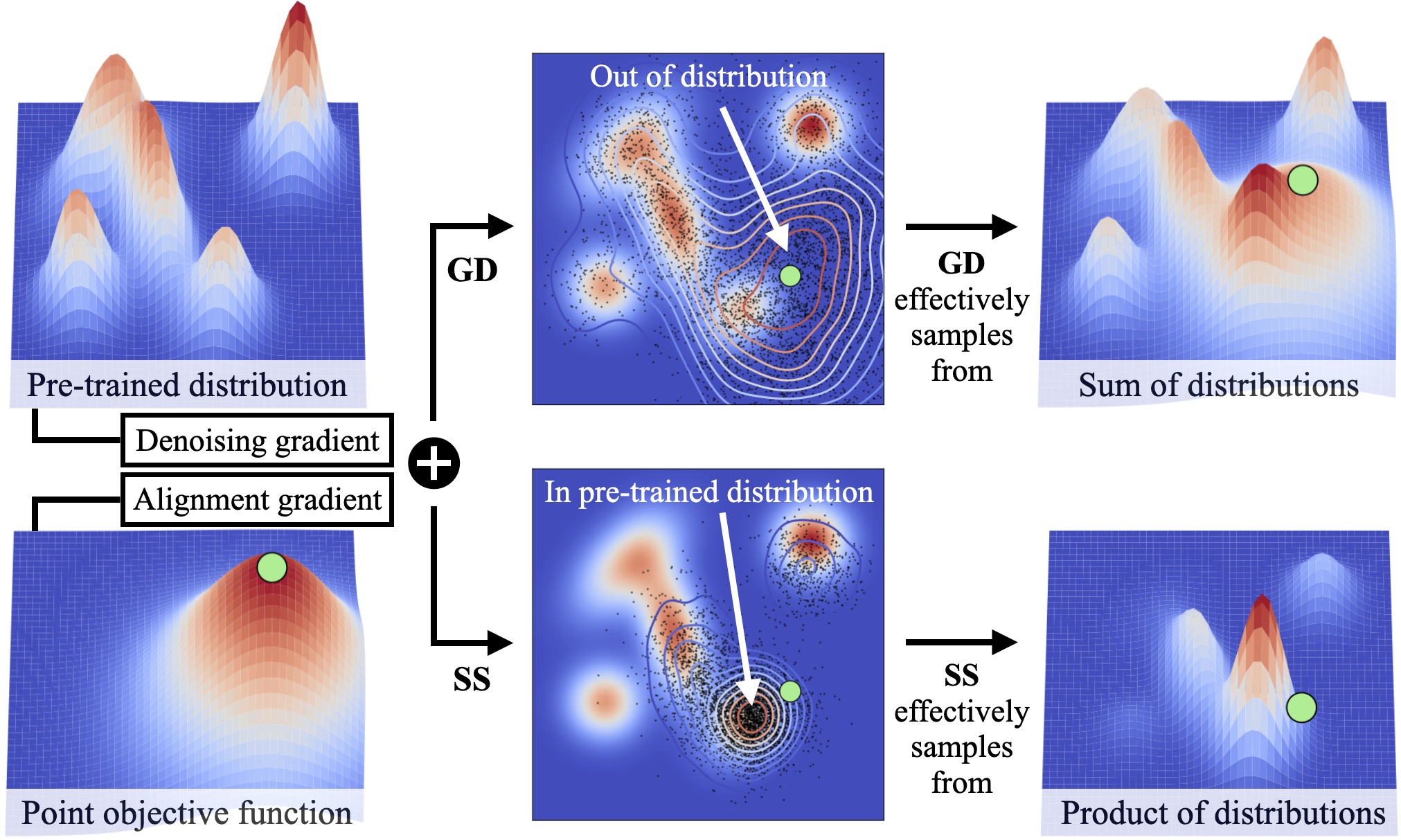}
    \captionof{figure}{\small \textbf{Guided Diffusion vs. Stochastic Sampling.} In a toy example aiming to sample likely data points from a pre-trained distribution while aligning with a target point, \GD{} samples approximate the sum of two distributions, whereas \SGS{} samples approximate their product, as illustrated by contour lines from kernel density estimation \cite{Waskom2021}. Consequently, when the point input does not align with any distribution mode, \GD{} introduces distribution shift, while \SGS{} identifies the closest in-distribution mode.}
    \label{fig:itps_gd_vs_ss}
\end{figure}

\textbf{Guided Diffusion (\GD).}
In \textit{Guided Diffusion}, we use the objective function $\xi(\tau, \mathbf{z})$ to guide the trajectory synthesis in the diffusion process~\cite{janner2022planning}. Specifically, at each diffusion timestep $i$, given $\mathbf{z^\text{point}}$ or $\mathbf{z^\text{sketch}}$, we compute the alignment gradient $\nabla_{\tau_i} \xi(\tau_i, \mathbf{z})$ to bias sampling:
\begin{equation}
\tau_{i-1} = \alpha_i(\tau_i - \gamma_i (\epsilon_\theta(\tau_i, i) + \beta_i \nabla_{\tau_i} \xi(\tau_i, \mathbf{z}))) + \sigma_i \eta,
\label{eqn:itps_guided_diffusion}
\end{equation}
 where $\epsilon_\theta(\tau_i, i)$ is the denoising network, $\eta \sim \mathcal{N}(0, I)$ is Gaussian noise, $\beta_i$ is the guide ratio that controls the alignment gradient's influence, $\alpha_i$, $\gamma_i$, $\sigma_i$ are diffusion-specific hyperparameters. This alignment gradient steers the reverse process toward trajectories aligned with $\mathbf{z}$, potentially discovering new behavior modes in states where unconditional predictions would otherwise be unimodal and far from  $\mathbf{z}$. However, sampling with a weighted sum of denoising and alignment gradients in Equation~\ref{eqn:itps_guided_diffusion} approximates sampling from the weighted sum of the policy distribution and the objective distribution rather than their product~\cite{du2023reduce}, which can result in out-of-distribution samples (Figure \ref{fig:itps_gd_vs_ss}).

\begin{figure}[t]
    \centering
        \includegraphics[width=0.7\linewidth]{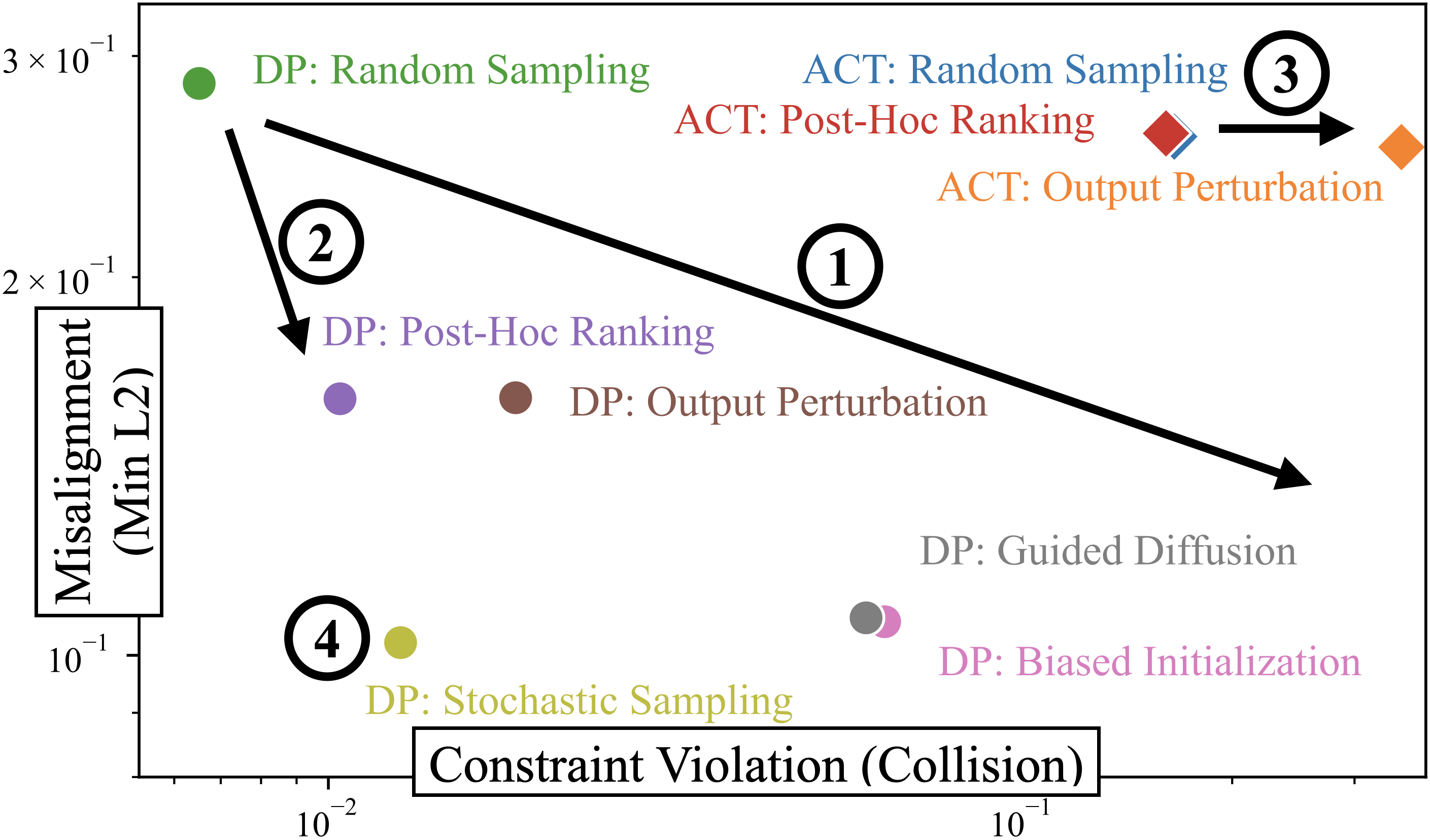}
        \captionsetup{width=0.7\linewidth}
        \caption{\small \textbf{Alignment vs. Collision in Maze2D.} 
        We compare various sampling methods with ACT and DP steered using sketch input. (1) Steering improves alignment at the cost of constraint satisfaction and increased collisions. Moreover, (2) Multimodal policies (DP) steered with \PR{} enhance alignment without significant distribution shift, while (3) unimodal policies (ACT) are harder to steer effectively, particularly if they lack robustness (see Figure \ref{fig:itps_maze_csr}). (4) Finally, DP steered with \SGS{} achieves the best alignment-constraint satisfaction trade-off.
    \label{fig:itps_maze_multi_trend}}
\end{figure}

\begin{figure}[t]
    \centering
    \includegraphics[width=1\linewidth]{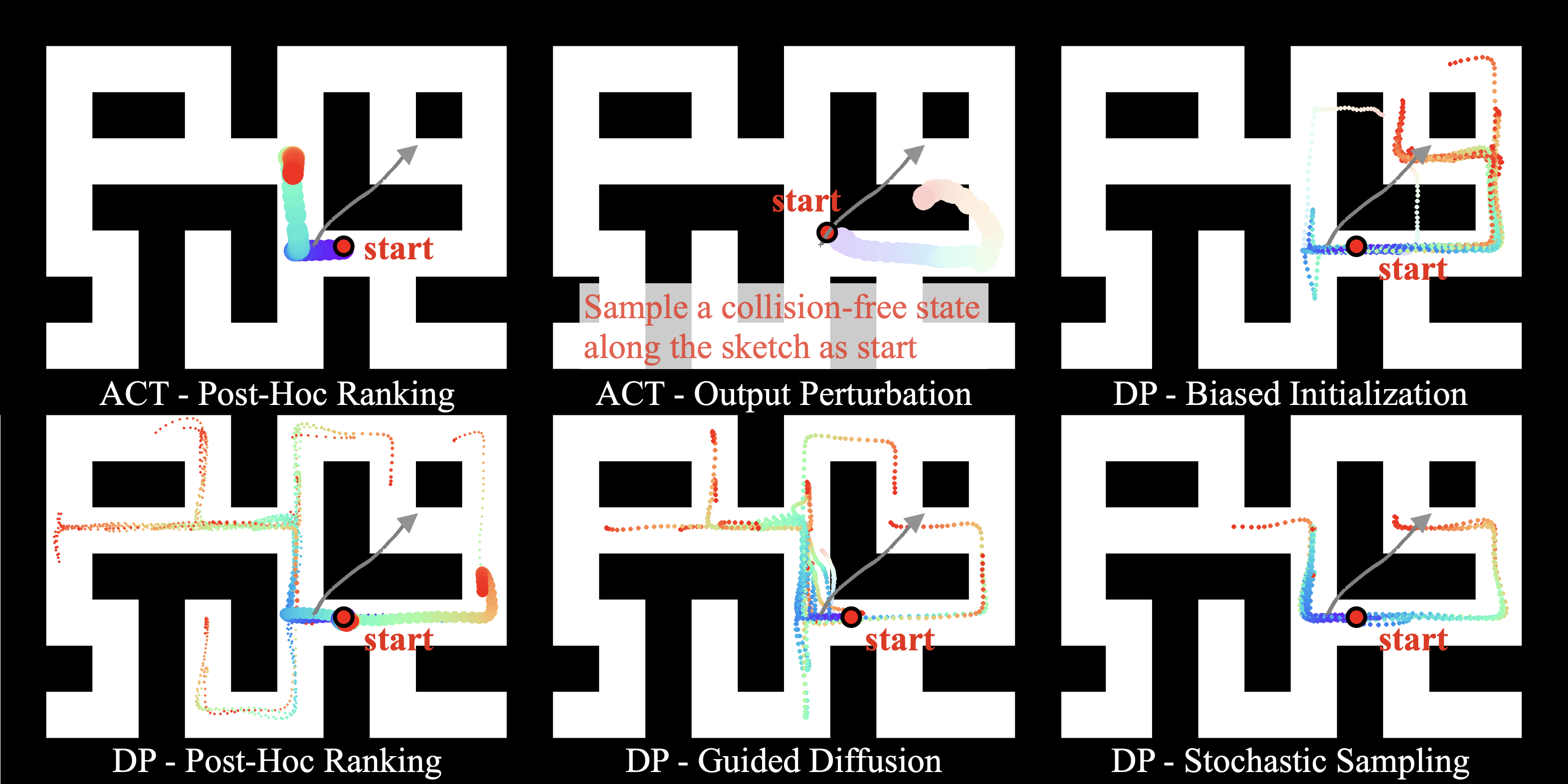}
    \caption{\small \textbf{Maze2D Qualitative Comparisons.} We visualize trajectories (color-coded from blue to red over time) sampled with various steering methods from two policy classes (ACT and DP) given a sketch in gray. Trajectory thickness reflects similarity to the sketch after ranking, and samples in collision are tinted white. \SGS{} preserves collision-free constraints while aligning with user intent.}
    \label{fig:itps_maze2d_qualitative}
\end{figure}

\begin{figure}[ht]
    \centering
    \includegraphics[width=0.70\linewidth]{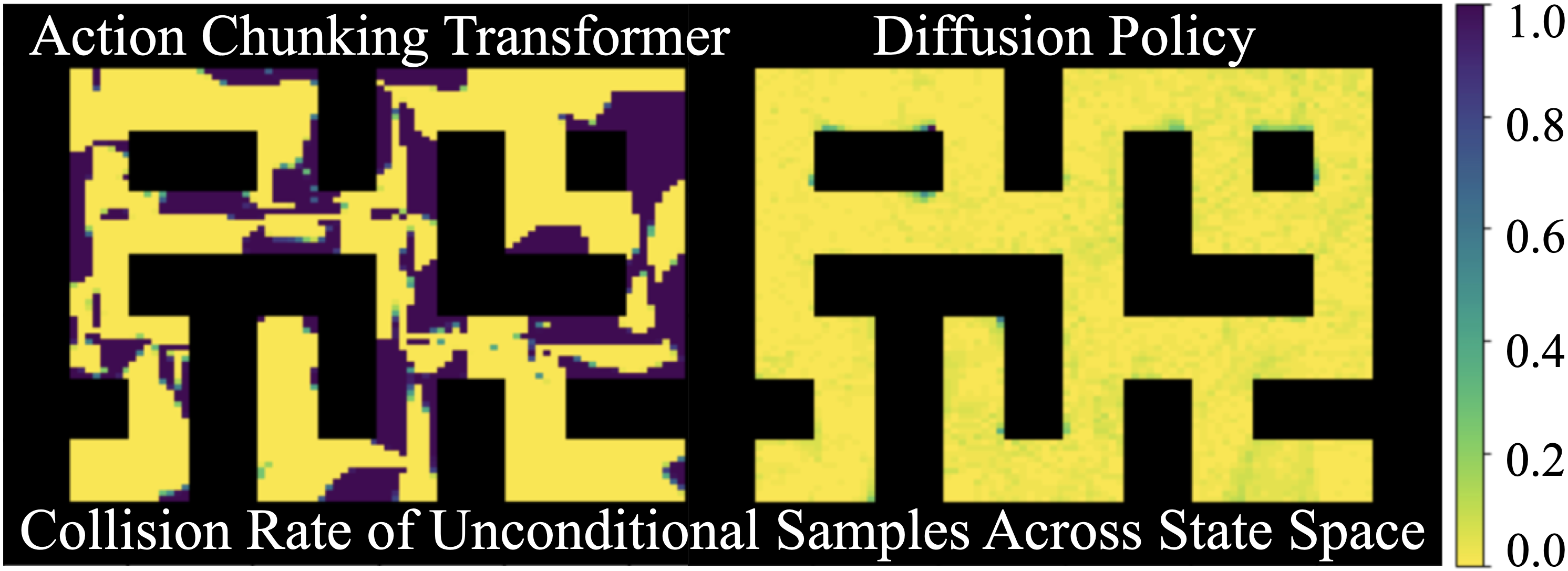}
    \captionsetup{width=0.9\linewidth}
    \caption{\small \textbf{Robustness of ACT/DP in Maze2D.}\\}
    \label{fig:itps_maze_csr}
\end{figure}

\begin{table}[]
    \centering
    \begin{tabular}{llcccccc}
    \toprule
    & & \textbf{\RS} & \textbf{\PR} & \textbf{\OP} & \textbf{\BI} & \textbf{\GD} & \textbf{\SGS} \\
    \midrule
    \multirow{3}{*}{\textbf{ACT}} 
    & Min $\mathcal{L}_2$ $\downarrow$ & 0.26 & 0.26 & 0.26 & - & - & - \\
    & Avg $\mathcal{L}_2$ $\downarrow$ & 0.26 & 0.26 & 0.26 & - & - & - \\
    & Collision $\downarrow$ & 0.16 & 0.16 & 0.35 & - & - & - \\
    \midrule
    \multirow{3}{*}{\textbf{DP}} 
    & Min $\mathcal{L}_2$ $\downarrow$ & 0.27 & 0.16 & 0.16 & 0.11 & 0.11 & \textbf{0.10} \\
    & Avg $\mathcal{L}_2$ $\downarrow$ & 0.28 & 0.28 & 0.28 & 0.14 & 0.18 & \textbf{0.12} \\
    & Collision $\downarrow$ & 0.01 & 0.01 & 0.02 & 0.06 & 0.06 & \textbf{0.01} \\
    \bottomrule
    \end{tabular}
    \captionsetup{width=0.7\linewidth}
    \captionof{table}{\small \textbf{Maze2D Results.} Mean collision rate and $\mathcal{L}_2$ distance between $\mathbf{z^\text{sketch}}$ and the closest sample (min) / all samples (ave) per batch across trials. \SGS{} achieves the best alignment with minimal collisions.}
    \label{tab:itps_maze}
\end{table}

\textbf{Stochastic Sampling (\SGS).}
Finally, in \textit{Stochastic Sampling}, we use annealed MCMC to optimize the composition of the diffusion model $\pi_\theta$ and the objective $\xi(\tau_i, \mathbf{z})$~\cite{du2023reduce}. Here, the denoising function $\epsilon_\theta(\tau_i, i)$ at each timestep $i$ represents the score $\nabla_\tau \log p_i(\tau)$ for a sequence of probability distributions $\{p_i(\tau)\}_{0\le i\le N}$, where $p_N(\tau)$ is Gaussian and $p_0(\tau)$ is the distribution of valid trajectories in the environment. Simultaneously, the objective $\xi(\tau, \mathbf{z})$ defines an energy-based model (EBM) distribution $q(\tau) \propto e^{-\xi(\tau, \mathbf{z})}$. Steering toward user intent then corresponds to sequentially sampling from $p_N(\tau) q(\tau)$ to $p_0(\tau) q(\tau)$, yielding final samples from $p_0(\tau) q(\tau)$ that are both valid within the environment and minimize the specified objective.

We implement this sequential sampling using the annealed ULA MCMC sampler, which can be implemented in a similar form to the guided diffusion code~\cite{du2023reduce}. First, we initialize a noisy trajectory $\tau_N \sim \mathcal{N}(0, I)$, corresponding to a sample from $p_N(\tau) q(\tau)$. We then run $M$ steps of MCMC sampling at timestep $i$ using the update equation:
\begin{equation}
\tau_{i} = \tau_i - \gamma_i (\epsilon_\theta(\tau_i, i) + \beta_i \nabla_{\tau_i} \xi(\tau_i, \mathbf{z})) + \sigma_i \eta,
\label{eqn:itps_mcmc_no_contract}
\end{equation}
repeated $M-1$ times, followed by a final reverse step in Equation \ref{eqn:itps_guided_diffusion}
to obtain a sample $\tau_{i-1}$ from $p_{i-1}(\tau) q(\tau)$. These steps closely resemble reverse sampling in Equation~\ref{eqn:itps_guided_diffusion} and can be implemented by modifying four lines in the guided diffusion code (Algorithm~\ref{alg:itps_psuedocode}). To implement the sampling of Equation~\ref{eqn:itps_mcmc_no_contract}, we take the intermediate clean trajectory prediction $\tilde{\tau}_0$ obtained via reverse sampling on $\tau_i$, followed by a forward diffusion step with noise level $i$ to update $\tau_{i}$. 
The addition of multiple reverse sampling steps at a fixed noise level better approximates sampling from a product distribution, as shown in Figure~\ref{fig:itps_gd_vs_ss}, producing samples that satisfy likelihood constraints and user objectives. 
Across our experiments, \SGS{} provides the most proficient policy steering.


\section{Experiments}

We evaluate the effectiveness of inference-time steering methods in improving continuous \textbf{M}otion \textbf{A}lignment (\texttt{MA}) in Maze2D and discrete \textbf{T}ask \textbf{A}lignment (\texttt{TA}) in the Block Stacking and Real World Kitchen Rearrangement tasks. Additionally, we report how steering affect \textbf{C}onstraint \textbf{S}atisfaction (\texttt{CS}) among samples.

\subsection{Maze2D - Continuous Motion Alignment (\texttt{MA})} 

For continuous motion alignment, we use Maze2D \cite{fu2020d4rl} to evaluate whether a generative policy trained exclusively on collision-free navigation demonstrations can remain on a collision-free motion manifold when steered with sketches that violate constraints. To test the impact of the pre-trained policy class, we train a VAE-based action chunking transformer (ACT) \cite{zhao2023learning} and a diffusion policy (DP) \cite{chi2023diffusion} on 4 million navigation steps between random locations in a maze environment. DP is trained with a DDIM \cite{song2020denoising} scheduler over 100 training steps $(N=100)$. The training objective focuses solely on modeling the data distribution (i.e., collision-free random walk) without any goal-oriented objectives.

At inference time, a given policy is kept frozen to benchmark various steering methods. We generate 100 random locations in the maze, each paired with a user sketch $\mathbf{z}^{\text{sketch}}$ that may not be collision-free. These sketch inputs steer the generation of a batch of 32 trajectories per trial from the policy. For DP, the scheduler is allocated 10 inference steps, with a guide ratio of $\beta_{i \leq N} = 20$ for \GD\ and $\beta_{i \leq N} = 60$ for \SGS\, where the MCMC sampling steps are set to $M=4$. To incorporate $\mathbf{z}^{\text{sketch}}$ in the \OP\ sampling procedure, an early portion of the sketch is sampled to identify a non-collision state, resetting the starting location accordingly. To evaluate steering, we report the collision rate ($1 - \texttt{CS}$) and the $\mathcal{L}_2$ distance between the sketch and the closest trajectory (Min $\mathcal{L}_2$) or all trajectories (Avg $\mathcal{L}_2$) per batch, which measures negative \texttt{MA}. Min $\mathcal{L}_2$ shows the best alignment, while Avg $\mathcal{L}_2$ captures the overall distribution alignment after steering. 

\begin{figure}[t]
    \centering
    \includegraphics[width=\linewidth]{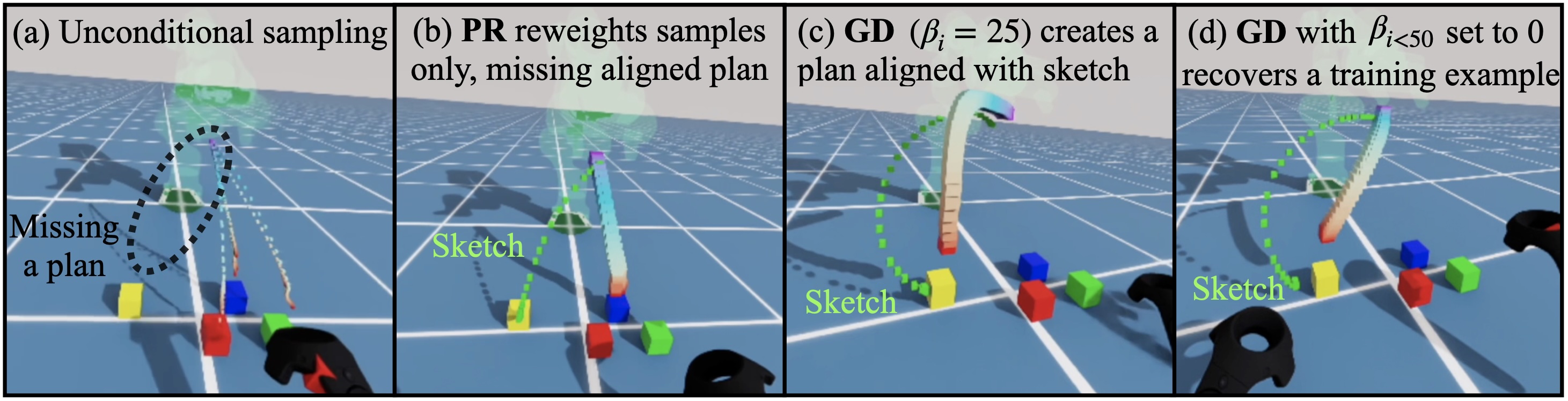} 
      \caption{\small \textbf{Block Stacking Qualitative Comparisons.} (a) Unconditional sampling from a DP may miss intended plans, which (b) \PR{} cannot recover, but (c) \GD{} can. (d) Adjusting the number of diffusion steps with steering (set guide ratio $\beta_i = 0$) balances similarity to the sketch versus adherence to the training distribution.}
      \label{fig:itps_block}
\end{figure}

\begin{figure}[t]
    \centering
    \setlength{\tabcolsep}{0pt} 
    \begin{tabularx}{\linewidth}{@{}l*{3}{>{\centering\arraybackslash}X}@{}}
    \toprule
    \textbf{Method Type (DP)} & \textbf{\PR} & \textbf{\GD} ($\beta_{i<50}$=$0$) & \textbf{\GD} ($\beta_{i}$=$100$) \\
    \midrule
    \tikzmark{TA}TA (Alignment: AS+AF)       & 33\% & 83\% & 86\% \\
    \tikzmark{CS}CS (Success: AS+MS)         & 100\% & 84\% & 15\% \\
    \tikzmark{AS}\textbf{Aligned Success (AS)} & 33\% & \textbf{67\%} & 15\% \\
    \tikzmark{AF}Aligned Failure (AF)          & 0\% & 16\% & 71\% \\
    \tikzmark{MS}Misaligned Success (MS)       & 67\% & 17\% & 0\% \\
    \tikzmark{MF}Misaligned Failure (MF)       & 0\% & 0\% & 14\% \\
    \bottomrule
    \end{tabularx}
    
    \captionof{table}{\small \textbf{Block Stacking Results.} \texttt{TA} is the percentage of interactions that achieve aligned execution, regardless of outcome. \texttt{CS} is the percentage of picking/placing success, regardless of alignment.}
    \label{tab:itps_block}
\end{figure}

Our findings, illustrated in Figure \ref{fig:itps_maze_multi_trend}, reveal a tradeoff between alignment and constraint satisfaction. Specifically, aggressive steering improves \texttt{MA} but reduces \texttt{CS} and increases collisions. Additionally, a policy with multimodal predictions (DP) combined with \PR\ effectively improves alignment without exacerbating distribution shift. However, if the intended plan is absent from the initial sampled batch, \PR\ cannot discover it (Figure \ref{fig:itps_maze2d_qualitative}). In contrast, a policy with unimodal predictions (ACT) cannot be steered to improve alignment with \PR. If the policy lacks robustness (Figure \ref{fig:itps_maze_csr}), \OP\ can introduce significant distribution shift. Finally, diffusion-specific steering methods can transform constraint-violating sketches into the nearest collision-free samples on the data manifold. Among these, \SGS\ achieves the best \texttt{MA} and \texttt{CS} tradeoff, as shown in Table \ref{tab:itps_maze} and Figure \ref{fig:itps_maze2d_qualitative}.

\subsection{Block Stacking - Discrete Task Alignment (\texttt{TA})}

We evaluate discrete task alignment by testing whether a multistep generative policy, with multimodal predictions at each step, can be steered to solve a long-horizon task following a user-preferred execution sequence. For this, we design a 4-block stacking domain in the Isaac Sim environment \cite{mittal2023orbit}. The simulation initializes four blocks at random positions, and motion trajectories are generated using CuRobo \cite{sundaralingam2023curobo}. The planner randomly selects blocks to pick and place, sometimes disassembling partial towers to rebuild them elsewhere. We train a DP (DDIM with $N=100$) on 5 million steps from this dataset to learn a motion manifold of valid pick-and-place actions without goal-oriented behavior. As shown in Figure \ref{fig:itps_block}(a), the learned policy exhibits multimodality across a discrete set of trajectories.

At inference time, we steer the policy to achieve a specific stacking sequence, completing a 4-block tower. To facilitate 3D steering, we develop a virtual reality (VR)-based system that allows users to provide 3D sketches within the simulation environment. In each interaction trial, the user observes the policy's unconditional rollouts before providing a sketch for conditional sampling. If the conditional sample with the smallest $\mathcal{L}_2$ distance to the sketch corresponds to the intended block, the trial is considered successfully aligned. If the policy execution also succeeds, the trial is deemed constraint-satisfying. We report \texttt{TA} and \texttt{CS} across interaction trials for \PR{} and \GD{} with $\beta_{i \leq N}=25$ in Table \ref{tab:itps_block}. Again, we see that higher \texttt{TA} correlates with lower \texttt{CS}.

Additionally, we experiment with a strategy to mitigate distribution shift during sampling with \GD{}. Rather than keeping the guide ratio $\beta_i$ constant for all $i = N, \dots, 1$, we deactivate steering by setting $\beta_{i \leq I} = 0$ for later steps. This approach aligns the low-frequency component of the noisy sample with user input in early diffusion steps while reverting to unconditional sampling after step $I$. Figure \ref{fig:itps_block}(c-d) demonstrate that the original \GD{} produces a curved trajectory resembling the sketch, while the modified \GD{} ($I = 50$) retrieves a straight-line trajectory from the CuRobo training dataset with the correct discrete alignment.

\subsection{Real World Kitchen - Discrete Task Alignment (\texttt{TA})}

\begin{wrapfigure}{l}{0.33\textwidth}
    \centering
    \includegraphics[width=0.33\textwidth]{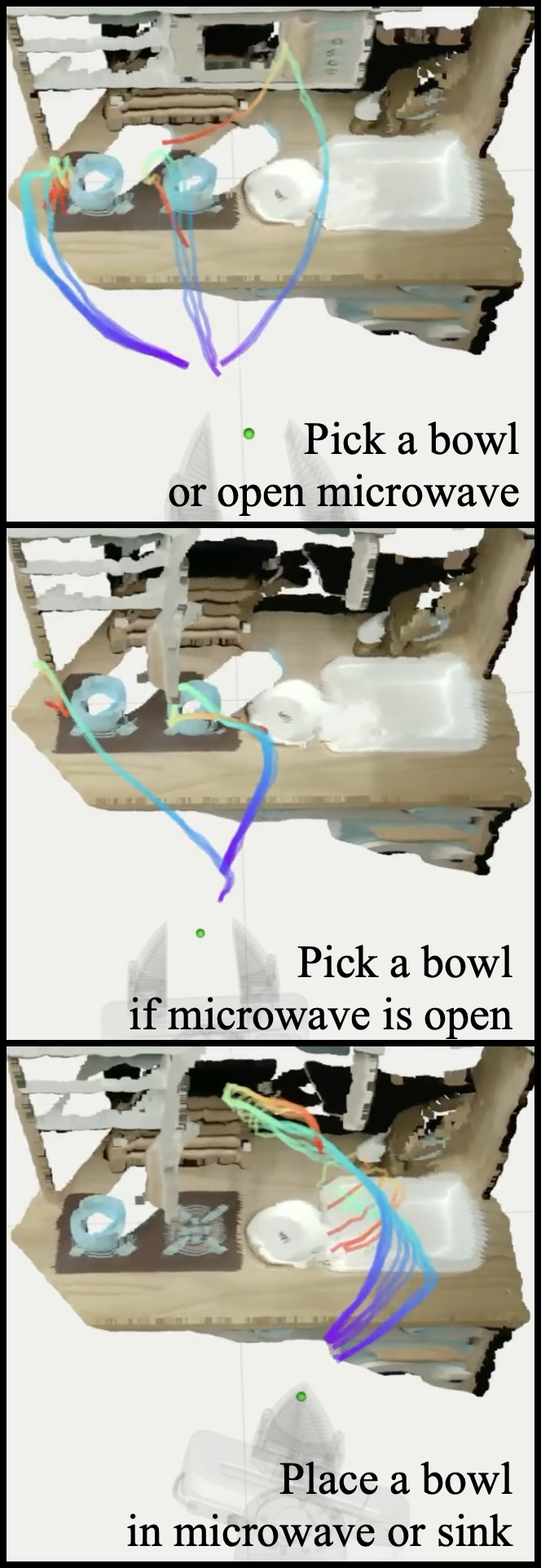} 
    \caption{\small \textbf{Multimodal Skills.}}
    \label{fig:itps_kitchen_multimodal}
\end{wrapfigure}

\begin{figure}[t] 
  \begin{minipage}[t]{\linewidth}
    \begin{minipage}[t]{\linewidth}
        \includegraphics[width=\linewidth]{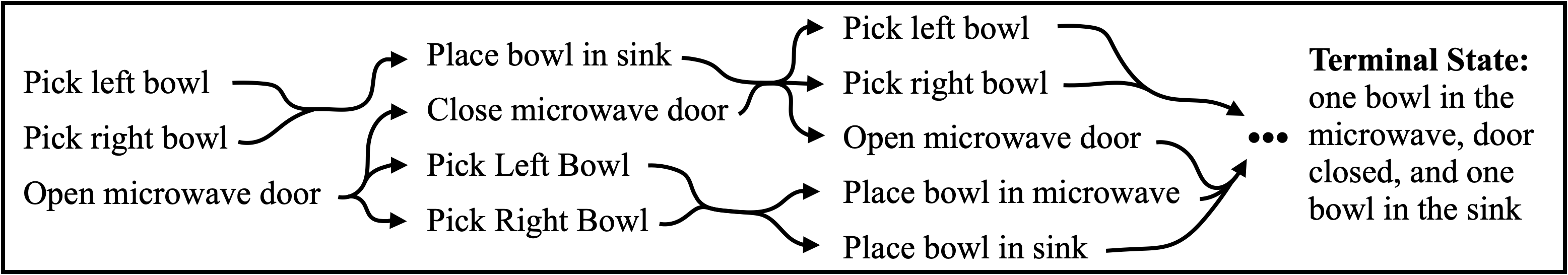}
        \caption{\small \textbf{Multimodal Valid Sequence for Kitchen Cleaning.} Steering selects a preferred legal sequence of skills to be executed until the terminal state is reached. This task requires a minimum of six steps.}
        \label{fig:itps_kitchen_tree}
        \vspace{5pt} 
    \end{minipage}
    
    \begin{minipage}[t]{\linewidth}
        \includegraphics[width=\linewidth]{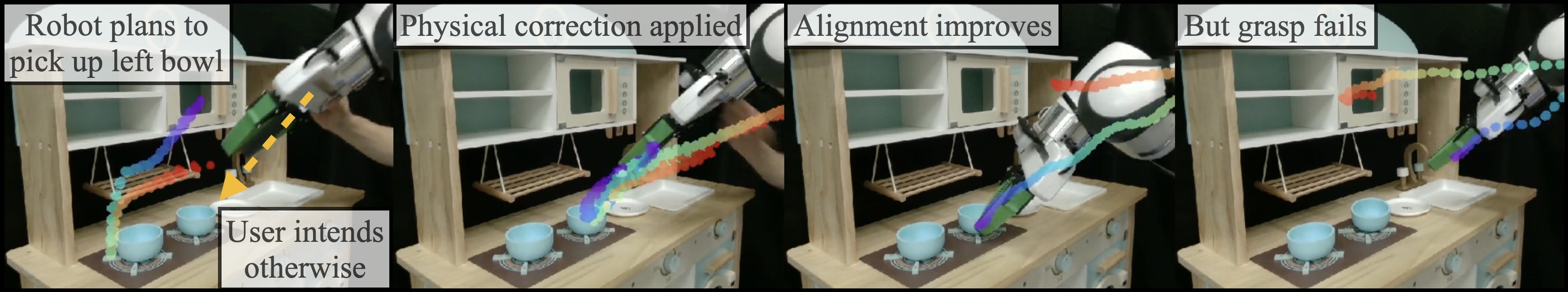}
        \caption{\small \textbf{Tradeoff Between Alignment and Distribution Shift.} As the user steers the policy to align with their intent, inference-time interactions may exacerbate distribution shift and lead to execution failure.}
        \label{fig:itps_kitchen_rollout}
        \vspace{5pt} 
    \end{minipage}

    \begin{minipage}[t]{\linewidth}
        \includegraphics[width=\linewidth]{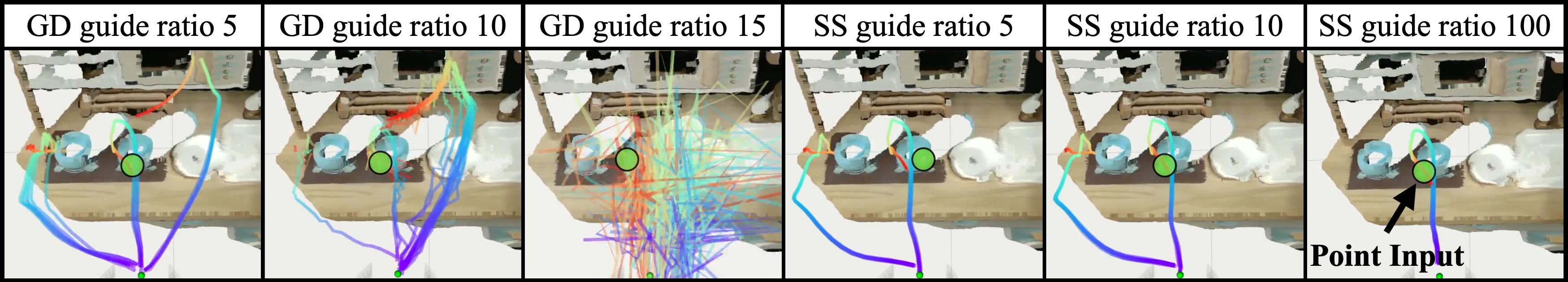}
        \caption{\small \textbf{Sensitivity to Guide Ratio $\beta_i$.} When $\beta_i$ is small, steering (via point input in this case) is ineffective for both \GD\ and \SGS. As $\xi_i$ increases, \GD\ begins to produce incoherent trajectories, while \SGS\ successfully identifies the intended skill. The same $\beta_i$ is applied for all $i \leq N$ ($N=100$).}
        \label{fig:itps_kitchen_ablation}
    \end{minipage}
  \end{minipage}
\end{figure}

To evaluate inference-time steering of multistep, multimodal policies in a real-world setting, we construct a toy kitchen environment and generate demonstrations using kinesthetic teaching. We focus on two tasks: (1) placing a bowl in the microwave and (2) placing a bowl in the sink. For each task, we collect 60 demonstrations and combine them into a dataset to train a diffusion policy (DP) over 40,000 steps. Figure \ref{fig:itps_kitchen_multimodal} illustrates that the learned motion manifold exhibits distinct multimodal skills based on the end-effector pose and gripper state. Unlike the block stacking experiment, merging datasets from different tasks introduces scenarios where skill sequences are not feasible—for example, placing a bowl in the microwave before opening the microwave door. Therefore, in this context, the \texttt{CS} metric not only measures the success of individual skills but also evaluates whether the resulting sequence is valid as shown in Figure \ref{fig:itps_kitchen_tree}.

At inference time, users can steer execution towards a preferred, valid sequence by clicking a pixel in the scene camera view to specify the intended skill. The corresponding 3D location of the pixel is visualized with a red sphere that turns green upon activation of the steering input. We also experiment with physical corrections to the end-effector pose to trigger behavior switches, but as shown in Figure \ref{fig:itps_kitchen_rollout}, this type of interaction often leads to execution failures. 

We evaluate the effectiveness of \GD, \SGS, and \OP{} in enabling users to achieve specific sequences of discrete skills. During real-time policy rollouts (7 Hz), users observe a randomly sampled skill and select a different one through interactions. We report whether the interaction successfully causes the intended behavior switch and whether it results in successful execution in Table \ref{tab:itps_kitchen}. For \GD, we use a guide ratio $\beta_i=5$ for all diffusion steps ($N=100$), while for \SGS{}, $\beta_i=100$ is used. These choices are based on the observation that increasing the guide ratio for \GD{ }disrupts the diffusion process without improving alignment (Figure \ref{fig:itps_kitchen_ablation}). In contrast, higher guide ratios for \SGS{} enhance alignment without producing noisy trajectories. Thus, \GD{} with $\beta_i=5$ serves as a baseline for weak steering, while \OP{}—allowing users to physically correct the robot end-effector trajectory during execution—functions as an aggressive steering baseline. Both \GD{} and \SGS{} are steered with pixel inputs. In Table \ref{tab:itps_kitchen}, as alignment \texttt{TA} increases, the constraint satisfaction rate \texttt{CS} decreases. The best steering method (\SGS) has a higher failure rate than rolling out randomly (\RS) but improves \textbf{Aligned Success by 21\%} without any fine-tuning.

\section{Related work}

\begin{table}[t]
\centering
\setlength{\tabcolsep}{0pt} 
\begin{tabularx}{0.8\linewidth}{@{}l*{4}{>{\centering\arraybackslash}X}@{}}
\toprule
\textbf{Method Type (DP)} & \textbf{\RS} & \textbf{\GD} & \textbf{\SGS} & \textbf{\OP} \\
\midrule
\textbf{Interaction Type} & \textbf{N/A} & \textbf{Point} & \textbf{Point} & \textbf{Correction} \\
\midrule
\tikzmark{ta}TA (Alignment: AS+AF) & 38\% & 37\% & 71\% & 89\% \\
\tikzmark{cs}CS (Success: AS+MS) & 90\% & 82\% & 73\% & 37\% \\
\tikzmark{as}\textbf{Aligned Success (AS)} & 34\% & 32\% & \textbf{55\%} & 30\% \\
\tikzmark{af}Aligned Failure (AF) & 4\% & 5\% & 16\% & 59\% \\
\tikzmark{ms}Misaligned Success (MS) & 56\% & 50\% & 18\% & 7\% \\
\tikzmark{mf}Misaligned Failure (MF) & 6\% & 13\% & 11\% & 4\% \\
\bottomrule
\end{tabularx}

\begin{tikzpicture}[overlay, remember picture]
    \draw[decorate, decoration={brace, mirror, amplitude=3pt}, thick]
        ($(as)+(-0.1cm, 0.2cm)$) -- ($(mf)+(-0.1cm, 0.0cm)$)
        node[midway, left=0.1cm]{\footnotesize 100\%};
\end{tikzpicture}

\caption{\small \textbf{Real World Kitchen Results.} We evaluate whether a user can steer a policy to switch from a randomly sampled skill to an intended skill and maintain successful execution. Overall, as alignment (\texttt{TA}) improves, the success rate (\texttt{CS}) decreases.}
\label{tab:itps_kitchen}
\end{table}

\textbf{Learning for Human-Robot Interaction.} Recently, learning from demonstrations \cite{chi2023diffusion,zhao2023learning} has achieved significant success in robotic manipulation. Despite this progress, real-time human input is often absent during inference-time policy rollouts. To address this gap, natural human-robot interfaces \cite{perzanowski2001building,berg2020review} have been employed when deploying robots in human environments. Various input forms, such as language, sketches, and goals \cite{team2024octo,brohan2023can,sundaresan2024rt,ding2019goal,mees2022matters,shi2024yell}, have also been studied to convey human intent to robots. Inspired by \cite{yoneda2023noise, ng2023diffusion}, our framework repurposes pre-trained generative policies for HRI settings, accommodating real-time human input. In this work, we focus on physical interactions, as they often provide grounding information that complements language prompts.

\textbf{Learning from Human Demonstrations.} Generative modeling \cite{urain2024deep,chi2023diffusion,lee2024behavior} has advanced imitation learning from multimodal, long-horizon demonstrations, enabling dexterous skill acquisition. Diffusion models \cite{chi2023diffusion}, are particularly effective at capturing the multimodal nature of human demonstrations, with their implicit function representation allowing flexible composition with external probability distributions \cite{janner2022planning, liu2022compositional, du2023reduce}. Previous research has explored using latent plans to support long-horizon tasks \cite{wang2022hierarchical,zhao2023learning,lynch2020learning}, but these focus on demonstrations with a single, high-quality behavior mode. In this work, we focus on generative modeling of multiple behavior modes \cite{wang2022temporal}, which is essential for enabling user interactions that require policies to adapt to inputs at inference time.

\textbf{Inference-Time Behavior Synthesis.} In robotics, inference-time composition has been explored as a method for achieving structured generalization \cite{du2019model,janner2022planning,gkanatsios2023energy, yang2023compositional, reuss2023goal,mishra2023generative,urain2023se}. Approaches like BESO \cite{reuss2023goal} leverage learned score functions combined with classifier-free guidance to enable goal-conditioned behavior generation. Similarly, SE3 Diffusion Fields \cite{urain2023se} use learned cost functions to generate gradients for joint motion and grasping planning, while V-GPS \cite{nakamoto2024steering} employs a learned value function to guide a generalist policy through re-ranking. PoCo \cite{wang2024poco} synthesizes behavior across diverse domains, modalities, constraints, and tasks through gradient-based policy composition, supporting out-of-distribution generalization. Building on PoCo, our work investigates how different types of real-time physical interaction can effectively steer policy at inference time.


\section{Conclusion}
In this work, we propose the Inference-Time Policy Steering (ITPS) framework, which integrates real-time human interactions to control policy behaviors during inference without requiring explicit policy training. We demonstrate how this approach enables humans to steer policies and benchmark several algorithms across both simulation and real-world experiments.
One limitation of our work is the reliance on an expensive sampling procedure to produce behaviors aligned with human intent. In future work, we aim to distill the steering process into an interaction-conditioned policy to achieve faster responses to human interactions and conduct a user study to further validate steerability. 

\chapter{Task and Motion Imitation with LTL Specification}

\begin{tcolorbox}
\textsc{$\textbf{Temporal Logic Imitation: Learning Plan-Satisficing Motion} \\
\textbf{Policies from Demonstrations}$ \\
\footnotesize Yanwei Wang, Nadia Figueroa, Shen Li, Ankit Shah, Julie Shah \\ 
\textbf{CoRL 2022 Oral}}
\end{tcolorbox}

\begin{flushright}
\vspace{1cm} 
\textit{"We must risk delight." \\
— Jack Gilbert}
\end{flushright}

\section{Introduction}
\begin{figure}[t] 
      \includegraphics[width=0.25\textwidth]{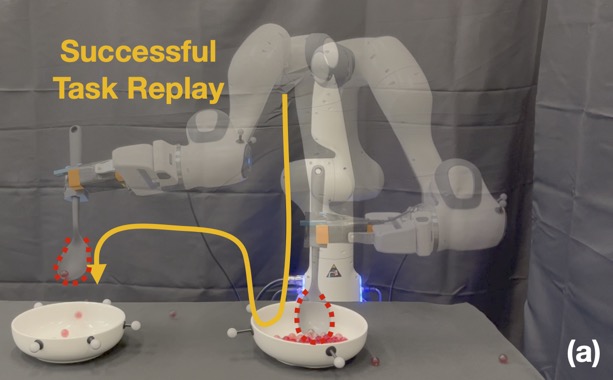}\includegraphics[width=0.25\textwidth]{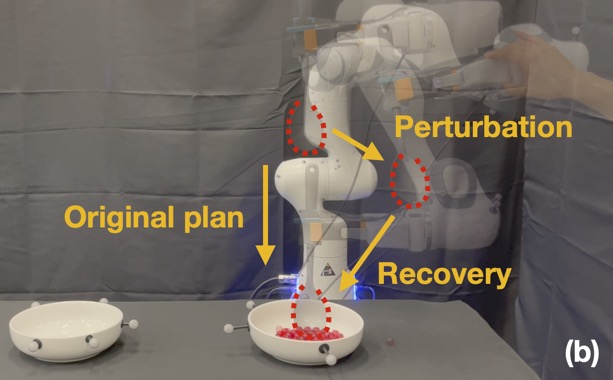}\includegraphics[width=0.25\textwidth]{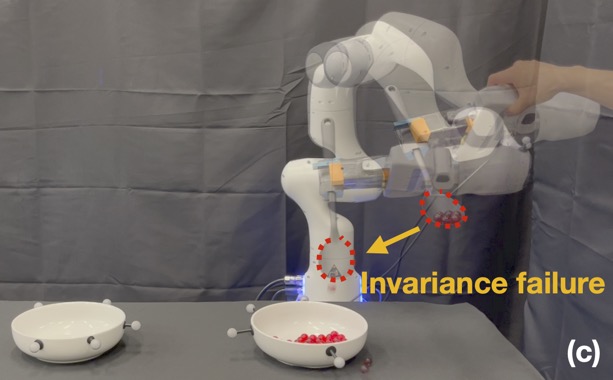}\includegraphics[width=0.25\textwidth]{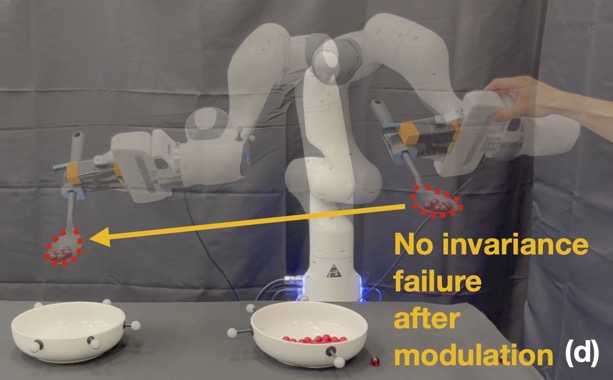}
     
      \caption{\textbf{(a)} A successful replay of the scooping task. The robot \textbf{(b)} is robust to motion-level perturbations; \textbf{(c)} experiences an invariance failure (i.e., drops material) after a task-level perturbation; and \textbf{(d)} re-scoops after a task-level perturbation, avoiding failure after DS motion policy modulation. 
      \label{fig:tli_robot_intro}
      }
\end{figure}

In prior work, learning from demonstration (LfD) \cite{argall2009survey, ravichandar2020recent} has successfully enabled robots to accomplish multi-step tasks by segmenting demonstrations (primarily of robot end-effector or tool trajectories) into sub-tasks/goals \cite{ekvall2008robot, grollman2010incremental, medina2017learning, gupta2019relay, mandlekar2020learning, pirk2020modeling}, phases \cite{pastor2012towards, kroemer2015towards}, keyframes \cite{akgun2012trajectories, perez2017c}, or skills/primitives/options \cite{konidaris2012robot, niekum2013incremental, fox2017multi, figueroa2019high}. Most of these abstractions assume reaching subgoals sequentially will deliver the desired outcomes; however, successful imitation of many manipulation tasks with spatial/temporal constraints cannot be reduced to imitation at the motion level unless the learned motion policy also satisfies these constraints. This becomes highly relevant if we want robots to not only imitate but also generalize, adapt and be robust to perturbations imposed by humans, who are in the loop of task learning and execution. LfD techniques that learn stable motion policies with convergence guarantees (e.g., Dynamic Movement Primitives (DMP) \cite{saveriano2021dmp}, Dynamical Systems (DS) \cite{DSbook}) are capable of providing such desired properties but only at the motion level. As shown in Fig.~\ref{fig:tli_robot_intro} (a-b) a robot can successfully replay a soup-scooping task while being robust to physical perturbations with a learned DS. Nevertheless, if the spoon orientation is perturbed to a state where all material is dropped, as seen in Fig.~\ref{fig:tli_robot_intro} (c), the motion policy will still lead the robot to the target, unaware of the task-level failure or how to recover from it. To alleviate this, we introduce an imitation learning approach that is capable of i) reacting to such task-level failures with Linear Temporal Logic (LTL) specifications, and ii) modulating the learned DS motion policies to avoid repeating those failures as shown in Fig.~\ref{fig:tli_robot_intro} (d). 

\begin{figure}[t]
    \includegraphics[trim={0cm 0cm 0 0},width=\textwidth, clip]{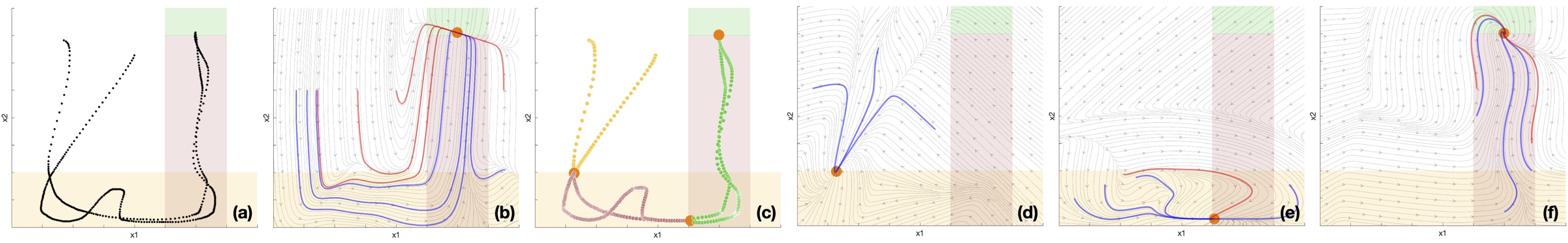}
    \caption{Mode abstraction of a 2D soup-scooping task: $x_1$ and $x_2$ denote the spoon's orientation and distance to the soup. \textbf{(a)} Task: To move the spoon's configuration from the white region (spoon without soup) $\Rightarrow$ yellow region (spoon in contact with soup) $\Rightarrow$ pink region (spoon holding soup) $\Rightarrow$ green region (soup at target). (Note that transitions (white $\Rightarrow$ pink) and (white $\Rightarrow$ green) are not physically realizable.) Black curves denote successful demonstrations. \textbf{(b)} Learning DS policies \cite{figueroa2018physically} over unsegmented data can result in successful task replay (blue trajectories), but lacks a guarantee due to invalid transitions (red trajectories). \textbf{(c)} Trajectories are segmented into three colored regions (modes) with orange attractors. \textbf{(d-f)} Learning DSs on segments may still result in \textit{invariance failures} (i.e., traveling outside of modes as depicted by red trajectories).
    \label{fig:tli_intro} 
    }
\end{figure}

\textbf{Example.} We demonstrate that successfully reaching a goal via pure motion-level imitation does not imply successful task execution. The illustrations in Fig. \ref{fig:tli_intro} represent a 2D simplification of the soup-scooping task, where task success requires a continuous trajectory to simulate a discrete plan of consecutive transitions through the colored regions. Human demonstrations, shown in Fig.~\ref{fig:tli_intro}~\textbf{(a)}, are employed to learn a DS policy \cite{figueroa2018physically}, depicted by the streamlines in Fig.~\ref{fig:tli_intro}~\textbf{(b)}. The policy is stress-tested by applying external perturbations, displacing the starting states of the policy rollouts. As shown, only blue trajectories succeed in the task, while the red ones fail due to discrete transitions that are not physically realizable (e.g., white $\Rightarrow$ pink). As shown in Fig.~\ref{fig:tli_intro}~ \textbf{(c-f)}, even if the demonstrations are further segmented by subgoals (and corresponding DS policies are learned), this issue is not mitigated. While one could treat this problem as covariate shift and solve it by asking a human for more demonstrations \cite{ross2011reduction}, in this work, we frame it as the mismatch between a learned continuous policy and a discrete task plan specified by the human in terms of a logical formula. Specifically, the core challenges illustrated by this example are two-fold: 1) subgoals only impose point constraints that are insufficient to represent the boundary of a discrete abstraction; and 2) the continuous policy can deviate from a demonstrated discrete plan when perturbed to unseen parts of the state space, and is incapable of replanning to ensure all discrete transitions are valid.

To address these challenges, our proposed approach employs ``modes” as discrete abstractions. We define a \textit{mode} as a set of robot and environment configurations that share the same sensor reading \cite{van2000introduction, garrett2021integrated}; e.g., in Fig. \ref{fig:tli_intro}, each colored region is a unique mode, and every mode has a boundary that imposes path constraints on motion policies. Additionally, we use a task automaton as a receding-horizon controller that replans when a perturbation causes the system to travel outside a mode boundary and triggers an unexpected sensor change; e.g., detecting a transition from yellow $\Rightarrow$ white instead of the desired yellow $\Rightarrow$ pink will result in a new plan: white $\Rightarrow$ yellow $\Rightarrow$ pink $\Rightarrow$ green. In this work, we synthesize a task automaton from a linear temporal logic formula (LTL) that specifies all valid mode transitions. We denote the problem of learning a policy that respects these mode transitions from demonstrations as \textit{temporal logic imitation} (TLI). In contrast to temporal logic planning (TLP) \cite{kress2018synthesis}, where the workspace is partitioned into connected convex cells with known boundaries, we do not know the precise mode boundaries. Consequently, the learned policy might prematurely exit the same mode repeatedly, causing the task automaton to loop without termination. To ensure any discrete plan generated by the automaton is feasible for the continuous policy, the bisimulation criteria \cite{alur2000discrete, fainekos2005temporal} must hold for the policy associated with each mode. Specifically, any continuous motion starting in any mode should stay in the same mode \textbf{(invariance)} until eventually reaching the next mode \textbf{(reachability)}. The violations of these conditions are referred to as \textit{invariance failures} and \textit{reachability failures} respectively.

\textbf{Contributions.} First, we investigate TLP in the setting of LfD and introduce TLI as a novel formulation to address covariate shift by proposing imitation with respect to a mode sequence instead of a motion sequence. Second, leveraging modes as the discrete abstraction, we prove that a state-based continuous behavior cloning (BC) policy with a global stability guarantee can be modulated to simulate any LTL-satisficing discrete plan. Third, we demonstrate that our approach LTL-DS, adapts to task-level perturbations via an LTL-satisficing automaton's replanning and recovers from motion-level perturbations via DS' stability during a multi-step, non-prehensile manipulation task.

\section{Temporal Logic Imitation: Problem Formulation}
\label{sec:tli}
Let $x \in \mathbb{R}^n$ represent the $n$-dimensional continuous state of a robotic system; e.g., the robot's end-effector state in this work. Let $\alpha = [\alpha_1,...,\alpha_m]^T \in \{0, 1\}^m$ be an $m$-dimensional discrete sensor state that uniquely identifies a mode $\sigma = \mathcal{L}(\alpha)$. We define a system state as a tuple, $s = (x, \alpha) \in \mathbb{R}^n \times \{0, 1\}^m$. Overloading the notation, we use $\sigma \in \Sigma$, where $\Sigma = \{\sigma_i\}_{i=1}^{\mathcal{M}}$, to represent the set of all system states within the same mode---i.e., $ \sigma_i = \{s=(x, \alpha) \mid \mathcal{L}(\alpha) = \sigma_i\}$. In contrast, we use $\delta_i=\{x | s=(x,\alpha) \in \sigma_i\}$ to represent the corresponding set of robot states. Note $x$ cannot be one-to-one mapped to $s$, e.g., a level spoon can be either empty or holding soup. Each mode is associated with a goal-oriented policy, with goal $x^*_i\in\mathbb{R}^n$. A successful policy that accomplishes a multi-step task $\tau$ with a corresponding LTL specification $\phi$ can be written in the form:
\begin{equation}
\label{eq:tli_imitation_pi}
\begin{aligned}
& \dot{x} =  \pi(x,\alpha; \phi) = \Sigma_{i=1}^{\mathcal{M}}\delta_{\Omega_{\phi}(\alpha)\sigma_i} f_i(x;\theta_i, x^*_i) 
\end{aligned}
\end{equation}
with $\delta_{\Omega_{\phi}(\alpha)\sigma_i}$ being the Kronecker delta that activates a mode policy $f_i(x;\theta_i,x^*_i):\mathbb{R}^{n}\rightarrow\mathbb{R}^n$ encoded by learnable parameters $\theta_i$ and goal $x^*_i$. Mode activation is guided by an LTL-equivalent automaton $\Omega_{\phi}(\alpha) \rightarrow \sigma_i$ choosing the next mode $\sigma_i$ based on the current sensor reading $\alpha$. 

\textbf{Demonstrations.} Let demonstrations for a task $\tau$ be $\Xi= \{\{x^{t,d},\dot{x}^{t,d}, \alpha^{t,d}\}_{t=1}^{T_d}\}_{d=1}^{D}$ where $x^{t,d}\text{, } \dot{x}^{t,d}\text{, } \alpha^{t,d}$ are robot state, velocity, and sensor state at time $t$ in demonstration $d$, respectively, and $T_d$ is the length of each $d$-th trajectory. A demonstration is successful if the continuous motion traces through a sequence of discrete modes that satisfies the corresponding LTL task specification.

\textbf{Perturbations.} External perturbations, which many works in Sec. \ref{sec:tli_related} avoid, constitute an integral part of our task complexity. Specifically, we consider (1) motion-level perturbations that displace a continuous motion within the same mode, and (2) task-level perturbations that drive the robot outside of the current mode. Critically, motion-level perturbations do not cause a plan change instantaneously, but they can lead to future unwanted mode transitions due to covariate shift. 

\textbf{Problem Statement.} Given (1) an LTL formula $\phi$ specifying valid mode transitions for a task $\tau$, (2) 
sensors that detect each mode abstraction defined in $\phi$, and (3) successful demonstrations $\Xi$, we seek to learn a policy defined in Eq. \ref{eq:tli_imitation_pi} that generates continuous trajectories guaranteed to satisfy the LTL specification despite arbitrary external perturbations. 

\section{Preliminaries} 
\label{sec:tli_prelims}
\subsection{LTL Task Specification}\label{sec:tli_ltl_spec} 
LTL formulas consist of atomic propositions (AP), logical operators, and temporal operators \cite{shah2018bayesian, kress2018synthesis}. Let $\Pi$ be a set of Boolean variables; an infinite sequence of truth assignments to all APs in $\Pi$ is called the trace $[\Pi]$. The notation $[\Pi],t\models\phi$ means the truth assignment at time $t$ satisfies the LTL formula $\phi$. Given $\Pi$, the minimal syntax of LTL can be described as:
\begin{equation} \label{eq:tli_ltl}
    \phi ::= p \mid \neg\phi_1 \mid \phi_1 \vee \phi_2 \mid \textbf{X}\phi_1 \mid \phi_1\textbf{U}\phi_2
\end{equation}
where $p$ is any AP in $\Pi$, and $\phi_1$ and $\phi_2$ are valid LTL formulas constructed from $p$ using Eq. \ref{eq:tli_ltl}. The operator \textbf{X} is read as `next,' and $\textbf{X}\phi_1$ intuitively means the truth assignment to APs at the next time step sets $\phi_1$ as true. \textbf{U} is read as `until' and, intuitively, $\phi_1\textbf{U}\phi_2$ means the truth assignment to APs sets $\phi_1$ as true until $\phi_2$ becomes true. Additionally, first-order logic operators $\neg$ (not), $\land$ (and), $\lor$ (or), and $\rightarrow$ (implies), as well as higher-order temporal operators \textbf{F} (eventually), and \textbf{G} (globally), are incorporated. Intuitively, $\textbf{F}\phi_1$ means the truth assignment to APs eventually renders $\phi_1$ true and $\textbf{G}\phi_1$ means truth assignment to APs renders $\phi_1$ always true from this time step onward. 
\subsection{Task-Level Reactivity in LTL} \label{sec:tli_fsm}
To capture the reactive nature of a system given sensor measurements, the \textit{generalized reactivity (1)} (GR(1)) fragment of LTL \cite{piterman2006synthesis, kress2009temporal} can be used. Let the set of all APs be $\Pi = \mathcal{X}\cup\mathcal{Y}$, where sensor states form environment APs $\mathcal{X} = \{\alpha_1,...,\alpha_m\}$ and mode symbols form system APs $\mathcal{Y} = \{\sigma_1,...,\sigma_l\}$. A GR(1) formula is of the form $\phi = (\phi_e \rightarrow \phi_s)$ \cite{piterman2006synthesis}, where $\phi_e$ models the assumed environment behavior and $\phi_s$ models the desired system behavior. Specifically,
\begin{equation} \label{eq:gr1}
    \phi_e=\phi^e_i\land \phi^e_t\land \phi^e_g, \quad\quad \phi_s=\phi^s_i\land \phi^s_t\land \phi^s_g
\end{equation}
$\phi^e_i$ and $\phi^s_i$ are non-temporal Boolean formulas that constrain the initial truth assignments of $\mathcal{X}$ and $\mathcal{Y}$ (e.g., the starting mode). $\phi^s_t$ and $\phi^e_t$ are LTL formulas categorized as safety specifications that describe how the system and environment should always behave (e.g., valid mode transitions). $\phi_g^s$ and $\phi_g^e$ are LTL formulas categorized as liveness specifications that describe what goal the system and environment should eventually achieve (e.g., task completion) \cite{kress2018synthesis}.
The formula $\phi$ guarantees the desired system behavior specified by $\phi_s$ if the environment is \textit{admissible}---i.e., $\phi_e$ is true---and can be converted to an automaton $\Omega_{\phi}$ that plans a mode sequence satisfying $\phi$ by construction \cite{kress2009temporal}.

\subsection{Motion-Level Reactivity in DS} \label{sec:tli_ds_learning}
Dynamical System \cite{figueroa2018physically} is a state-based BC method with a goal-reaching guarantee despite arbitrary perturbations. A DS policy can be learned from as few as a single demonstration and has the form:


\begin{minipage}{.4\textwidth}
\begin{equation}
\label{eq:tli_ds_eq}
  \dot{x} = f(x) = \sum_{k=1}^{K}\gamma_k(x)(A^kx+b^k)
\end{equation}
\end{minipage}\hspace{1pt} 
\begin{minipage}{.575\textwidth}
\begin{equation}
\label{eq:tli_stability}
  \begin{cases}
    (A^k)^T P+PA^k =Q^k, Q^k=(Q^k)^T \prec 0 \\
    b^k=-A^k x^*
    \end{cases}
    \forall k
\end{equation}
\end{minipage}

where $A^k\in \mathbb{R}^{n\times n}$, $b^k\in \mathbb{R}^n$ are the k-th linear system parameters, and $\gamma_k(x):\mathbb{R}^n \rightarrow \mathbb{R}^+$ is the mixing function. To certify global asymptotic stability (G.A.S.) of Eq. \ref{eq:tli_ds_eq}, a Lyapunov function $V(x)=(x-x^*)^T P(x-x^*)$ with $P=P^T\succ 0$, is used to derive the stability constraints in Eq. \ref{eq:tli_stability}. Minimizing the fitting error of Eq. \ref{eq:tli_ds_eq} with respect to demonstrations $\Xi$ subject to constraints in Eq. \ref{eq:tli_stability} yields a non-linear DS with a stability guarantee \cite{figueroa2018physically}. To learn the optimal number $K$ and mixing function $\gamma_k(x)$ we use the Bayesian non-parametric GMM fitting approach presented in \cite{figueroa2018physically}. 


\subsection{Bisimulation between Discrete Plan and Continuous Policy}
To certify a continuous policy will satisfy an LTL formula $\phi$, one can show the policy can simulate any LTL-satisficing discrete plan of mode sequence generated by $\Omega_{\phi}$. To that end, every mode's associated policy must satisfy the following bisimulation conditions \cite{fainekos2005temporal, kress2018synthesis}:

\begin{condition}[\textbf{Invariance}] Every continuous motion starting in a mode must remain within the same mode while following the current mode's policy; i.e., $\forall i \ \forall t \ (s^0 \in \sigma_i \rightarrow s^t \in \sigma_i)$
\end{condition}

\begin{condition}[\textbf{Reachability}] 
Every continuous motion starting in a mode must reach the next mode in the demonstration while following the current mode's policy; i.e., $\forall i \ \exists T \ (s^0 \in \sigma_i \rightarrow s^{T} \in \sigma_j)$
\end{condition}


\section{Method}
\label{sec:tli_ltlds}
To solve the TLI problem in Sec. \ref{sec:tli}, we introduce a mode-based imitation policy---LTL-DS:
\begin{equation}
\label{eq:tli_ltlds_pi}
\dot{x} = \pi(x,\alpha; \phi) = \textcolor{blue}{\underbrace{\Sigma_{i=1}^{\mathcal{M}}\delta_{\Omega_{\phi}(\alpha)\sigma_i}}_{\text{offline learning}}} \textcolor{orange}{\underbrace{M_i\big(x;\Gamma_i(x), x^*_i\big)}_{\text{online learning}}}\textcolor{blue}{\underbrace{f_i(x;\theta_i, x^*_i)}_{\text{offline learning}}},
\end{equation}
During offline learning, we synthesize the automaton $\Omega_{\phi}$ from $\phi$ as outlined in Sec. \ref{sec:tli_fsm} and learn DS policies $f_i$ from $\Xi$ according to Sec. \ref{sec:tli_ds_learning}. While the choice of DS satisfies the reachability condition as explained later, nominal DS rollouts are not necessarily bounded within any region. Neither do we know mode boundaries in TLI. Therefore, an online learning phase is necessary, where for each mode policy $f_i$ we learn an implicit function, $\Gamma_i(x): \mathbb{R}^n \rightarrow \mathbb{R^+}$, that inner-approximates the mode boundary in the state-space of a robot $x\in\mathbb{R}^n$. With a learned $\Gamma_i(x)$ for each mode, we can construct a modulation matrix $M_i$ that ensures each modulated DS---$M_if_i$---is mode invariant.

\begin{figure}[!tbp]
  \centering
  \includegraphics[width=0.95\textwidth]{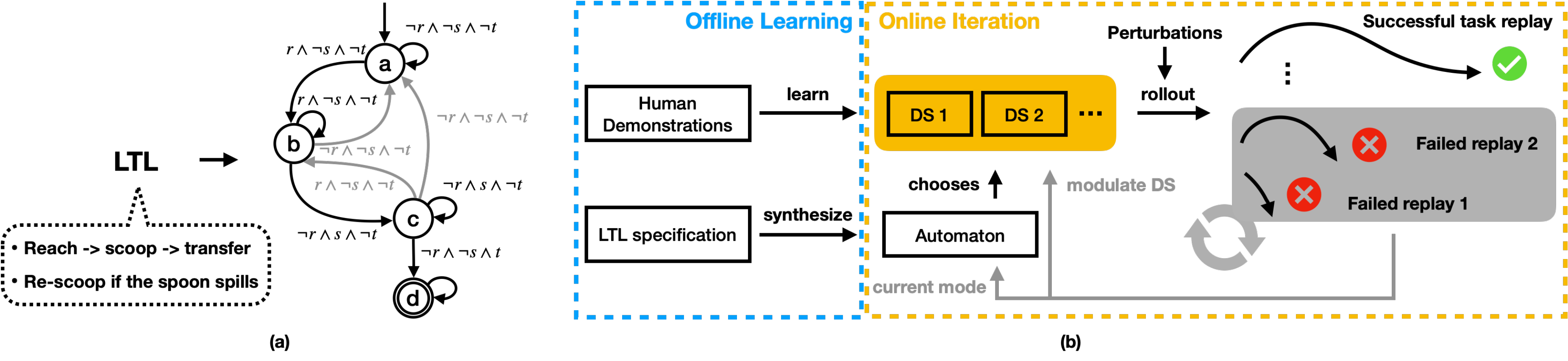}
  \caption{\textbf{(a)} Task automaton for a scooping task LTL. Mode $a, b, c, d$ are reaching, scooping, transporting, and done mode respectively. Atomic proposition $r, s, t$ denote sensing the spoon reaching the soup, soup on the spoon, and task success respectively. 
  During successful demonstrations, only mode transitions in black, $a \Rightarrow b \Rightarrow c \Rightarrow d$, are observed. Additional valid transitions in gray, $b \Rightarrow a$, $c \Rightarrow a$, and $c \Rightarrow b$, are given by the LTL to help recover from unexpected mode transitions. 
  \textbf{(b)} System flowchart of LTL-DS.}
  \label{fig:tli_method} 
\end{figure} 

\subsection{Offline Learning Phase}
\textbf{Synthesis of LTL-Satisficing Automaton}
We convert an LTL to its equivalent automaton with \cite{duret.16.atva2}, which plans the next mode given the current sensor reading. Assuming all possible initial conditions for the system are specified in the LTL, the automaton is always deployed from a legal state.

\textbf{Sensor-based Motion Segmentation and Attractor Identification} Given demonstrations in $\Xi$ and accompanying sensor readings related to the set of $\mathcal{M}$ modes, we can automatically segment the trajectories into $\mathcal{M}$ clusters and corresponding attractor set $X^*$. Refer to Appendix \ref{sec:tli_segmentation} for details.

\textbf{Ensuring Goal Reachability with Learned DS Mode Policies}
While any BC variant with a stability guarantee can satisfy reachability (see Sec. \ref{sec:tli_related}), we focus on the G.A.S. DS formulation and learning approach defined in Section \ref{sec:tli_ds_learning} that ensures every $x\in \mathbb{R}^n$ is guaranteed to reach $x^*_{i}$. By placing $x^*_{i}$ within the boundary set of $\delta_j$ for a mode $\sigma_j$, we ensure mode $\sigma_j$ is reachable from every $s$ in mode $\sigma_i$. Note $f(x)$ cannot model sensor dynamics in $\alpha$. Yet, we employ mode abstraction to reduce the imitation of a system state trajectory in $s$---which includes the evolution of both the robot and sensor state---to just a robot state trajectory in $x$. 

\subsection{Online Learning Phase}
\label{sec:tli_invariance}
\begin{figure}[!tbp]
    \includegraphics[trim={0cm 0cm 0 0},width=\textwidth, clip]{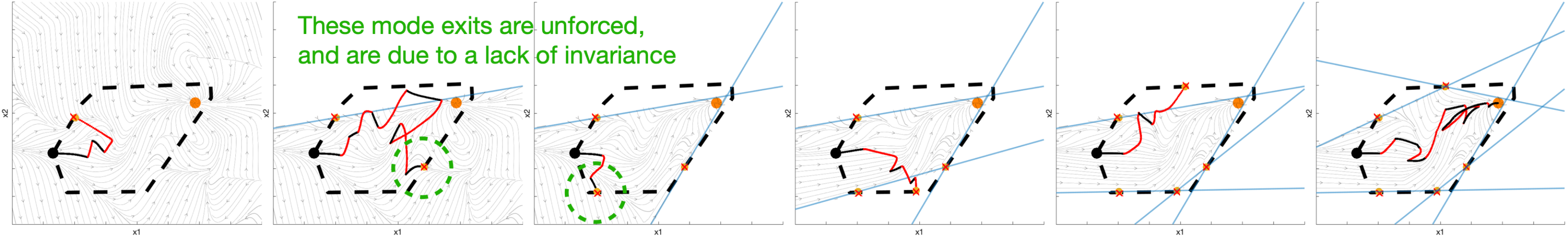}
    \caption{An illustration of iterative estimation of a mode boundary with cutting planes. A system enters a mode with an unknown boundary (dashed line) at the black circle, and is attracted to the goal at the orange circle. The trajectory in black shows the original policy rollout, and the trajectory in red is driven by perturbations. After the system exits the mode and before it eventually re-enters the same mode through replanning, a cut is placed at the last in-mode state (yellow circle) to bound the mode from the failure state (red cross). When the system is inside the cuts, it experiences modulated DS and never moves out of the cuts (flows moving into the cuts are not modulated); when the system is outside the cuts but inside the mode, it follows the nominal DS. Note only mode exits in black are invariance failures in need of modulation (green circles); mode exits in red are driven by perturbations to illustrate that more cuts lead to better boundary approximation.\label{fig:tli_cut}
    }
\end{figure}

\textbf{Iterative Mode Boundary Estimation via Invariance Failures} As shown in Fig. \ref{fig:tli_intro}, DS can suffer from \textit{invariance} failures in regions without data coverage. Instead of querying humans for more data in those regions \cite{ross2011reduction}, we leverage sparse events of mode exits detected by sensors to estimate the unknown mode boundary. Specifically, for each invariance failure, we construct a cut that separates the failure state, $x^{T_f}$, from the mode-entry state, $x^{0}$, the last in-mode state, $x^{T_f-1}$, and the mode attractor, $x^*$. We ensure this separation constraint with a quadratically constrained quadratic program (QCQP) that searches for the normal direction (pointing away from the mode) of a hyper-plane that passes through each $x^{T_f-1}$ such that the plane's distance to $x^*$ is minimized. The intersection of half-spaces cut by hyper-planes inner approximates a convex mode boundary, as seen in Fig. \ref{fig:tli_cut}. Adding cuts yields better boundary estimation, but is not necessary unless the original vector field flows out of the mode around those cuts. For more details, refer to Appendix \ref{sec:tli_qcqp}.

\textbf{Ensuring Mode Invariance by Modulating DS} We treat each cut as a collision boundary that deflects DS flows following the approach in \cite{khansari2012dynamical,huber2019avoidance}. In our problem setting the mode boundary is analogous to a workspace enclosure rather than a task-space object. Let existing cuts form an implicit function, $\Gamma(x): \mathbb{R}^n \rightarrow \mathbb{R^+}$, where $\Gamma(x)<1$, $\Gamma(x)=1, \Gamma(x)>1$ denote the estimated interior, the boundary and the exterior of a mode. $0<\Gamma(x)<\infty$ monotonically increases as $x$ moves away from a reference point $x^r$ inside the mode. For $x$ outside the cuts, or inside but moving away from the cuts, we leave $f(x)$ unchanged; otherwise, we modulate $f(x)$ to not collide with any cuts as $\dot{x}=M(x)f(x)$ by constructing a modulation matrix $M(x)$ through eigenvalue decomposition:
\begin{equation}\label{eq:tli_modulate}
    \begin{cases}
        M(x) = E(x)D(x)E(x)^{-1},~~ 
        E(x) = [\mathbf{r}(x)\text{ } \mathbf{e}_1(x)\text{ } ...\text{ } \mathbf{e}_{d-1}(x)],~~\mathbf{r}(x) = \frac{x-x^r}{\|x-x^r\|}\\ D(x) = 
        \textbf{diag(}\lambda_r(x), \lambda_{e_1}(x),...,\lambda_{e_{d-1}}(x)\textbf{)},~~
        \lambda_r(x) = 1-\Gamma(x),~~\lambda_e(x) = 1
    \end{cases}
\end{equation}
The full-rank basis $E(x)$ consists of a reference direction $\mathbf{r}(x)$ stemming from $x^r$ toward $x$, and $d-1$ directions spanning the hyperplane orthogonal to $\nabla\Gamma(x)$, which in this case is the closest cut to $x$. In other words, all directions $\mathbf{e}_1(x)...\mathbf{e}_{d-1}(x)$ are tangent to the closest cut, except $\mathbf{r}(x)$. By modulating only the diagonal component, $\lambda_r(x)$, with $\Gamma(x)$, we have $\lambda_r(x) \rightarrow 0$ as $x$ approaches the closest cut, effectively zeroing out the velocity penetrating the cut while preserving velocity tangent to the cut. Consequently, a modulated DS will not repeat invariance failures that its nominal counterpart experiences as long as the mode is bounded by cuts. Notice this modulation strategy is not limited to DS and can be applied to any state-based BC method to achieve mode invariance.

\section{Proof}
Next, we prove LTL-DS produces a continuous trajectory that satisfies an LTL specification. We start with assumptions and end with theorems. Detailed proofs are provided in Appendix \ref{sec:tli_proof}. 
\begin{assumption} \label{eq:tli_convexity}
    All modes are convex. 
\end{assumption}
This assumption leads to the existence of at least one cut---i.e., the supporting plane \cite{boyd2004convex}, which can separate a failure state on the boundary from any other state within the mode. A corollary is that the boundary shared by two modes, which we call a guard surface, $G_{ij} = \delta_i \cap \delta_j$, is also convex. Since all transitions out of a mode observed during demonstrations reside on the mode boundary, their average location, which we use as the attractor for the mode, will also be on the boundary.

\begin{assumption} \label{eq:tli_finiteness}
    There are a finite number of external perturbations of arbitrary magnitudes.
\end{assumption}
Given zero perturbation, all BC methods should succeed in any task replay, as the policy rollout will always be in distribution. If there are infinitely many arbitrary perturbations, no BC methods will be able to reach a goal. In this work, we study the setting in between, where there are finitely many motion- and task-level perturbations causing unexpected mode exits. Environmental stochasticity is ignored, as its cumulative effects can also be simulated by external perturbations. 

\begin{assumption} \label{eq:tli_non-oracle}
    Perturbations only cause transitions to modes already seen in the demonstrations.
\end{assumption}
While demonstrations of all valid mode transitions are not required, they must minimally cover all possible modes. If a system encounters a completely new sensor state during online interaction, it is reasonable to assume that no BC methods could recover from the mode unless more information about the environment is provided. 
\begin{theorem} \label{eq:tli_kc1}
    (Key Contribution 1) A nonlinear DS defined by Eq. \ref{eq:tli_ds_eq}, learned from demonstrations, and modulated by cutting planes as described in Section \ref{sec:tli_invariance} with the reference point $x^r$ set at the attractor $x^*$, will never penetrate the cuts and is G.A.S. at $x^*$. \hfill\textbf{Proof:} See Appendix \ref{sec:tli_proof}.~~\qedsymbol{}
\end{theorem} 
\begin{theorem} (Key Contribution 2) \label{eq:tli_kc2}
    The continuous trace of system states generated by LTL-DS defined in Eq. \ref{eq:tli_ltlds_pi} satisfies any LTL specification $\phi$ under Asm. \ref{eq:tli_convexity}, \ref{eq:tli_finiteness}, and \ref{eq:tli_non-oracle}.~\textbf{Proof:} See Appendix \ref{sec:tli_proof}.\qedsymbol{} 
\end{theorem}

\section{Experiments}
\subsection{Single-Mode Invariance and Reachability}

\begin{figure}[!tbp]
    \includegraphics[width=1\linewidth]{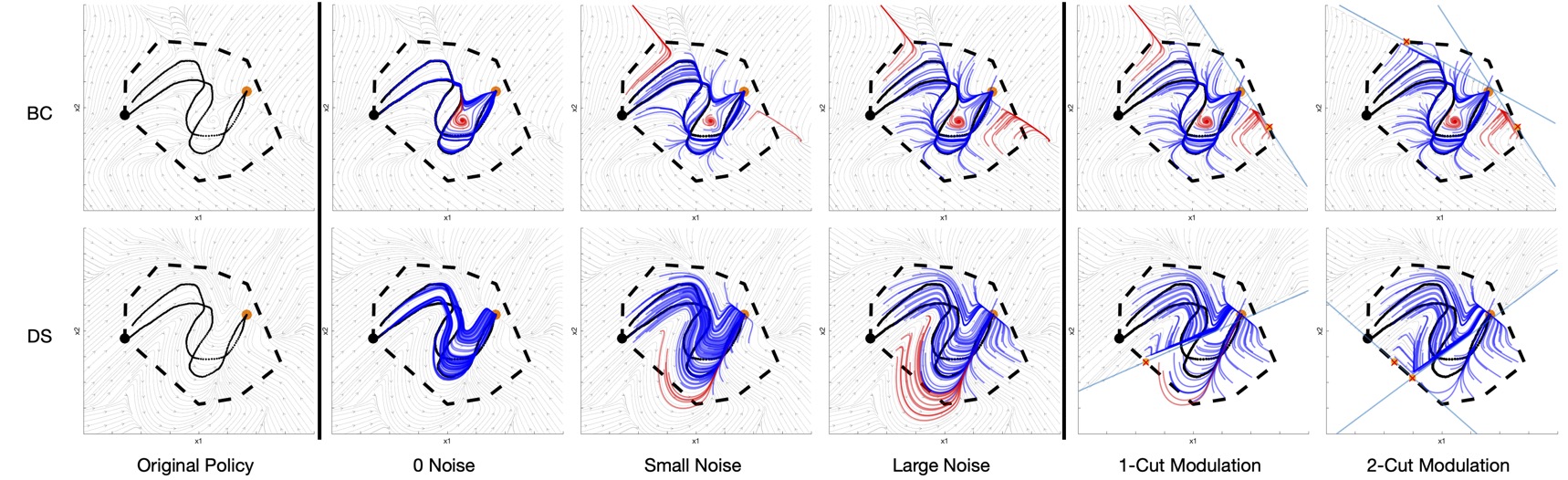}
\caption{Policy rollouts from different starting states for a randomly generated convex mode. The top row shows BC results, and the bottom row depicts DS results. The left column visualizes the nominal policies learned from two demonstrations (black trajectories) reaching the orange attractor. The middle columns add different levels of Gaussian noise to the initial states sampled from the demonstration distribution. Blue trajectories successfully reach the attractor, while red trajectories fail due to either invariance failures or reachability failures. (Note that these failures only occur at locations without data coverage.) The right columns show that cutting planes (blue lines) separate failures (red crosses) from last-visited in-mode states (yellow circles) and consequently bound both policies to be mode-invariant. Applying cutting planes to BC policies without a stability guarantee cannot correct reachability failures within the mode. More results in Appendix \ref{sec:tli_single-mode}.}\label{fig:tli_rollout}
\end{figure}

\begin{figure}[t]
    \centering
    \resizebox{0.6\columnwidth}{!}{%
    \begin{tabular}{cccccc}
    \hline
    Policy & Reachability & Invariance & No Noise & Small Noise & Large Noise \\ \hline
    BC     & \color{red}\xmark & \color{red}\xmark  & 88.9 & 72.4 & 58.6  \\
    BC+mod & \color{red}\xmark & \color{green}{\cmark}  & 91.9 & 83.6 & 76.0  \\
    DS     & \color{green}{\cmark} & \color{red}\xmark  & 100  & 97.0 & 86.9  \\
    DS+mod & \color{green}{\cmark} & \color{green}{\cmark}  & \textbf{100}  & \textbf{100}  & \textbf{100}   \\ 
    \hline
    \end{tabular}}
    \vspace{5pt}
    \\
    \includegraphics[width=0.6\linewidth]{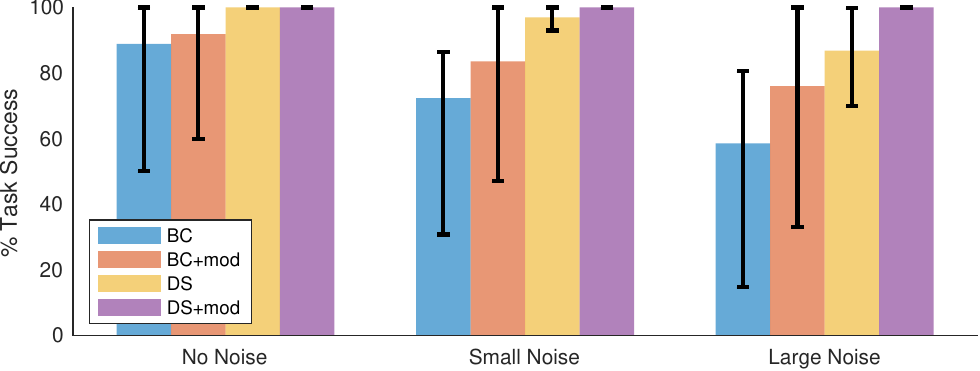}
    \caption{The success rate (\%) of a single-mode reaching task. As we begin to sample out of distribution by adding more noise to the demonstrated states, the BC's success rate degrades more rapidly than the DS'. After modulation, DS (+mod) maintains a success guarantee, which BC (+mod) falls short of due to the base policy's lack of a stability guarantee.}
    \label{fig:tli_result} 
\end{figure}

\begin{figure}[h!]
    \centering
  \includegraphics[width=0.6\linewidth]{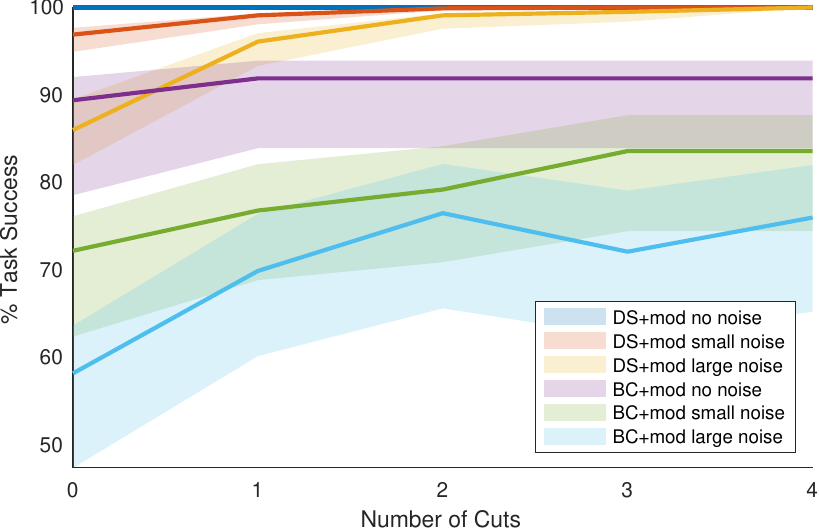}
  \caption{Empirically, the invariance of a single mode requires only a finite number of cuts for a nominal policy with a stability guarantee. Regardless of the noise level, DS achieves a $100\%$ success rate after four cuts, while BC struggles to improve performance with additional cuts. Thick lines represent mean statistics and shaded regions show the interquartile range. More details in Appendix \ref{sec:tli_single-mode}.}
\end{figure}

We show quantitatively  both reachability and invariance are necessary for task success. We compare DS and a NN-based BC policy (denoted as BC) to represent policies with and without a stability guarantee. Figure \ref{fig:tli_rollout} shows that policy rollouts start to fail (turn red) as increasingly larger perturbations are applied to the starting states; however, DS only suffers from invariance failures, while BC suffers from both invariance and reachability failures (due to diverging flows and spurious attractors). Figure \ref{fig:tli_rollout} (right) shows that all flows are bounded within the mode for both DS and BC after two cuts. In the case of DS, flows originally leaving the mode are now redirected to the attractor by the cuts; in the case of BC, while no flows leave the mode after modulation, spurious attractors are created, leading to reachability failures. This is a counterfactual illustration of Thm. \ref{eq:tli_kc1}, that policies without a stability guarantee are not G.A.S. after modulation. Figure \ref{fig:tli_result} verifies this claim quantitatively and we empirically demonstrate that a stable policy requires only four modulation cuts to achieve a perfect success rate---which an unstable policy cannot be modulated to achieve.

\begin{figure}[!tbp]
    \includegraphics[width=1\linewidth]{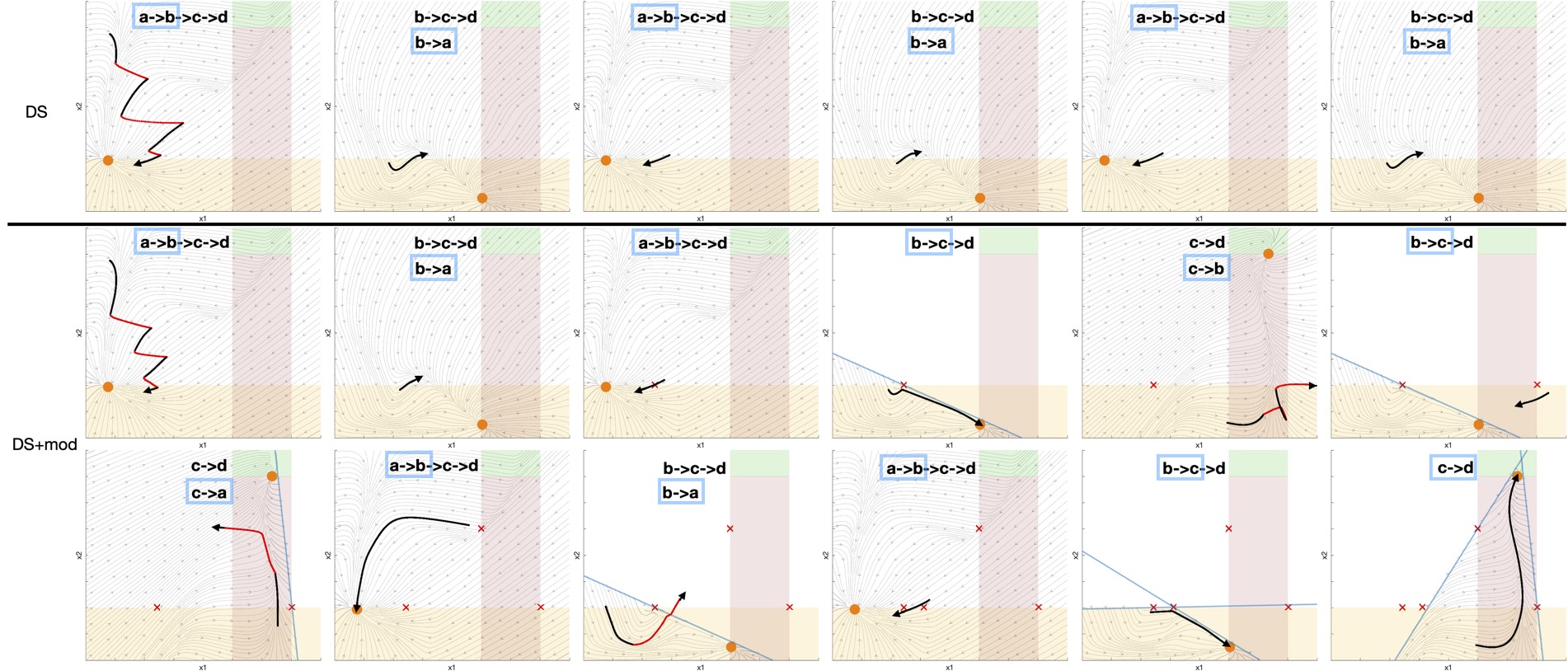}
\caption{Rollouts of a multi-step scooping task under perturbations. The first row shows that sequencing DS policies with an automaton can lead to looping without boundary estimation. The second and third rows show that modulation prevents looping and enables the system to eventually reach the goal mode despite repeated perturbations. We show the desired discrete plan at the top of each sub-figure and annotate the current mode transition detected in the blue box. Black and red trajectories signify original and perturbed rollouts. \label{fig:tli_modulation} 
}
\end{figure}

\subsection{Multi-Modal Reactivity and Generalization to New Tasks}\label{sec:tli_scooping_soup_task}
We now empirically demonstrate that having a reactive discrete plan alone is insufficient to guarantee task success without mode invariance for tasks with multiple modes. Consider the multi-modal soup-scooping task introduced in Fig. \ref{fig:tli_intro}. Formally, we define three environment APs, $r, s, t$, sensing the spoon is in contact with the soup, has soup on it, and has arrived at a target location respectively. Given successful demonstrations, sensors will record discrete transitions $(\lnot r\land \lnot s \land \lnot t) \Rightarrow (r\land \lnot s \land \lnot t) \Rightarrow (\lnot r \land s \land \lnot t) \Rightarrow (\lnot r \land \lnot s \land t)$, from which four unique sensor states are identified. We label each sensor state as a mode with robot AP $a \text{ (reaching) }\Rightarrow b \text{ (scooping) } \Rightarrow c \text{ (transporting) } \Rightarrow d \text{ (done)}$. The Invariance of mode $b$ enforces contact with soup during scooping, and the invariance of mode $c$ constrains the spoon's orientation to avoid spilling. We follow the TLP convention to assume LTL formulas are provided by domain experts (although they can also be learned from demonstrations \cite{shah2018bayesian, chou2021learning}.) The specific LTL for the soup-scooping task is detailed in Appendix \ref{sec:tli_multi-modal}, and can be converted into a task automaton as shown in Fig. \ref{fig:tli_method}. One might assume the automaton is sufficient to guarantee task success without modulation, as it only needs to replan a finite number of times assuming a finite number of perturbations; however, not enforcing mode invariance can lead to looping at the discrete level, and ultimately renders the goal unreachable, as depicted in the top row of Fig. \ref{fig:tli_modulation}. In contrast, looping is prevented when modulation is enabled, as the system experiences each invariance failure only once. 


\textbf{Robot Experiments.} First, we implement the soup-scooping task on a Franka Emika robot arm as shown in Fig.~\ref{fig:tli_robot_intro}. We show in videos on our website that (1) DS allows our system to compliantly react to motion-level perturbations while ensuring system stability; (2) LTL allows our system to replan in order to recover from task-level perturbations; and (3) our modulation ensures the robot learns from previous invariance failures to avoid repeating them. To test robustness against unbiased perturbations, we collect 30 trials from 6 humans as seen in Appendix \ref{sec:tli_robot_scoop}. All trials succeed eventually in videos. We do not cherry-pick these results, and the empirical 100\% success rate further corroborates our theoretic success guarantee. Second, we implement an inspection task as a permanent interactive exhibition at \href{https://yanweiw.github.io/tli/#Museum}{MIT Museum}, with details documented in Appendix \ref{sec:tli_inspection}. Lastly, we show a color tracing task testing different automaton structures with details in Appendix \ref{sec:tli_color}.

\textbf{Generalization.} LTL-DS can generalize to a new task by reusing learned DS if the new LTL shares the same set of modes. Consider another multi-step task of adding chicken and broccoli to a pot. Different humans might give demonstrations with different modal structures (e.g., adding chicken first vs adding broccoli first). LTL-DS can be reformulated to learn a policy for each mode transition (each mode can now have multiple policies), resulting in a collection of DS skills that can be flexibly recombined to solve new tasks. To generate different task LTLs, a human only needs to edit the $\phi_t^s$ portion of the original LTL formula. We provide further details of this analysis in Appendix \ref{sec:tli_generalization}.

\section{Related Work} 
\label{sec:tli_related}
\textbf{Temporal Logic Motion Planning.} LTL is a task specification language widely used in robot motion planning \cite{belta2007symbolic, wolff2013automaton, plaku2016motion, kress2018synthesis}. Its ease of use and efficient conversion \cite{piterman2006synthesis} to an automaton have spurred substantial research into TLP \cite{fainekos2005temporal, kress2009temporal, decastro2015synthesis}, which studies how to plan a continuous trajectory that satisfies an LTL formula. However, TLP typically assumes known workspace partitioning and boundaries \textit{a priori}, both of which are unknown in the rarely explored TLI setting. While a robot can still plan in uncertain environments \cite{ayala2013temporal, lahijanian2016iterative}, LfD bypasses the expensive search in high-dimensional space. Recent works \cite{innes2020elaborating, puranic2021learning} have considered temporal logic formulas as side-information to demonstrations, but these formulas are treated as additional loss terms or rewards and are not guaranteed to be satisfied. The key motivation for using LTL is to generate a reactive discrete plan, which can also be achieved by a finite state machine \cite{niekum2013incremental} or behavior tree \cite{li2021reactive}.

\textbf{Behavior Cloning.} We consider a subclass of LfD methods called state-based behavior cloning (BC) that learns the state-action distribution observed during demonstrations \cite{osa2018algorithmic}. DAGGER \cite{ross2011reduction}, a BC-variant fixing covariate shift, could reduce the invariance failures depicted in Fig. \ref{fig:tli_intro}, but requires online data collection, which our framework avoids with an LTL specification. To satisfy goal reachability, we employ a DS-based LfD technique \cite{khansari2011learning}. Alternatives to this choice include certified NN-based methods \cite{neumann2013neural, dawson2022safe}, DMPs \cite{ijspeert2013dynamical}, partially contracting DS \cite{ravichandar2017learning}, and Euclideanizing-flows \cite{rana2020learning}. To satisfy mode invariance, we modulate the learned DS to avoid invariance failure as state-space boundaries \cite{khansari2012dynamical}, similar to how barrier functions are learned to bound a controller \cite{robey2020learning, saveriano2019learning, dawson2022survey}.
For comparison of stable LfD variants, please refer to \cite{saveriano2021dmp, DSbook, dawson2022survey}.

\textbf{Multi-Step Manipulation.} Prior LfD works \cite{konidaris2012robot, niekum2013incremental, kroemer2015towards, ye2017guided} tackle multi-step manipulation by segmenting demonstrations via a hidden Markov model. Using segmented motion trajectories, \cite{konidaris2012robot} learned a skill tree, \cite{niekum2013incremental} learned DMPs, \cite{kroemer2015towards} learned phase transitions, and \cite{bowen2014closed} learned a task model. Most of these works assume a linear sequence of prehensile subtasks (pick-and-place) without considering how to replan when unexpected mode transitions happen. \cite{ye2017guided, bowen2014closed} considered a non-prehensile scooping task similar to ours, but their reactivity only concerned collision avoidance in a single mode. \cite{rajeswaran2017learning, gupta2019relay} improved BC policies with RL, but offered no guarantee of task success.




\section{Conclusion}
\label{sec:tli_conclusion}
In this paper, we formally introduce the problem of \textit{temporal logic imitation} as imitating continuous motions that satisfy an LTL specification. We identify the fact that learned policies do not necessarily satisfy the bisimulation criteria as the main challenge of applying LfD methods to multi-step tasks. To address this issue, we propose a DS-based approach that can iteratively estimate mode boundaries to ensure invariance and reachability. Combining the task-level reactivity of LTL and the motion-level reactivity of DS, we arrive at an imitation learning system able to robustly perform various multi-step tasks under arbitrary perturbations given only a small number of demonstrations. We demonstrate our system's practicality on a real Franka robot. 

\textbf{Limitations.} TLI assumes the existence of suitable mode abstractions, reactive logic formulas and perfect sensors to detect mode transitions, which can be difficult to obtain without non-trivial domain knowledge. Our work is based on the assumption that for well-defined tasks (e.g., assembly tasks in factory settings), domain expertise in the form of a logic formula is a cheaper knowledge source than collecting hundreds of motion trajectories to avoid covariate shift (we use up to $3$ demonstrations in all experiments). Moreover, even when abstractions for a task are given by an oracle, an LfD method without either the invariance or the reachability property will not have a formal guarantee of successful task replay, which is this work's focus. In future work, we will learn mode abstractions directly from sensor streams such as videos so that our approach gains more autonomy without losing reactivity.

\chapter{Learning Grounding Classifiers for Task and Motion Imitation}

\begin{tcolorbox}
\textsc{$\textbf{Grounding Language Plans in Demonstrations Through} \\
\textbf{Counterfactual Perturbations}$ \\
\footnotesize Yanwei Wang, Tsun-Hsuan Wang, Jiayuan Mao, Michael Hagenow, Julie Shah \\ 
\textbf{ICLR 2024 Spotlight}}
\end{tcolorbox}

\begin{flushright}
\vspace{1cm} 
\textit{"It's only a point." \\
— Roger Federer}
\end{flushright}

\section{Introduction}

Language models, in particular, pretrained large language models (LLMs) contain a large amount of knowledge about physical interactions in an abstract space. However, a grand open challenge lies in extracting such semantic knowledge and grounding it in physical domains to solve multi-step tasks with embodied agents. Previous methods, given the symbolic and abstract nature of language, primarily focus on leveraging LLMs to propose abstract actions or policies in purely symbolic spaces or on top of manually defined high-level primitive abstractions \cite{liu2023llmp,ahn2022can,wang2023voyager}. Such approaches inherently require a set of predefined primitive skills and additional toolkits for estimating affordances or feasibility before executing a plan generated by an LLM \cite{ahn2022can,lin2023text2motion}.

To address this important limitation, in this paper, we consider the problem of grounding plans in abstract language spaces into robot demonstration trajectories, which lie in the low-level robot configuration spaces. Our key idea is that many verbs; such as reach, grasp, and transport; are all grounded on top of mode families that are lower-dimensional manifolds in the configuration space~\cite[as in manipulation mechanics, see][]{mason2001mechanics, hauser2010multi}. Therefore, LLMs can be prompted to describe the multi-step structure of demonstrations in terms of semantic mode abstractions: valid mode transitions describe pre-conditions for mode-based skills, and mode boundaries explicitly encode motion constraints in the physical space that are critical for task success.

\begin{figure}[tp]
    \centering
    \includegraphics[width=\linewidth]{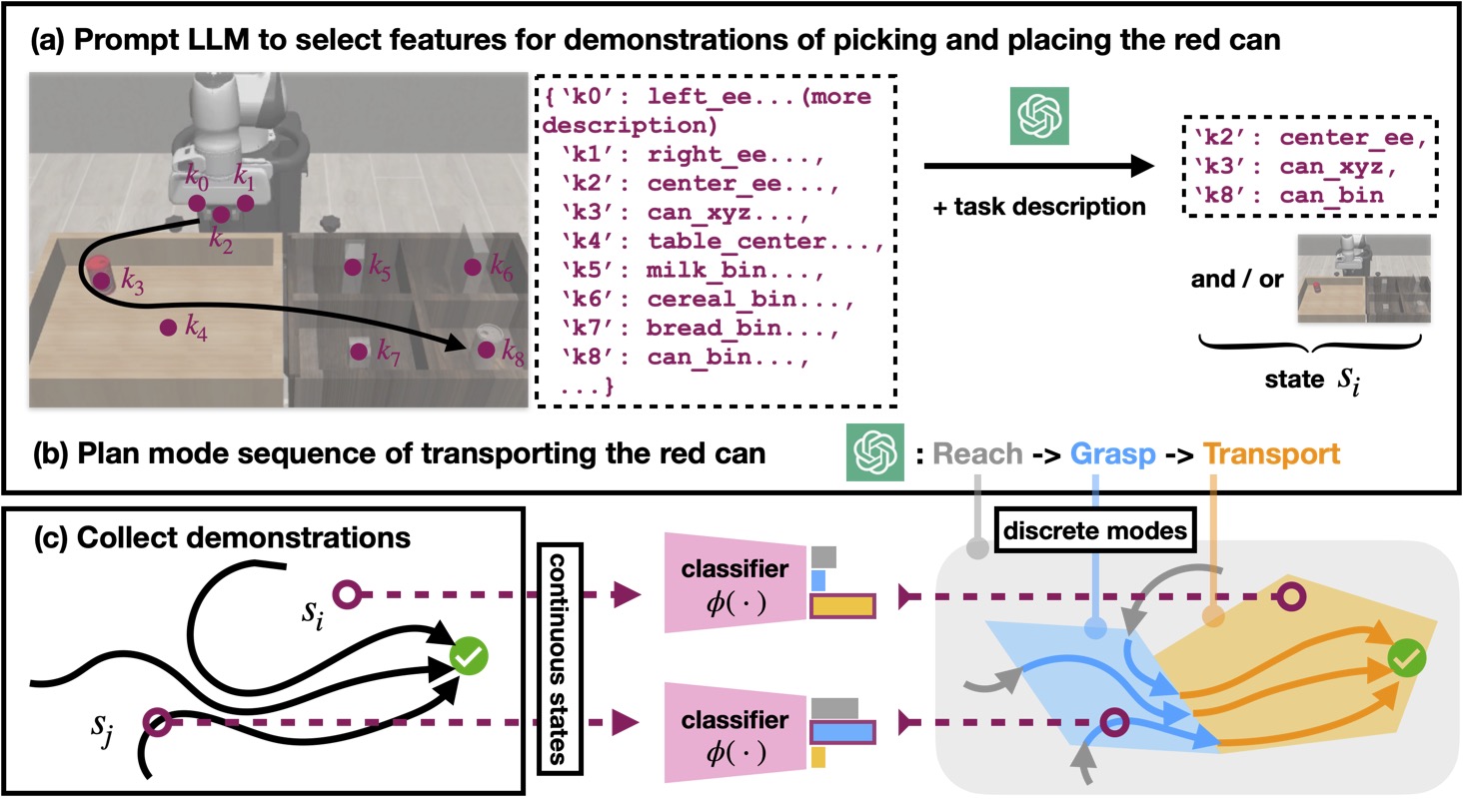}
    \caption{\textbf{\model framework.} Given a common-sense LLM that understands (a) the appropriate state abstractions for a task and (b) how to solve the task via a sequence of manipulation modes in semantic space and (c) a few unsegmented human demonstrations that embody the transitions through these modes, we learn a grounding classifier that maps continuous physical states and observations to discrete semantic modes. Mode boundaries discovered by the classifier encode constraints implicit in the demonstrations that are critical for task success.}
    \label{fig:glide_framework}
  \end{figure}

Building upon this idea, we propose {\it \modelfull} (\model, illustrated in \fig{fig:glide_framework}), which casts the language grounding problem into two stages: learning to classify current modes from states, and learning mode-specific policies. The main challenge in mode classification is that learning a decision boundary fundamentally requires both positive and negative labeled examples. To avoid having humans exhaustively provide dense mode annotation that covers the entire state space, we propose to systematically perturb demonstrations to generate ``counterfactual'' trajectories and use a simple ``overall'' task success predictor as sparse supervision. Intuitively, perturbations to inconsequential parts of a successful replay add unseen state coverage, while perturbations that cause counterfactual failing outcomes reveal constraints in the demonstration.
Next, we use an explanation-based learning paradigm \cite{dejong1986explanation,segre1985explanation} to recover the mode families that successful demonstrations implicitly transition through. With a learned classifier that maps continuous physical states to discrete abstract modes, we can then learn mode-specific policies and also use LLMs to plan for recovery from external perturbations or other sources of partial failures. Our system improves both the interpretability and reactivity of robot learning of multi-step tasks.

Our framework of grounding language plans as recovering modes and learning mode-specific policies brings two important advantages. First, compared to frameworks that generate robot behavior solely based on text, we do not require pre-built policies and feasibility predictors for primitive actions. Experiments show that our learning paradigm can successfully identify each mode from the demonstration data without any human segmentation annotations, and from only a small number of expert-generated demonstrations.
Second, connecting demonstrations with language suggests a principled way to improve the interpretability and reactivity of motion imitation. While plenty of data collection systems \cite{zhao2023learning,fang2023low, wu2023gello, fu2024mobile, chi2024universal} allow humans to demonstrate complex multi-step tasks, these demonstrations are typically unsegmented without semantic annotations of individual steps. Neither do humans elaborate on the task constraints that successful trajectories implicitly satisfy. Consequently, the resulting imitation policies cannot detect whether current actions fail to achieve pre-conditions \cite{garrett2021integrated} of subsequent actions or replan to recover from mistakes due to covariate shift \cite{ross2011reduction}. Our system enables the usage of LLMs for replanning and improves the overall system robustness.

\section{Method}
\label{sec:glide_method}

Our framework, GLiDE, takes in a language description of the target task, and a small set of successful human demonstrations as input, and aims to produce a robust policy that can accomplish the task successfully even under perturbations.
GLiDE first uses a perturbation strategy to augment a small set of human demonstrations with additional successful executions and failing counterfactuals. (\sect{ssec:glide_data}). Next, it prompts a large language model (LLM) to decompose the very high-level instruction into a step-by-step abstract plan in language. At this step, the most important outcome is a feasibility matrix that encodes how we can transition between different modes in this task (\sect{ssec:glide_llm}). Given the augmented demonstration and perturbation dataset and the LLM-generated abstract plan, we ground each mode onto trajectories (\sect{ssec:glide_algorithm}) and generate motions for individual modes to be sequenced by a language plan (\sect{ssec:glide_policy}).

\begin{figure}[t]
    \centering
    \includegraphics[width=1.0\linewidth]{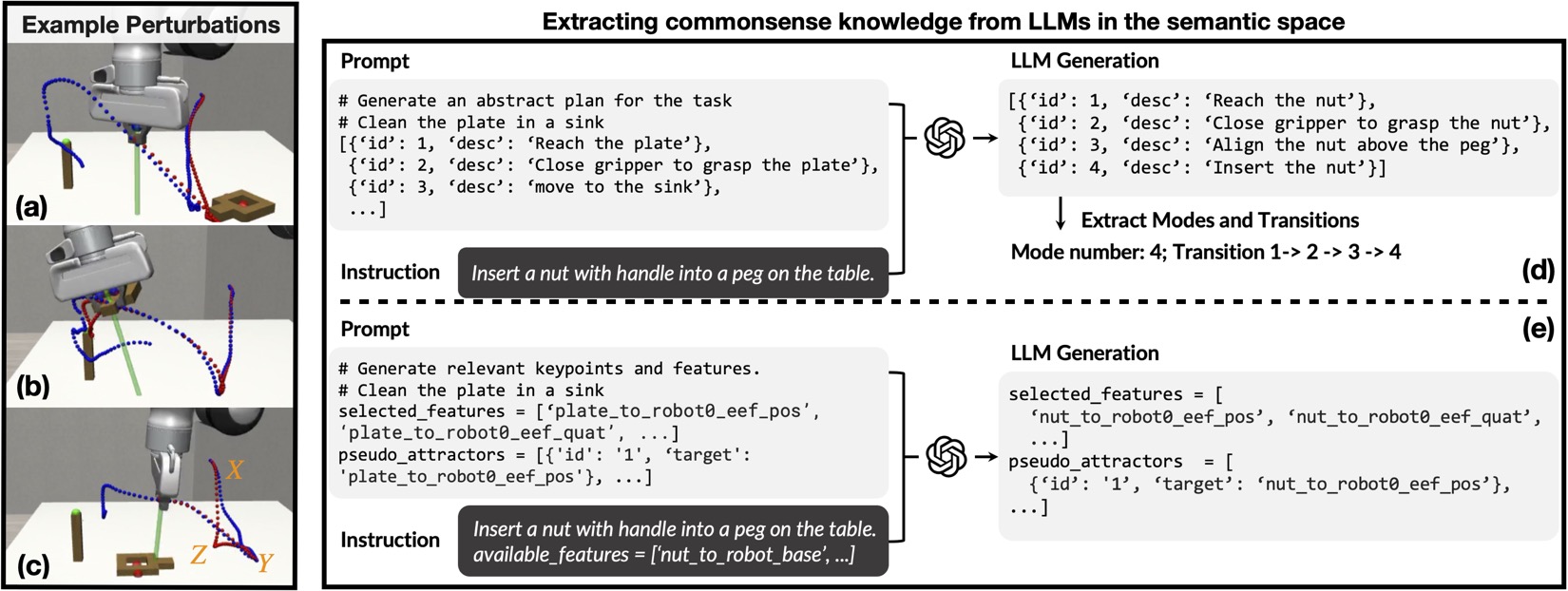}
    \caption{(a-c) Example perturbations causing replays (red) to deviate from successful demonstrations (blue). The task is to pick up the square nut and place it on the peg. End-effector perturbations at different locations (a) may or (b) may not cause grasp failures. (c) The gripper picks up the nut despite an initial end-effector perturbation but later drops it due to a gripper perturbation. LLMs can be prompted (d) to describe a task solution via a discrete mode sequence or (e) to select relevant features and pseudo attractors for solving a task.}
    \label{fig:glide_demo_llm}
  \end{figure}

\subsection{Demonstration Data Augmentation with Counterfactual Perturbations}
\label{ssec:glide_data}

To learn a grounding classifier that can partition the state space being considered into mode families, we need data coverage beyond the regions explored in a few successful demonstrations. Additionally, to learn mode abstractions that can be used to predict task success--as opposed to clustering data based on statistical similarity--negative data that fail by crossing infeasible boundaries are necessary. Assuming an oracle that can label the execution outcome of a synthetically generated trajectory, we propose the following perturbations to demonstration replays that might reveal task constraints:

\xhdr{End-effector perturbations} Illustrated in \fig{fig:glide_demo_llm}c, given a successful demonstration shown in blue, we first sample two points on the trajectory, namely $X$ and $Y$. Next, we randomly sample a third point $Z$ in the state space. During the replay shown in red, we replace the $XY$ segment with $XZ$ and $ZY$. Depending on the location and magnitude of the perturbations, the robot may still succeed in the task (\fig{fig:glide_demo_llm}b) or fail (\fig{fig:glide_demo_llm}a), revealing that grasping the square nut is a pre-condition for the next step of peg insertion to be successful.

\xhdr{Gripper perturbations} Illustrated in \fig{fig:glide_demo_llm}c, we randomly toggle the gripper state while otherwise adhering to the original trajectory. Failure replays where the gripper drops the nut pre-maturely reveal the motion constraint of holding the nut during transportation.

Given the perturbed trajectories, we execute them using a trajectory-following controller in the environment and collect a binary task success signal for each trajectory. Essentially, this gives us a dataset of paired trajectories and their task success labels: $\langle\tau^i, \textit{succ}^i\rangle$, where $\tau^i = \{s_1, s_2, \cdots, s_T\}$, and $\textit{succ}^i \in \{0, 1\}$. To learn the grounding classifier $\phi(\cdot)$ that can map $\tau^i$ to its corresponding mode sequence $\phi(\tau^i)=\{m_1, m_2, \cdots, m_T\}$ (mode and mode families are used interchangeably in this work), we ask LLMs what modes there are in a demonstration, how they are connected, and what constitutes a state $s_t$ for a given task.

\subsection{Semantic description of demonstrations and task structure from LLMs}
\label{ssec:glide_llm}

\xhdr{Explaining continuous demonstrations with a discrete mode sequence.} First, we assume a given small set of demonstrations $\{\langle\xi^i, 1\rangle\}_{i=1}^D$, which can be variable at the motion level, satisfy the same sequential transition through $K$ modes, defined as $\sigma \in \Sigma = \{\sigma_i\}_{i=1}^{K}$ and $\phi(s_t) = m_t \in \Sigma$. In other words, if we reduce self-transitions in the demonstrations where $m_t = m_{t+1}$, mode sequence $\phi(\xi^i)$ for all demonstrations can be reduced to the same $K$-step transitions $\sigma_1 \rightarrow \sigma_2 \rightarrow \cdots \sigma_K$. This is the form of the language plan we prompt LLMs to generate to describe demonstrations. The plan informs the number of modes there are as well as the semantic grounding of each mode as seen in \fig{fig:glide_demo_llm}d. 

\textbf{Representing states with task-informed abstraction.} Second, we further prompt LLMs to define the state representation $s_t$ as a set of keypoint-based features or image observations that are relevant to mode classification. In particular, the keypoint-based features come from a pre-defined exhaustive list of keypoints describing the scene as seen in \fig{fig:glide_framework}a. Each keypoint definition contains (1) the keypoint name and (2) a short description of its semantic meaning. Given a task description, an LLM can be prompted to either select a subset of keypoints tracking absolute locations or combine pairs of keypoints to track relative positions as shown in \fig{fig:glide_demo_llm}e. For image observations, we either use the raw image as a state representation or use a pre-trained vision model \cite{kirillov2023segment} or vision-language model \cite{huang2023voxposer} to extract features from the image.

\begin{figure}
    \centering\small
    \includegraphics[width=\textwidth]{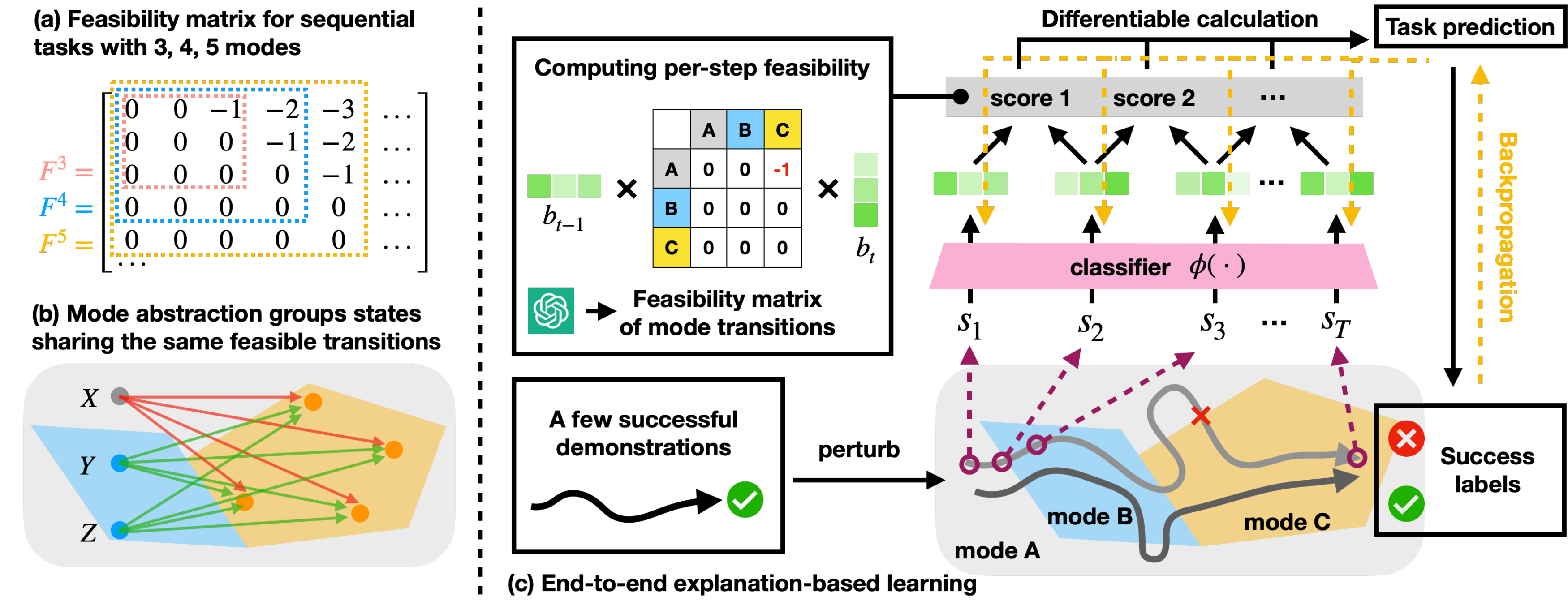}
    \caption{(a) Example feasibility matrices. Specifically, $F^3$ can describe the modal structure for a pick-and-place task with solution reach$\rightarrow$grasp$\rightarrow$transport, where reach$\rightarrow$transport directly is infeasible. (b) The definition of a mode transition implies every state in the second mode is reachable from every state in the first mode (states Y and Z are in the same mode but not X). We leverage this connection between the continuous states and the discrete modes to design (c) a fully-differentiable pipeline that calculates overall trajectory success based on the mode classification of individual states in the trajectory.}
    \label{fig:glide_algo_detail}
\end{figure}

\textbf{Encoding discrete modal structure in a feasibility matrix.} Lastly, while successful demonstrations $\xi^i$ can be reduced to a K-step language plan, not every perturbed trajectory $\tau^i$ can be as it might not be successful or correspond to a minimal solution.  Therefore, the reduced mode sequence may contain back-and-forth steps such as $\sigma_1 \rightarrow \sigma_2 \rightarrow \sigma_1 \rightarrow \cdots$ or simply invalid mode transitions. To describe the modal structure of a task in terms of the feasible transitions between modes, we generate a feasibility matrix $F^K$ with $K$ modes by first querying LLMs whether two semantic modes are directly connected. Then we compute the matrix entry $F_{ij}$ from LLM responses as the negative shortest path between each pair of modes. 
In the case of sequential tasks with a linear temporal structure (true for most experiments considered in this work), zero entries $F_{ij}$ encode valid transitions that incur zero costs. Negative entries $F_{ij}$ encode infeasible transitions, and the magnitudes denote the number of missing modes in between. In particular, in \fig{fig:glide_algo_detail}a diagonal entries $F_{ii}$ are feasible self-transitions, and entries $F_{i,i+1}$ are demonstrated mode transitions towards the goal.
Note for tasks with complex structures, the matrix may have more negative entries than the ones shown in \fig{fig:glide_algo_detail}a. The feasibility matrix is also interpretable and can be modified manually by humans. 

\subsection{End-to-end explanation-based learning for mode classification}
\label{ssec:glide_algorithm}

Given a language plan, a task-informed state representation, and a feasibility matrix as discrete structural information about a task, learning the grounding classifier is an inverse problem that tries to recover the underlying modal structure from sparsely labeled continuous trajectories. To this end, we design a differentiable decision-making pipeline to explain the task success of a trajectory on top of mode predictions. Having trajectory coverage with contrasting execution outcomes allows for recovering the precise grounding in terms of mode boundaries. 

\textbf{Mode classifier.} Our mode classifier is a neural network (with softmax output layers) that inputs a state $s_t$ and outputs a categorical distribution of the abstract mode at that state. Overloading the notation $\phi(\cdot)$ to output both a predicted mode $m_t$ and a mode belief, we have $b_t = \phi(s_t)$.
The architecture of the classifier depends on the state representation.
The number of softmax categories $K$ is chosen based on the sequence length of the LLM-generated plan. If we had dense mode annotations $\{\langle s_t, m_t\rangle\}_{t=1}^T$, we could train the classifier directly with a cross-entropy loss. However, we only have supervision at the trajectory level via task success. Therefore, we need a differentiable forward model that can predict task success from a sequence of mode beliefs $\{b_t\}_{t=1}^T$.

\textbf{Differentiable forward model to predict task success.} What makes a perturbed trajectory rollout unsuccessful (or still successful) in solving a task? Following the approach by \cite{wang2022temporal}, we consider a successful trajectory as one that both $(1)$ contains only feasible mode transitions according to $F^K$ and $(2)$ eventually reaches the final mode $\sigma_K$ seen in the demonstrations.
Since the perturbations we consider in this work do not affect the starting state $s_1$ and final state $s_T$, the success criteria for a trajectory $\tau^+$ solely concerns intermediate transitions: $\phi(s_t^+)F^K\phi(s_{t+1}^+)=0\textrm{, } \forall s_t^+ \in \tau^+$.
Similarly, a failure trajectory $\tau^-$ is one that contains at least one invalid mode transition.

To operationalize this idea, let's consider the dataset of trajectories $\mathcal{T}$ containing both successful trajectories $\mathcal{T}^+=\{\tau^{i+}\}_{i=1}^M$ and failure trajectories $\mathcal{T}^-=\{\tau^{j-}\}_{j=1}^N$.
First, we use a cross-entropy loss to enforce that the starting and ending continuous states for all trajectories must be in the initial and final mode being demonstrated: $\gL_{\textit{init}} = \mathbb{E}_{\tau_i\sim \mathcal{T}}\gL_{\textit{CE}}(\phi(s_1^i), \sigma_1)$ and $\gL_{\textit{final}} =\mathbb{E}_{\tau_i\sim \mathcal{T}}\gL_{\textit{CE}}(\phi(s_T^i), \sigma_K)$. Second, we define the success and failure loss using  $f_{t,t+1}$, which is a shorthand for transition feasibility score $\phi(s_t)F^K\phi(s_{t+1})$ between two states:
\begin{equation}
    \gL_{\textit{succ}} = -\frac{1}{M}\sum_{\tau_i\in\mathcal{T}^+} \frac{1}{T-1}\sum_{t=1}^{T-1}f_{t,t+1} \quad\quad 
    \gL_{\textit{fail}} = \frac{1}{N}\sum_{\tau_j\in\mathcal{T}^-} \max(-1, \sum_{t=1}^{T-1}f_{t,t+1}) \\
\end{equation}
Intuitively, minimizing $\gL_{\textit{succ}}$ encourages the classifier to predict mode beliefs such that all transitions between consecutive states are feasible. Minimizing $\gL_{\textit{fail}}$ encourages the classifier to predict mode beliefs such that there exists at least one invalid mode transition. The clipping in $\gL_{\textit{fail}}$ at $-1$ makes the loss well-defined and treats all invalid mode transitions described by the negative entries in \fig{fig:glide_algo_detail}a equally\footnote{Empirically, setting all negative entries in the matrix to be $-1$ can get gradient descent optimization stuck.}.

\fig{fig:glide_algo_detail}b gives another intuitive example, where states $Y$ and $Z$ constitute the same mode but not state $X$. A necessary condition to test if a state $s$ is in mode $\sigma_i$ is to check if $s$ can directly transition to at least one state in mode $\sigma_{i+1}$ in the trajectory. Adding everything together, we have $\gL_{\textit{full}}$ in Eq.~\ref{eq:glide_full_loss}, where $\lambda_s$, $\lambda_f$, and $\lambda_i$ are hyperparameters for balancing loss terms:
\begin{equation}
    \gL_{\textit{full}} = \lambda_s \gL_{\textit{succ}} + \lambda_f \gL_{\textit{fail}} + \lambda_i (\gL_{\textit{init}} + \gL_{\textit{final}})
    \label{eq:glide_full_loss}
\end{equation}

\xhdr{Extension to underactuated systems.}
These conditions are sufficient for recovering modes from a fully-actuated system. For underactuated systems \cite{underactuated} such as object manipulation where objects cannot directly move from one configuration to another via teleportation, it is not possible to generate a direct transition between any two modes such as the ones shown in \ref{fig:glide_algo_detail}b using synthetic perturbations. Hence, we need an additional regularization at the motion level to infer precise boundaries.
Specifically, states in the same mode should go through similar dynamics. In other words, one should be able to infer $(s_{t+1}-s_t)$ from $(s_t - s_{t-1})$. Such mapping should be different for different modes. For example, the relative transformation between the end-effector pose and the object pose should remain the same when the robot is rigidly holding the object and change otherwise. Based on this observation, we instantiate a forward dynamics model $\psi(\cdot)$ that inputs the current state change and predicts how the state should change next for each mode. Coupled with a mode belief, we can predict the next state change as $\phi(s_t)^\intercal\psi(s_{t}-s_{t-1})$. Consequently, we can train mode classifiers for underactuated systems by introducing a dynamics loss $\gL_{\textit{dyn}}$:

\begin{equation}
    \gL_{\textit{under}} = \gL_{\textit{full}} + \lambda_d \gL_{\textit{dyn}} \quad \text{where} \quad
    \gL_{\textit{dyn}} = \sum_{\tau_j\in\mathcal{T}} \sum_{t=1}^{T-1} \| \phi(s_t)^\intercal\psi(s_{t}-s_{t-1}) - (s_{t+1}-s_t) \|_2^2
    \label{eq:glide_under_loss}
\end{equation}

Minimizing $\gL_{\textit{dyn}}$ groups states into modes based on similarity in dynamics. Since losses are differentiable with respect to $\phi$ and $\psi$, we use stochastic gradient descent to optimize learnable parameters.

\subsection{Mode-Based Motion Generation}
\label{ssec:glide_policy}

Having learned the explicit mode boundaries, we can leverage them in motion planning to ensure that the robot avoids invalid mode transitions ~\cite{lavalle1998rapidly}. Alternatively, we can use the classifier $\phi(\cdot)$ to segment demonstrations into mode-specific datasets, with which we can learn imitation policies $\pi_k(a | s)$ for each mode $\sigma_k$ and sequence them using a discrete plan \cite{wang2022temporal}. To further improve the robustness of the learned policy for manipulation tasks, we use the mode feature identified by the LLM to construct a pseudo-attractor for each mode. If the mode feature is the absolute pose of the robot end-effector, we compute the mean end-effector poses at which mode transitions $\sigma_k \rightarrow \sigma_{k+1}$ occur as the pseudo-attractor; if it is a relative pose, we transform that into an absolute pose of the end-effector at test time. We use this pseudo-attractor to construct a potential field that guides the robot to move towards the next mode at inference time. Specifically, the final mode-based policy $\pi^*_k(a | s)$ is a weighted sum of the original $\pi_k(a | s)$ and a control command that moves the end-effector towards the pseudo-attractor for mode $\sigma_k$. We only apply the pseudo-attractor term when the distance between the current state and the pseudo-attractor is greater than a threshold. Intuitively, when a large perturbation leads to out-of-distribution states, the potential field will drive the system back to the demonstration distribution before the imitation policy $\pi_k$ takes sole effect.

\begin{figure}[tp]
  \centering\small
  \includegraphics[width=\textwidth]{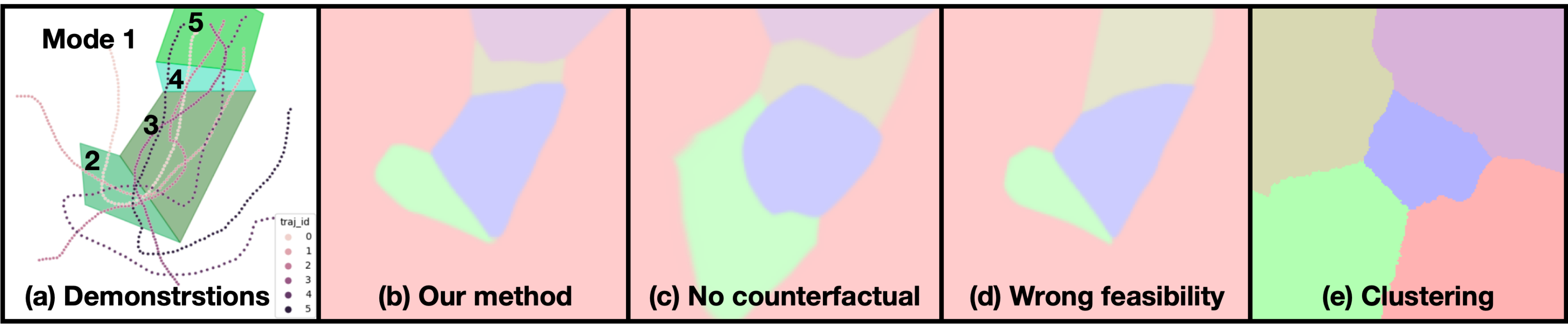}
  \caption{Grounding of 2D navigation task. (a) Given six demonstrations that start in mode 1 and end in mode 5, visualized on top of the ground truth, (b) our method \model recovers the underlying mode abstractions. (c) Without counterfactual data, \model fails to learn precise boundaries. (d) Without a correct feasibility matrix (\eg 4-mode instead of 5-mode), \model results will miss modes. (e) Lastly, clustering the 2D state space to the nearest mode centers, discovered in the demonstrations by kmeans++, produces an incorrect modal structure.}
\label{fig:glide_polygon-visualization}
\end{figure}

\begin{table}[tp]
\centering
  \setlength\tabcolsep{4pt}
  \begin{tabular}{p{4.5cm}p{3.7cm}p{3.7cm}p{3.4cm}}
    \toprule
    Method & 3-Mode (+perturb) & 4-Mode (+perturb) & 5-Mode (+perturb) \\    
    \midrule
    Behavior Cloning (BC) & 0.967 \quad (0.908) & 0.814 \quad (0.614)  & 0.810 \quad (0.596)\\
    \model+BC             & 0.963 \quad (0.887) & 0.892 \quad  (0.753) & 0.893 \quad (0.753) \\
    \model+Planning       & \textbf{0.996 \hspace{0.2em} (0.996)} & \textbf{0.987 \hspace{0.2em} (0.966)} & \textbf{0.991 \hspace{0.2em} (0.974)} \\
    \bottomrule
  \end{tabular}
  \caption{2D navigation success rates for mode-agnostic imitation (BC) and mode-based (\model) imitation or planning for environments consisting of 3, 4, and 5 modes. 
  We test both perturbed and non-perturbed settings 1000 times and report the average success rate. For neural network-based BC policies, we report the average performance 
  across 10 different random seeds.}
  \label{table:glide_polygon}
\end{table}

\section{Experiments}

We evaluate our method on three sets of experiments: (1) a 2D navigation task, (2) simulated robot manipulation tasks in RoboSuite~\cite{zhu2020robosuite}, and (3) a real-robot implementation of the 2D navigation and a marble-scooping task. Since our robot experiments use end-effector pose control, we refer to task-space features as states rather than configurations.

\subsection{2D Navigation}

\xhdr{Setup.} The 2D navigation environment consists of a sequence of connected randomly generated polygons, and the goal is to traverse from any state in the free space (mode 1) through the polygon sequence consecutively as demonstrated until reaching the final polygon. This environment serves as a 2D abstraction of the modal structure for multi-step manipulation tasks, where each polygon represents a different mode with its boundaries showing the constraint of the mode. Illegal transitions include non-consecutive jumps between modes such as direct transitions from free space to any later modes other than mode 2. 

This system is fully-actuated with $(x, y)$ coordinate as the state representation and $(\dot{x}, \dot{y})$ as the agent action. For all environments, we use fewer than 10 successful demonstrations for classifier learning and policy learning. 

\textbf{Results: Mode classification.} We visualize the learned grounding classifier in Fig. \ref{fig:glide_polygon-visualization}b and Appendix~\ref{app:glide_polygon}. Compared to baselines in Fig. \ref{fig:glide_polygon-visualization}(c-e), the mode boundaries recovered by \model are the closest to the ground truth shown in Fig. \ref{fig:glide_polygon-visualization}a. In particular, the poor grounding learned in Fig. \ref{fig:glide_polygon-visualization}(c-e) shows respectively the importance of learning with counterfactual data, a correct task specification from LLMs, and a task prediction loss beyond clustering solely based on statistical similarities in the data. Quantitative results and more visualizations can be found on our website.  

\textbf{Results: Task execution.} Next, we show the learned grounding classifier can be used to improve task success rates, especially in the face of external perturbations. We use behavior cloning (BC) as a mode-agnostic baseline to learn a single policy $\pi(a | s)$ from all successful trajectories. By contrast, our method (\model+BC) first segments the demonstrations and then learns mode-specific policies. Additionally, instead of mode-based imitation, we could also do planning to stay in the mode boundaries recovered by the classifier, since the system is fully-actuated with a single-integrator dynamics. Specifically, (\model+Planning) uses RRT to compute waypoints to guide motion in non-convex mode 1 and then uses potential fields in convex polygons to generate trajectories that stay in the mode until entering the next mode. \tbl{table:glide_polygon} shows that our methods perform slightly better than BC across different environments. However, when external perturbations are introduced, BC suffers the biggest performance degradation as recovery at the motion level without attention to mode boundaries may incur invalid transitions leading to task failures. The fact that (\model+Planning) can almost maintain the same success rate despite perturbations validates the learned grounding. 

\textbf{Interpretability.} In the 2D environment, visualization of learned mode families can expose mode constraints and explain why some but not all perturbed demonstration replays fail the task execution.

\begin{table}[t]
  \centering
  \begin{minipage}{0.8\textwidth}
    \centering
    \setlength\tabcolsep{4pt}
    \begin{tabular}{lccc}
      \toprule
      Method & Can & Lift & Square \\    
      \midrule
      \model & \textbf{0.83} & \textbf{0.83} & \textbf{0.67} \\
      \model \,- Dynamics Loss &  0.67 & 0.75 & 0.46 \\
      \model \,- Prediction Loss & 0.67 & 0.68 & 0.56 \\
      \model \,- Feature Selection & 0.55 & 0.70 & 0.57 \\
      Traj. Seg. Baseline & 0.66 & 0.56 & 0.54 \\
      \bottomrule
    \end{tabular}
    \captionsetup{width=\textwidth}
    \caption{Ablation study of the influence of different loss terms and baselines on the mode classification accuracy based on overlap (percentage) with the ground truth.}
    \label{table:glide_manipablation}
  \end{minipage}
\end{table}

\begin{table}[t]
  \centering
  \begin{minipage}{0.8\textwidth}
    \centering
    \setlength\tabcolsep{4pt}
    \begin{tabular}{lccc}
      \toprule
      Method & Can & Lift & Square \\    
      \midrule
      BC & 0.93 & 0.99 & 0.38 \\
      BC (p) &  0.20 & 0.18 & 0.03 \\
      \model+BC & 0.85 & 0.99 & 0.25 \\
      \model+BC (p) & 0.40 & 0.39 & 0.15 \\
      \bottomrule
    \end{tabular}
    \captionsetup{width=\textwidth}
    \caption{The success rate of mode-agnostic imitation (BC) drops more than that of mode-conditioned imitation (\model+BC) after introducing perturbations (denoted by \textit{p}).}
    \label{table:glide_manipexp}
  \end{minipage}
\end{table}

\subsection{Robosuite}

\xhdr{Setup.} We test \model across three tasks from  Robosuite: placing a can in a bin (\textit{can}), lifting a block (\textit{lift}), and inserting a square nut into a peg (\textit{square}).
We use the default action and observation space of each environment unless an LLM suggests different features (\eg, relative distance to an object or keypoints). Since the manipulation tasks define underactuated systems, we use Eq. \ref{eq:glide_under_loss} for training.

\textbf{Results: Mode classification.} To evaluate the mode classification accuracy, we manually define the ground truth modes for each environment (details in Appendix~\ref{app:glide_mode-definition-robosuite}). \tbl{table:glide_manipablation} shows the percentage of overlap between mode predictions from different methods and the ground truth mode segmentation. The results show that including all of the loss terms in our method achieves the best boundary alignment with the ground truth as visualized in Appendix~\ref{app:glide_mode-definition-robosuite}. Ablating the dynamics loss or the task prediction loss ($\mathcal{L}_{succ}$ and $\mathcal{L}_{fail}$) degrades the prediction accuracy as the classifier misses the precise location of important events such as dropping a grasped object. Comparing \model to training without feature selection shows the importance of using an LLM to down-sample the feature space for efficient learning. A trajectory segmentation baseline using kmeans++ clustering on the features also underperforms \model, highlighting the limitation of similarity-based segmentation methods. 

\textbf{Results: Task execution.}  
To show the learned grounding can help recover from perturbations, we compare a mode-agnostic BC baseline, which is trained on unsegmented successful demonstrations, to a mode-conditioned method (\model+BC)  described in Section \ref{ssec:glide_policy}, where each per-model BC policy is augmented with a pseudo-attractor.  
While our method is insufficient to recover from all potential failures, our goal is to demonstrate how even a basic control strategy that leverages the underlying mode families can benefit policy learning in robotics. 
\tbl{table:glide_manipexp} summarizes the methods' performance without and with perturbations, which will randomly displace the end-effector or open the gripper. 
We see that for both methods, adding perturbations introduces some amount of performance drop. We find that the performance degradation for the BC baseline is much higher than with \model+BC.

\textbf{Interpretability.} In manipulation environments, it is challenging to directly visualize the mode families given the high-dimensional state space. However, exposing the mode families allows us to easily identify mode transition failures which can be used to generate post-hoc explanations of failures (\eg, videos on our website show invalid mode transitions associated with a task failure).

\subsection{Real Robot Experiments: 2D Navigation and Scooping Tasks}

\begin{figure}[!htb]
  \centering\small
  \includegraphics[width=\textwidth]{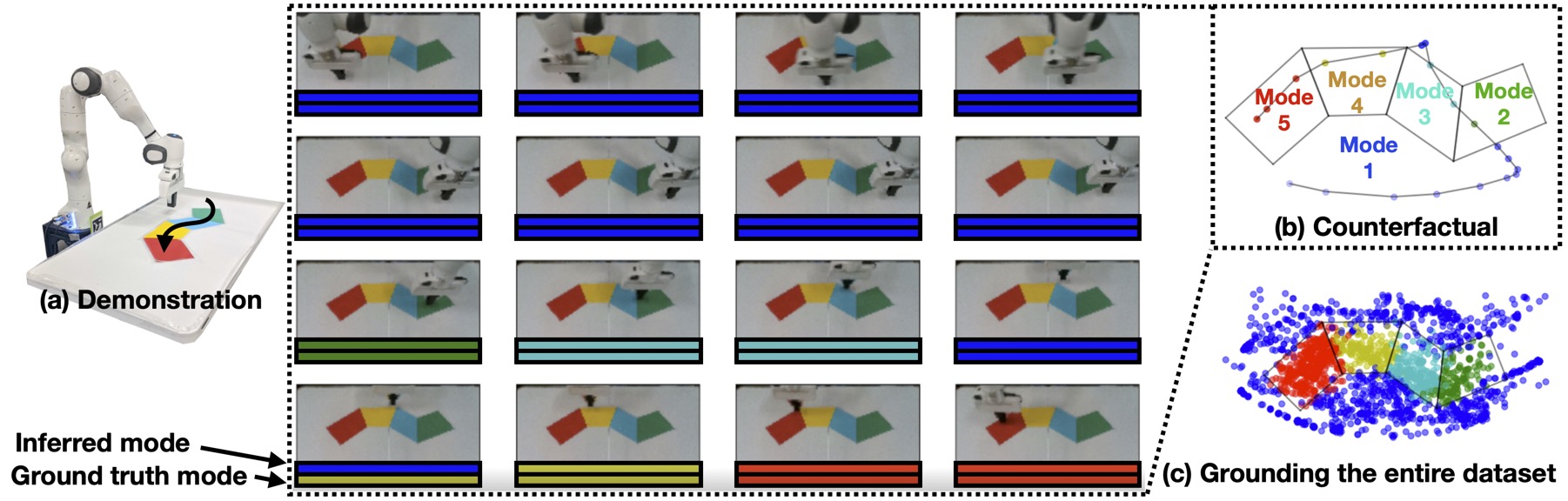}
  \caption{Illustration of the real robot 2D navigation task (a), where the end-effector traces through a sequence of colored polygons. 
  (b) shows a perturbed trajectory, overlaid on ground truth mode boundaries, experiences an invalid transition from mode 1 to mode 4. A vision-based classifier can predict from only pixels the inferred modes (first color bar) that match ground truth (second color bar) with high probability. (c) visualizes the mode prediction of individual image states seen in the dataset. The location of the scattered dots indicates where the images are recorded while the colors show the predictions, which are well-aligned with mode boundaries.}
  \label{fig:glide_realrobot2dnav}
\end{figure}

\xhdr{2D navigation.} To illustrate \model can also learn grounding classifiers directly from vision inputs, we implement the simulated 2D navigation task on a real Franka robot, where the end-effector traces through a sequence of colored polygons in a plane. First, we record 20 human demonstrations through kinesthetic teaching in the end-effector's state space that start in various parts of mode 1 and end in mode 5 as seen in \ref{fig:glide_realrobot2dnav}a. Second, we add end-effector perturbations to the demonstration replays to generate coverage over the entire tabletop area and record the perturbed trajectories in image sequences as seen in \ref{fig:glide_realrobot2dnav}b. For this planar task, we use a simple reset mechanism that brings the end-effector back to one of the demonstrations' starting states after each rollout to collect data continuously. Since we can check if the end-effect is within the convex hull of any colored regions whose vertices are known, we log the mode sequence of each perturbed trajectory and automatically label if the trajectory is a successful task execution by checking if all mode transitions are feasible. Consequently, we were able to collect 2000 labeled trajectories in 2 hours continuously without human supervision. To learn a vision-based classifier, we switch from using multi-layer perceptrons (MLP) for state-space inputs to convolutional neural networks (CNN) to encode image inputs. Figure \ref{fig:glide_realrobot2dnav}c shows the learned classifier can group image observations into correct modes according to the ground truth mode boundaries. 

\begin{figure}[!htb]
  \centering\small
  \includegraphics[width=\textwidth]{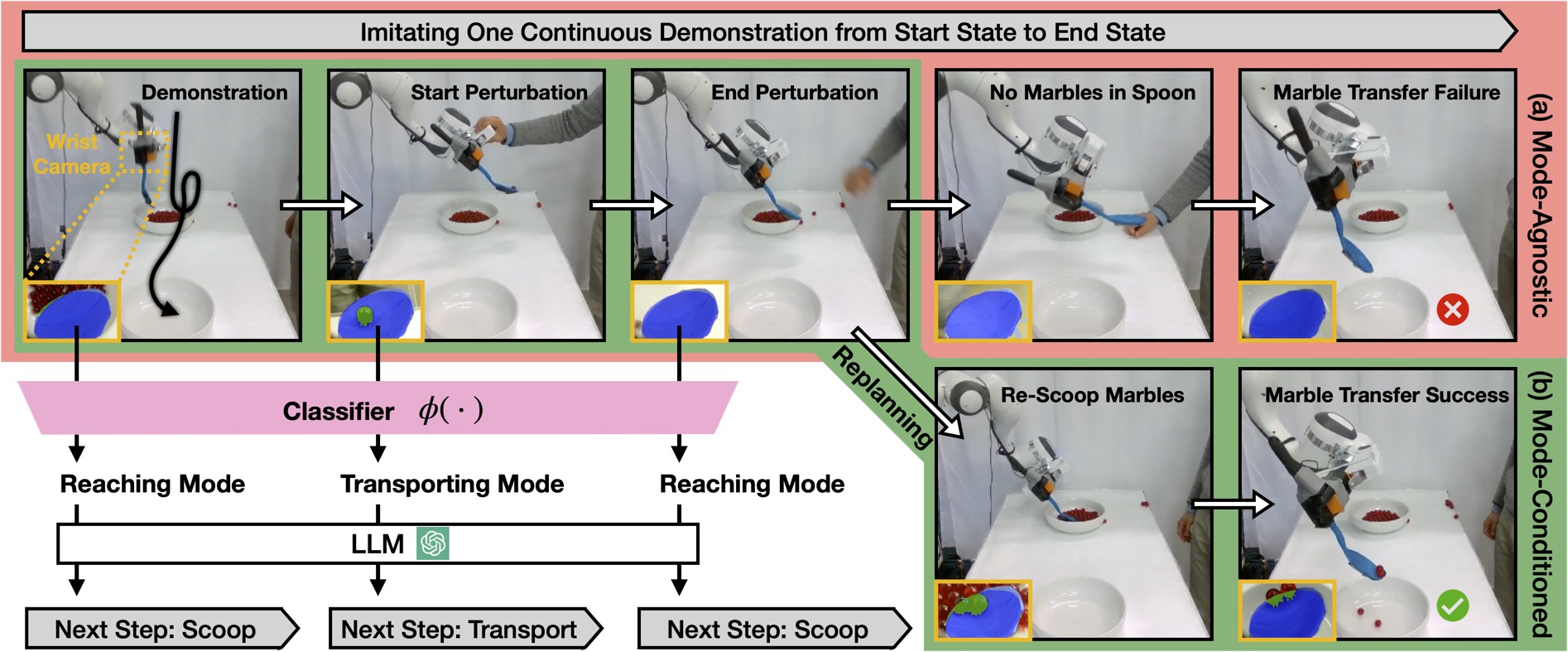}
  \caption{Illustration of the robot scooping task. Here the perturbations are human-initiated (e.g., moving the spoon to drop or fail to grasp marbles). A mode-agnostic BC that imitates continuous motion demonstrations cannot replan at the task level when all marbles are dropped during the transporting mode (a). In contrast, our mode-conditioned policy enabled by the grounding classifier can leverage LLMs to replan given external perturbations (b).}
  \label{fig:glide_realscooping}
\end{figure}

\textbf{Marble scooping task.} The second task requires a spoon-holding robot to scoop marbles from a bowl and then transport at least one marble to a second bowl across the table. A typical robot implementation might require engineering a marble detector to check if the spoon is holding marbles and plan actions accordingly \cite{wang2022temporal}. Instead, we learn a marble classifier on a wrist camera view to leverage LLM-based replanning as shown in Fig. \ref{fig:glide_realscooping}. 
To collect successful executions, we record the end-effector's pose and wrist camera view as a human demonstrates scooping from various starting states. Since it is non-trivial to engineer a reset mechanism for this task, we ask humans to demonstrate various failures through kinesthetic teaching as well. To improve learning efficiency, we preprocess the raw wrist image to extract a mask corresponding to marble objects, where an empty spoon returns an empty mask. Specifically, we prompt an LLM for relevant object types to track, with which we employ the Segment Anything Model (SAM) \cite{kirillov2023segment} to generate segmentation masks. The classifier is then trained on a state representation consisting of end-effector poses and marble masks (details in Appendix~\ref{app:glide_scooping}). To show the utility of the learned mode abstractions, we show a mode-agnostic policy cannot recover from task-level perturbations (dropping marbles) while a mode-based policy can leverage LLM to replan to ensure successful execution in Fig. \ref{fig:glide_realscooping}. Videos can be found on our website \url{https://yanweiw.github.io/glide/}. 

\section{Related Work}

\xhdr{Learning abstractions from demonstrations.}
A large body of work focuses on learning action abstractions from language and interaction. This includes the grounding of natural language \cite{corona-etal-2021-modular,andreas2017modular,andreas-klein-2015-alignment,jiang2019language,sharma2021skill,luo2023learning}, programs \cite{sun2020program}, and linear temporal logic (LTL) formulas~\cite{bradley2021learning,icarte-teaching-2018,tellex2011understanding}. In contrast to learning policies for individual action terms, this paper focuses on learning mode families in robot manipulation domains. These learned mode families enable us to construct robust policies under perturbation \cite{wang2022temporal}. Furthermore, our framework is capable of recovering the underlying modes from a small number of {\it unsegmented} demonstrations. Through the use of synthetic noise \cite{delaney2021instance, wang2022visual}, we relieve humans from the burden of providing negative demonstrations that fail a task by automatically generating positive and negative variations of task executions.

\textbf{Grounding language in robot behavior.}
With the rise of LLMs that can decompose high-level commands into sequences of actions, there has been much recent interest in the ability to ground these commands in embodied agents. Given that data from environment interactions (including human demonstrations) do not explicitly identify constraints and success criteria of a task, previous work has investigated how to infer affordances directly from observations \cite{ahn2022can}. Compared to prior work (e.g., \cite{lynch2023interactive}), our method does not require dense labels to learn a grounding operator. We are also not directly using large language models for planning \cite{huang2022language, li2022pre, huang2023voxposer}. Rather, we are using LLM to guide the discovery of mode abstractions in demonstrations, and as a result, we can also acquire a grounding operator for high-level language commands. In contrast to the discovery of language-conditioned skills \cite{lynch2020language, garg2022lisa}, which can consist of multiple modes, our mode decomposition occurs at a lower level and can explain why certain trajectories fail a task execution. 

\textbf{Counterfactuals.}
Counterfactuals describe hypothetical situations of alternative outcomes compared to the original data \cite{byrne2019counterfactuals}. In other words, they are fake (non-human generated) data with a different result (e.g., failing a task instead of succeeding) \cite{karimi2020survey}. In this paper, we define counterfactual perturbations as non-human-generated synthetic probes that test which parts of the time-series trajectory data \cite{delaney2021instance} demonstrated by humans have implicit constraints, the violation of which will change the outcome of the successful human demonstrations. 

\section{Conclusion}

In conclusion, this work introduces a framework, \textit{\modelfull} (\model), to effectively ground the knowledge within large language models into physical domains, via mode families. Given a small number of human demonstrations and task descriptions, we show how \model successfully recovers mode families and their transitions required in the task and enables the learning of robust robot control policies.

\xhdr{Limitations and future work.}
While GLiDE does not need a large number of human demonstrations, it requires a large number of trial-and-errors and an environment with a reset capability in order to collect task success labels of a trajectory. This data inefficiency, however, can be addressed through active learning where the current belief of mode segmentation can be used to probe demonstrations only in regions with high uncertainty. Additionally, prompting the LLM to find a suitable state representation for learning the classifier also requires skill. In future work, we would like to learn the state representation in conjunction with the mode classifiers in an end-to-end fashion.

\chapter{Conclusion}

\begin{flushright}
\textit{"We are all in the gutter, but some of us are looking at the stars." \\
— Oscar Wilde}
\end{flushright}

\section{Summary of Contributions}

This thesis presents ITPS as one framework for real-time policy steering, introduces TAMI as another framework for ensuring task success under physical steering, and proposes TLI and GLiDE as concrete instantiations addressing constraint recovery and mode abstraction within the TAMI framework.

With ITPS, we propose a novel inference-time framework that incorporates real-time user interactions to steer frozen imitation policies. We introduce a set of alignment objectives along with sampling methods for optimizing these objectives, illustrating the alignment–constraint satisfaction trade-off. To minimize distribution shift during steering, we develop a new inference algorithm for diffusion policies—stochastic sampling—that improves sample alignment with user intent while ensuring trajectories remain within the data manifold. Limitations of this work include the reliance on an expensive sampling procedure to generate behaviors and the lack of formal success guarantees during steering.

To address these limitations, we propose TAMI. Specifically, through TLI, an instantiation of TAMI using temporal logic specifications, we mitigate distribution shift during steering by recovering task constraints through oracle grounding classifiers. Leveraging modes as discrete abstractions, we prove that state-based continuous policies with global stability guarantees can be modulated by task constraints to simulate any discrete symbolic plan satisfying an LTL specification. Consequently, this provides task success guarantees despite arbitrary inference-time physical steering. However, TLI assumes the existence of suitable mode abstractions, reactive logic formulas, and perfect sensors for detecting mode transitions, which can be difficult to obtain without nontrivial domain knowledge.

To relax these assumptions and specifically enable learning mode classifiers directly from sensor streams, we introduce GLiDE, which grounds knowledge from large language models into physical domains via modes. Given a small number of human demonstrations and task descriptions, we demonstrate that GLiDE successfully recovers mode boundaries, enabling reactive behavior synthesis. Although GLiDE does not require extensive human demonstrations, it relies on substantial trial-and-error exploration and environment resets for collecting task-success labels. Additionally, effectively prompting the LLM to select suitable state representations requires expertise. Future work could explore learning state representations jointly with mode classifiers in an end-to-end manner.

\section{Future Directions}

This thesis demonstrated how recent advances in policy learning can enhance human-robot interaction (HRI) by enabling skill acquisition through imitation. The contributions focus on reusing these skills by making pretrained autonomous policies steerable through user interactions. Future work could further investigate how HRI can, in turn, improve policy learning by enabling more efficient data collection, improving constraint annotation, and developing better algorithms for translating corrective feedback into user preferences—ultimately reducing the data burden required for policy customization.

Recent successes in imitation learning have been driven by kinesthetic teaching and natural teaching systems \cite{zhao2023learning, chi2024universal}, which optimize usability for rapid data collection. Systems that enable seamless switching among various teaching modes, such as teleoperation, kinesthetic teaching, and natural teaching \cite{hagenow2024versatile}, provide an important foundation. Future studies could systematically identify which teaching mechanisms are best suited for specific task domains, thereby maximizing data collection efficiency.

To further improve policy learning and alignment, promising directions include exploring methods for explicitly labeling constraints during data collection and developing algorithms that infer constraints from negative or corrective feedback \cite{newman2023towards, gutierrez2018towards}. Pretrained policies often disregard critical constraints because demonstrators typically provide continuous trajectories without annotating discrete modes or associated requirements. Improved constraint labeling and learning processes could enable more effective policy customization with minimal user effort.

Ultimately, to make robots more accessible and customizable in everyday environments, large-scale real-world deployments of steerable pretrained policies will be important. For instance, deploying low-cost mobile manipulators \cite{wu2024tidybot++} across homes, factories, and retail stores could generate large-scale datasets of user interactions. Such datasets would enable the development and open-sourcing of vision-interaction-action models that are explicitly conditioned on inference-time interactions, i.e., $a = \pi(s, z)$, rather than pretraining non-interactive policies $a = \pi(s)$ and making them steerable post hoc (which inevitably introduces some distribution shift). These models could significantly accelerate progress for the broader robotics community.

\section{The Need for Steering in Robotics}

\begin{figure}[t]
    \centerline{\includegraphics[width=1\textwidth]{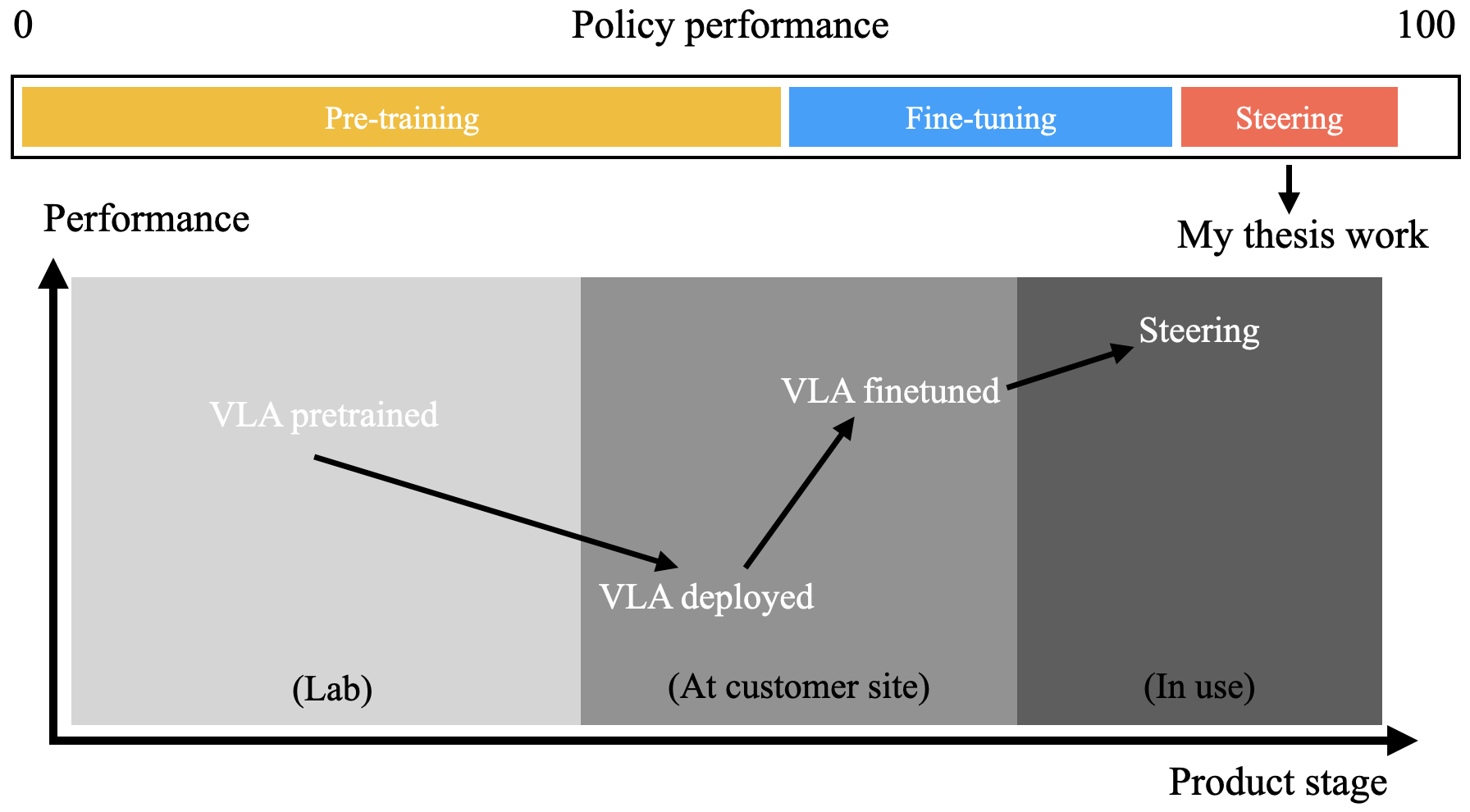}}
    \caption{
        (Left) \textbf{VLA Performance vs. Product Stage.} To enable end-users to achieve individualized task success with a generalist policy, we require robust pretraining to establish a strong base policy, finetuning to adapt the model to specific environments and tasks, and steerability to allow users to customize robot behavior when a residual performance gap remains.
    }
    \label{fig:conclusion}
\end{figure}

While the preceding sections detailed specific technical contributions and future research directions, this final section reflects on the broader necessity of enabling real-time steering to achieve robust, customizable robotic systems in everyday environments. Following the success of large-scale data collection and model training in non-embodied domains \cite{brown2020language, radford2021learning}, roboticists are beginning to adopt a similar paradigm \cite{sutton2019bitter}, training Vision-Language-Action (VLA) models at increasingly larger scales in pursuit of generalist policies with open-world generalization capabilities \cite{black2024pi_0, team2025gemini}. However, robotics fundamentally differs from non-embodied domains such as vision and language: the amount of available training data cannot feasibly cover the breadth of environmental and task variations encountered at inference time. As a result, even VLAs with strong pretraining performance often experience significant degradation when deployed in new environments or tasked with novel instructions (Figure~\ref{fig:conclusion}).

While one could argue that continued data scaling may eventually yield pretrained VLAs that succeed 99.99\% of the time out of the box, achieving such reliability is unlikely without deploying existing pretrained models into real-world settings—homes, factory floors, retail stores—and collecting data during actual use. Only when robots are integrated into daily life will we create a scalable stream of diverse, real-world data to further improve base models during subsequent pretraining phases. This raises a critical question: how do we motivate end-users to adopt robotic automation if current systems only perform reliably some of the time?

The prevailing approach is to finetune pretrained models with domain-specific data collected from end-users \cite{bommasani2021opportunities}. For instance, consider a grocery store that acquires a robot to stack shelves. After receiving the robot and its pretrained VLA model, the store may collect task-specific data to finetune the model before deploying it operationally. Indeed, after finetuning with store-specific data, the robot’s performance can significantly improve, as illustrated in Figure~\ref{fig:conclusion}. However, once deployed, the robot inevitably encounters out-of-distribution situations or novel task commands for which it was not specifically trained. Consequently, failures will still occur.

Although it is technically possible to pause the robot and initiate additional finetuning whenever failures arise, doing so creates an unacceptable user experience: users expect issues to be addressed within seconds, not hours. Frequent disruptions would erode user trust and discourage the adoption of automation solutions that require constant retraining. What is needed to bridge this gap is the introduction of mechanisms that allow users to steer robot behavior at inference time, enabling quick corrections without interrupting ongoing workflows. These user interventions can then be logged automatically and utilized to further finetune the model during off-duty periods when the robot is not actively deployed.

This work calls on the robotics community: if we recognize that passive data collection in laboratory settings will only yield pretrained VLAs with limited out-of-the-box generalization, then we must prioritize the development of inference-time steering methods that bring humans into the execution loop. Through a virtuous cycle of on-the-job corrections and off-duty finetuning in real-world environments, we can ultimately realize generalist policies capable of achieving 99.99\% success rates at deployment. By making inference-time steering a core capability of robotic systems, we can bridge the gap between imperfect pretrained policies and the dynamic, evolving needs of real-world environments. Building systems that are both steerable and continually improvable will be key to realizing the vision of customizable, trustworthy robots that integrate seamlessly into everyday life.

\appendix
\chapter{Task and Motion Imitation with LTL Specification}

\section{Proofs}\label{sec:tli_proof}
\setcounter{theorem}{0}
\begin{theorem}
    (Key Contribution 1) A nonlinear DS defined by Eq. \ref{eq:tli_ds_eq}, learned from demonstrations, and modulated by cutting planes as described in Section \ref{sec:tli_invariance} with the reference point $x^r$ set at the attractor $x^*$, will never penetrate the cuts and is G.A.S. at $x^*$.
\end{theorem} 
\textit{Proof\quad} Let the region bounded by cuts be $\mathcal{D}$, which is non-empty as it contains at least one demonstration. If $x \notin \mathcal{D}$, i.e., $x$ is outside the cuts, the nominal DS $f(x)$ will not be modulated. Since $f(x)$ is G.A.S. at $x^*$ and $x^* \in \mathcal{D}$, a robot state at $x$ will enter $\mathcal{D}$ in a finite amount of time. If $x \in \mathcal{D}$ and $[E(x)^{-1}f(x)]_1<0$, which corresponds to $f(x)$ having a negative component in the direction of $\mathbf{r^*}(x)=\frac{x-x^*}{\|x-x^*\|}$, $f(x)$ is moving away from cuts and toward the attractor. In this case, we leave $f(x)$ unmodulated and the original G.A.S. property holds true. If $x \in \mathcal{D}$ and $[E(x)^{-1}f(x)]_1\geq0$, where the nominal DS could flow out of the cuts, we apply modulation---and, by construction, $M(x)f(x)$ stays inside the cuts. To prove the stability of the modulated DS, we show that the Lyapunov candidate, $V(x)=(x-x^*)^T P(x-x^*)$, in satisfying $\dot{V}(x)=\frac{\partial V(x)}{\partial x}f(x) < 0$ for $f(x)$, also satisfies the Lyapunov condition for $M(x)f(x)$ (omitting matrix dependency upon $x$ to reduce clutter):
\begin{equation}
\begin{aligned}
\dot{V}(x) &= \frac{\partial V(x)}{\partial x}Mf(x) = \frac{\partial V(x)}{\partial x}EDE^{-1}f(x)\\
& = \frac{\partial V(x)}{\partial x}E~\textbf{diag(}1-\Gamma(x),1,...,1\textbf{)}E^{-1}f(x)\\
& = \frac{\partial V(x)}{\partial x}f(x) - \frac{\partial V(x)}{\partial x}E~\textbf{diag}(\Gamma(x),0,...,0)E^{-1}f(x)\\
& < 0 - \frac{\partial V(x)}{\partial x}\mathbf{r^*}(x)^T\Gamma(x)[E^{-1}f(x)]_1\\
& < 0 - \underbrace{2(x-x^*)^T P \frac{x-x^*}{\|x-x^*\|}}_{>0 ~\text{as}~ P\succ 0}\underbrace{\Gamma(x)}_{>0}\underbrace{[E^{-1}f(x)]_1}_{\geq0} < 0
\end{aligned}
\end{equation}
Therefore, $M(x)f(x)$ is G.A.S. \qedsymbol{}\\

The following lemmas support the proof of \textbf{Theorem 2}.
\begin{lemma} \label{eq:tli_lemma1}
    LTL-DS generates a discrete reactive plan of mode sequence that satisfies any LTL formula provided to the algorithm. 
\end{lemma}
\textit{Proof\quad} Since the task automaton is converted from an LTL formula, all resulting discrete plans of mode sequence (including the replanned sequence caused by perturbations) are correct by construction as long as the environment is admissible. \qedsymbol{}

\begin{lemma} \label{eq:tli_lemma2}
    If a mode transition $\sigma_i \Rightarrow \sigma_j$ has been observed in the demonstrations, $\sigma_j$ is reachable from $\sigma_i$ by DS $f_{i}$.
\end{lemma}
\textit{Proof\quad} Since $\sigma_i \Rightarrow \sigma_j$ has been demonstrated, $\sigma_i$ and $\sigma_j$ must be connected; let them share a guard, $G_{ij}$. Assigning a globally stable DS $f(\cdot):\mathbb{R}^n\rightarrow \mathbb{R}^n$ to each mode $\sigma_i$ with region $\delta_i \subset \mathbb{R}^n$ guarantees asymptotic convergence of all $x$ in $\delta_i$ to the attractor, $x^*_{i}$ by DS $f_{i}$. Placing $x^*_{i}$ on guard $G_{ij}$ ensures that $x^*_{i} \in \delta_j$, and thus $\forall s ~\sigma_i \Rightarrow \sigma_j ~\text{as} ~x \rightarrow x^*_{i}$. \qedsymbol{}

\begin{lemma} \label{eq:tli_lemma3}
    If an unseen mode transition $\sigma_i \Rightarrow \sigma_j$ occurs unexpectedly, the system will not be stuck in $\sigma_j$.
\end{lemma}
\textit{Proof\quad} While the transition $\sigma_i \Rightarrow \sigma_j$ has not been seen in demonstrations, Asm. \ref{eq:tli_non-oracle} ensures that mode $\sigma_j$ has been observed and its associated DS $f_{j}$ has been learned. Since the LTL GR(1) fragment does not permit clauses in the form of (\textbf{F G$\phi$}), which states $\phi$ is eventually globally true (i.e., the system can stay in $\sigma_j$ forever), every discrete plan will have to in finite steps result in $\sigma_j \Rightarrow \sigma_k$ for some $k,\ j\neq k$. Having learned $f_{j}$ also validates the existence of $x^*_{j}$---and, thus, a continuous trajectory toward $G_{jk}$. \qedsymbol{}

\begin{theorem} (Key Contribution 2)
    The continuous trace of system states generated by LTL-DS satisfies any LTL specification $\phi$ under Asm. \ref{eq:tli_convexity}, \ref{eq:tli_finiteness}, and \ref{eq:tli_non-oracle}. 
\end{theorem}

\textit{Proof\quad} Lemma \ref{eq:tli_lemma1} proves any discrete plan generated by LTL-DS satisfies the LTL specification. Lemmas \ref{eq:tli_lemma2} and \ref{eq:tli_lemma3} and Asm. \ref{eq:tli_finiteness} ensure the reachability condition of all modes. Thm. \ref{eq:tli_kc1} certifies the modulated DS will be bounded inside the cuts, and thus the mode these cuts inner-approximate. Consequently, a finite number of external perturbations only require a finite number of cuts in order to ensure mode invariance. Given that bisimulation is fulfilled, the continuous trace generated by LTL-DS simulates a LTL-satisficing discrete plan, and thus satisfies the LTL. \qedsymbol{}

\section{Motivation for Mode-based Imitation}
Our work aims to achieve generalization in regions of the state space not covered by initial demonstrations. A well-studied line of research is to collect more expert data \cite{ross2011reduction} so that the policy will learn to recover from out-of-distribution states. Our central observation in Fig. \ref{fig:tli_intro} is that there exists some threshold that separates trajectories deviating from expert demonstrations (black) into successes (blue) and failures (red). The threshold can be embodied in mode boundaries, which lead to the notion of a discrete mode sequence that acts as the fundamental success criteria for any continuous motions. In fact, online data collection to improve policies in DAGGER \cite{ross2011reduction} can be seen as implicitly enforcing mode invariance. We take the alternative approach of explicitly estimating mode boundaries and shift the burden from querying for more data to querying for a task automaton in the language of LTL.

\section{Sensor-based Motion Segmentation and Attractor Identification} \label{sec:tli_segmentation}

\begin{table}[!h]
    \centering
    \begin{tabular}{ccccccccccc}
    \hline
    time step & 1 & 2 & 3 & 4 & 5 & 6 & 7 & 8 & 9 & 10 \\ \hline
    demo 1    & \color{red}$x^{1,1}$ & \color{red}$x^{2,1}$ & \color{blue}$x^{3,1}$ & \color{blue}$x^{4,1}$ & \color{blue}$x^{5,1}$ & \color{blue}$x^{6,1}$ & \color{green}$x^{7,1}$ & \color{green}$x^{8,1}$ & \color{green}$x^{9,1}$ & \color{green}$x^{10,1}$\\
    demo 2    & \color{red}$x^{1,2}$ & \color{red}$x^{2,2}$ & \color{red}$x^{3,2}$ & \color{red}$x^{4,2}$ & \color{blue}$x^{5,2}$ & \color{blue}$x^{6,2}$ & \color{blue}$x^{7,2}$ & \color{blue}$x^{8,2}$ & \color{blue}$x^{9,2}$ & \color{green}$x^{10,2}$\\
    \hline
    \end{tabular}
    \caption{Demonstrations are segmented into three AP regions (shown by color) based on three unique sensor states for DS learning. We use the average location of the last states (transition states to the next AP) in each AP as the attractor for the corresponding DS.}
    \label{tab:tli_segmentation}
\end{table}

Let $\{\{x^{t,k},\dot{x}^{t,k}, \alpha^{t,k}\}_{t=1}^{T_k}\}_{k=1}^{K}$ be $K$ demonstrations of length $T_k$. The motion trajectories in $x^{t, k}$ are clustered and segmented into the same AP region if they share the same sensor state $\alpha^{t,k}$. For example, in Table. \ref{tab:tli_segmentation} two demonstrations of ten time steps form three AP regions (colored by red, blue and green) based on three unique sensor readings. To obtain the attractor for each of the three DS to be learned , we average the last state in the trajectory segment. For instance, the average location of $x^{2,1}$ and $x^{4,2}$, $x^{6,1}$ and $x^{9,2}$, $x^{10,1}$ and $x^{10,2}$ become the attractor for the red, blue and green AP's DS respectively. 

\section{Relation of TLI to Prior Work}
\begin{figure}[!h]
  \centering
  \includegraphics[width=0.9\linewidth]{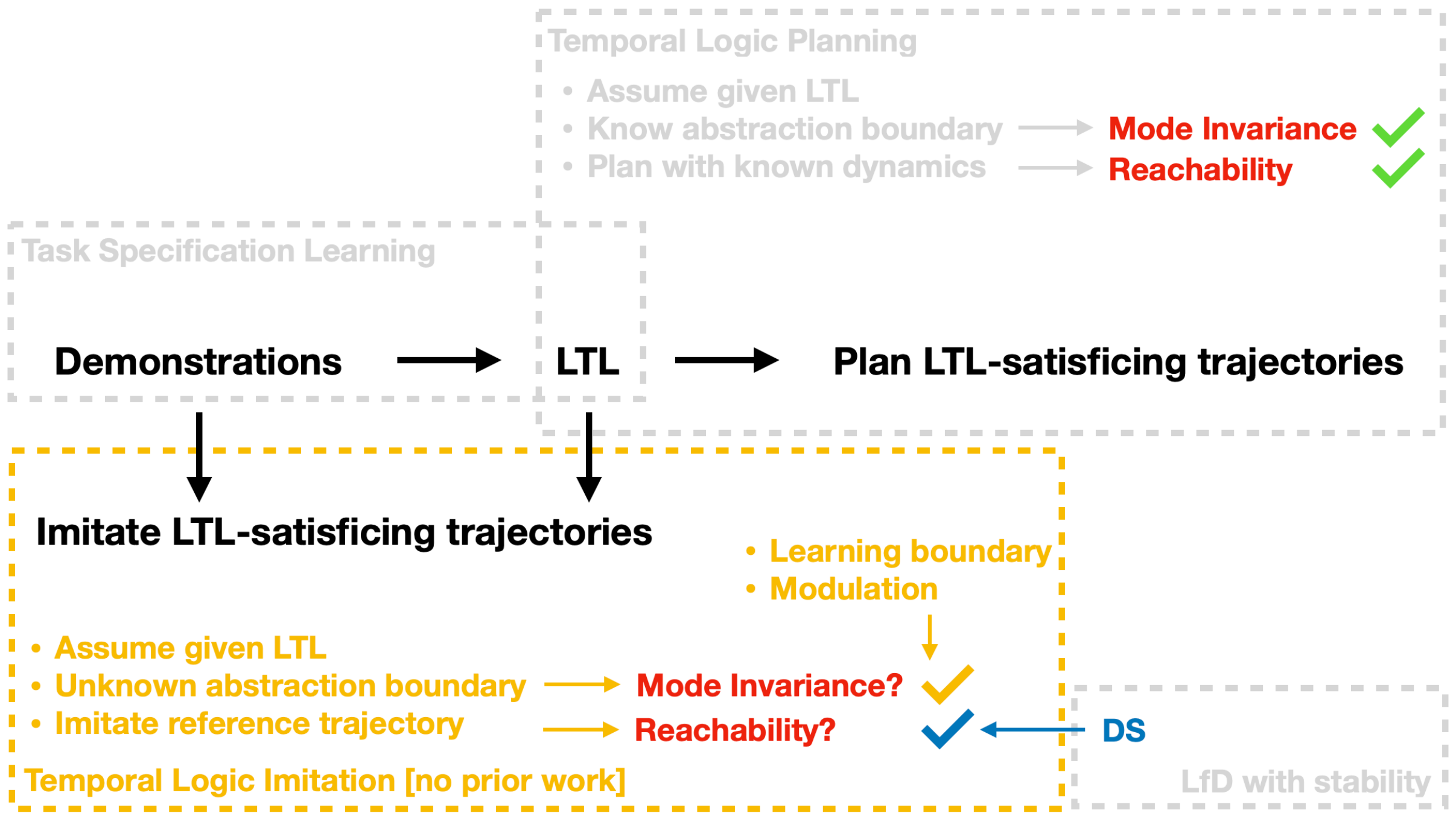}
  \caption{How TLI fits in relation to prior work, where gray dashed boxes represent prior work and yellow dashed box highlights our contribution.}
  \label{fig:tli_prior_work}
\end{figure}

This work explores a novel LfD problem formulation (temporal logic imitation) that is closely related to three research communities. First, there is a large body of work on learning task specifications in the form of LTL formulas from demonstrations \cite{shah2018bayesian, chou2021learning, kasenberg2017interpretable}. We do not repeat their endeavor in this work and assume the LTL formulas are given. Second, given LTL formulas there is another community (temporal logic planning) that studies how to plan a continuous trajectory that satisfies the given LTL \cite{belta2007symbolic, wolff2013automaton, plaku2016motion, kress2018synthesis}. Their assumption of known abstraction boundaries and known dynamics allow the planned trajectory to satisfy the invariance and reachability (bisimulation) criteria respectively, thus certifying the planned continuous trajectory will satisfy any discrete plan. Our observation is that the bisimulation criteria can also be used to certify that a LfD policy can simulate the discrete plan encoded by any LTL formula, which we dub as the problem of TLI. To the best of our knowledge, our work is the first to formalize TLI and investigate its unique challenges inherited from the LfD setting. On the one hand, we no longer have dynamics to plan with but we have reference trajectories to imitate. To satisfy reachability, it is necessary to leverage a third body of work--LfD methods with global stability guarantee (DS) \cite{khansari2011learning, figueroa2018physically, billard2022learning}. On the other hand, we note LfD methods typically do not satisfy mode invariance due to unknown mode boundaries that are also innate to the LfD setting. Thus, we propose learning an approximate mode boundary leveraging sparse sensor events and then modulating the learned policies to be mode invariant. We prove DS policies in particular after modulation will still satisfy reachability, and consequently certifying they will satisfy any given LTL formulas. Figure \ref{fig:tli_prior_work} summarizes how TLI's relationship to prior work, where gray dashed boxes represent prior work and yellow dashed box highlights our contribution.

\section{Single-mode Experiments} \label{sec:tli_single-mode}
\subsection{Experiment Details}
\begin{figure}[h!] 
    \centering
    \newcommand{\myimgwidth}{0.935\textwidth} 

    \begin{minipage}[b]{\myimgwidth}
        \includegraphics[width=\textwidth]{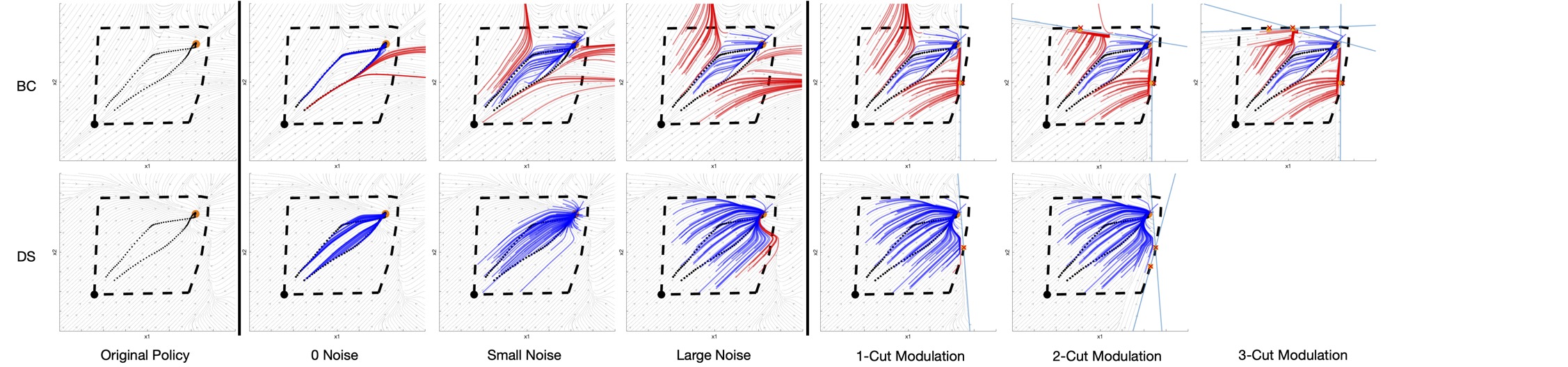}
    \end{minipage}
    \hfill
    \begin{minipage}[b]{\myimgwidth}
        \includegraphics[width=\textwidth]{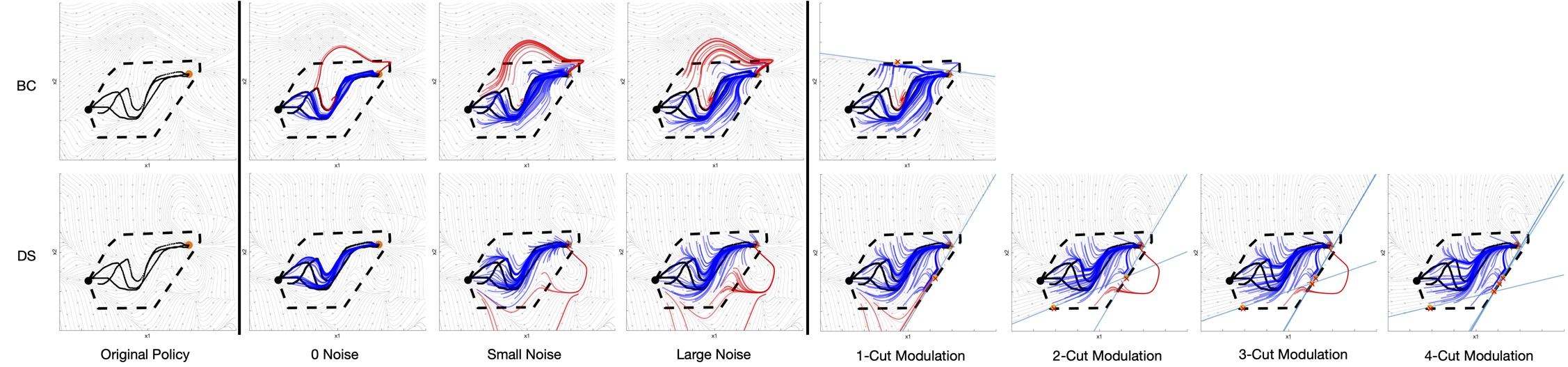}
    \end{minipage}
    \hfill
    \begin{minipage}[b]{\myimgwidth}
        \includegraphics[width=\textwidth]{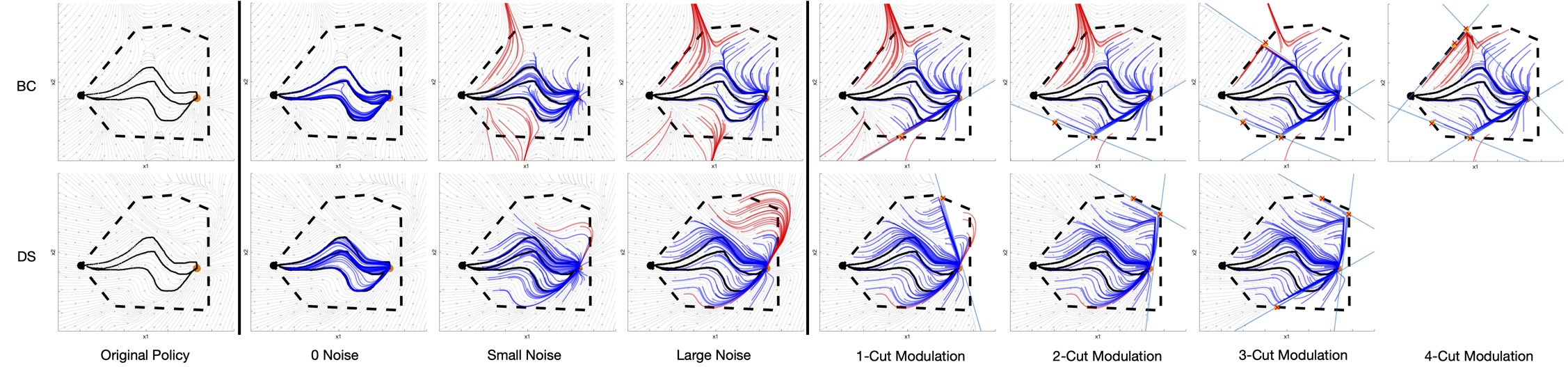}
    \end{minipage}
    \hfill
    \begin{minipage}[b]{\myimgwidth}
        \includegraphics[width=\textwidth]{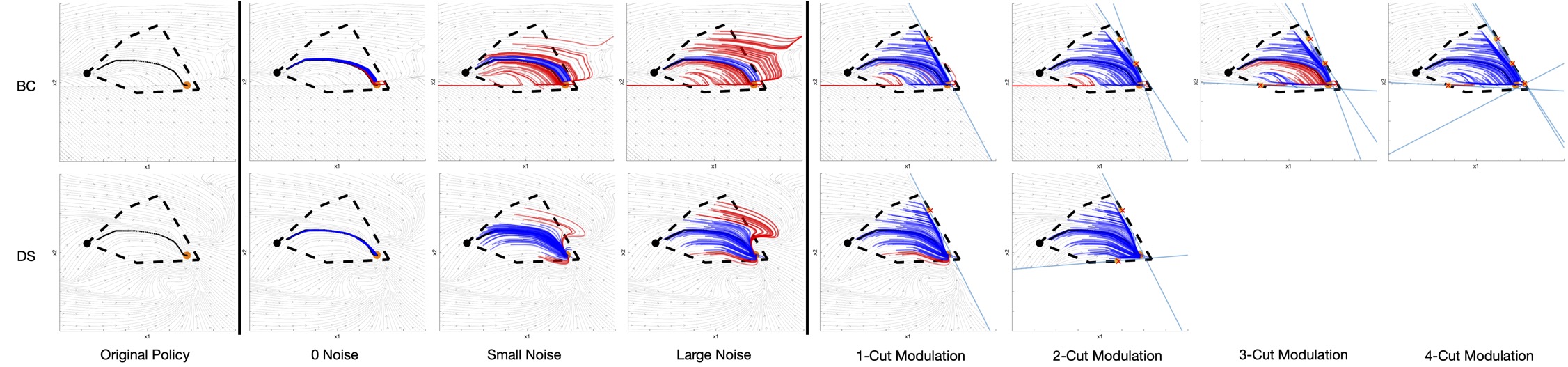}
    \end{minipage}
    \hfill
    \begin{minipage}[b]{\myimgwidth}
        \includegraphics[width=\textwidth]{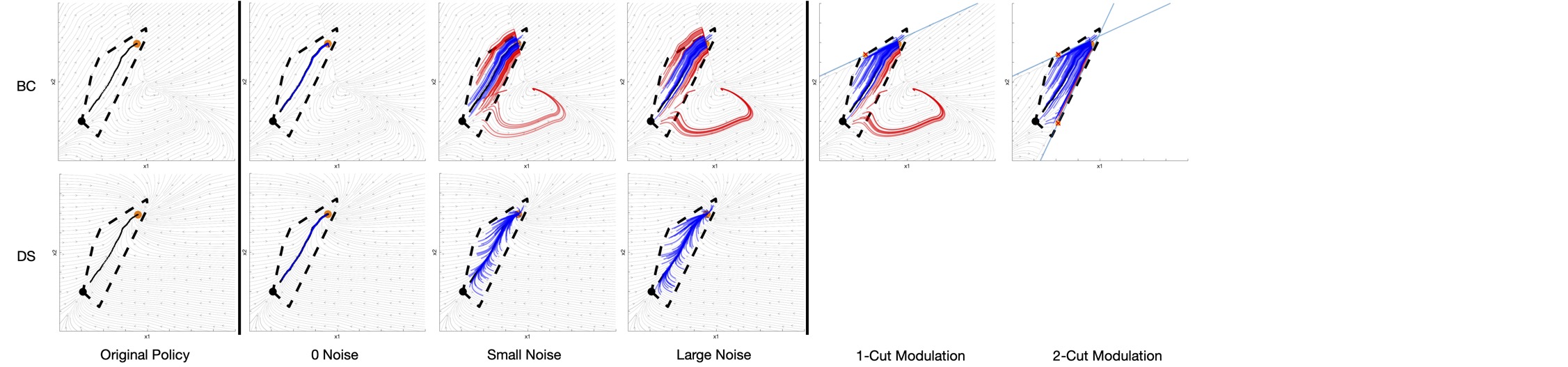}
    \end{minipage}
    
    \caption{For each convex mode, we use 1--3 demonstrations for learning (black). Successful rollouts are shown in blue, unsuccessful rollouts in red. We apply modulation to the large noise case; within four cuts, all DS policies are modulated to be mode invariant. While BC policies can also be modulated, they suffer from reachability failures both pre- and post-modulation. For example, BC flows that originally flow out of the mode can lead to spurious attractors at the cuts after modulation. We prove this cannot happen for DS due to its stability guarantee.}
    \label{fig:tli_additional_rollouts}
\end{figure}

We randomly generate convex modes and draw $1-3$ human demonstrations, as seen in Fig. \ref{fig:tli_rollout} (left). To check mode invariance, we sample starting states from the demonstration distribution perturbed by Gaussian noise with standard deviation of 0\%, 5\%, and 30\% of the workspace dimension. Sampling with zero noise corresponds to sampling directly on the demonstration states, and sampling with a large amount of noise corresponds to sampling from the entire mode region. To enforce invariance, we iteratively sample a failure state and add a cut until all invariance failures are corrected. A task replay is successful if and only if an execution trajectory both reaches the goal and stays within the mode. For each randomly generated convex mode, we sampled 100 starting states and computed the average success rate for 50 trials. We show DS+modulation ensures both reachability and invariance for five additional randomly sampled convex modes in Fig. \ref{fig:tli_additional_rollouts}.

\subsection{BC Policy Architecture and Training Details}
For the Neural-network-based BC policy, we use an MLP architecture that consists of 2 hidden layers both with 100 neurons followed by ReLU activations. We use tanh as the output activation, and we re-scale the output from tanh layer to [-50 -- 50]. Each demonstration trajectory consists of about 200 pairs of states and velocities as the training data to the network. Since we are training a state-based policy that predicts velocities from input states, we treat these data points as i.i.d. For training, we use Adam as the optimizer with a learning rate of 1e-3 for max 5000 epochs. 

\subsection{QCQP Optimization Details} \label{sec:tli_qcqp}
To find the normal direction of a hyperplane that goes through a last-in-mode states $x^{T_f-1}$ in Sec. \ref{sec:tli_invariance}, we solve the following optimization problem, where $w$ is the normal direction we are searching over; $f=1, 2, ...$ indexes a set of failure states; and $T_f$ is the corresponding time-step of first detecting an invariance failure ($T$ alone is used as matrix transpose.)
\begin{equation}
\begin{aligned}
\min_{w} \quad & (w^T(x^*-x^{T_f-1}))^2\\
\textrm{s.t.} \quad & \|w\|=1 \\
& w^T(x^*-x^{T_f-1}) \leq 0\\
& w^T(x^0-x^{T_f-1}) \leq 0\\
& w^T(x^{T_{f^{\prime}}-1} - x^{T_f-1}) \leq 0 \quad \forall f^{\prime}
\end{aligned}
\end{equation}
While specialized QCQP packages can be used to solve this optimization problem, we use a generic nonlinear Matlab function \texttt{fmincon} to solve for $w$ in our implementation.


\section{Multi-modal Experiments} \label{sec:tli_multi-modal}
After abstractions---environment APs, $r,s,t$, and robot APs, $a,b,c,d$---for the soup-scooping task in Sec. \ref{sec:tli_scooping_soup_task} are defined, the reactive LTL formula can be written as $\phi = ((\phi^e_i\land \phi^e_t\land \phi^e_g) \rightarrow (\phi^s_i\land \phi^s_t\land \phi^s_g))$. $\phi_i^s$ and $\phi_i^e$ specify the system's initial mode, $a$, and the corresponding sensor state. $\phi_g^s$ and $\phi_g^e$ set mode $d$ as the eventual goal for the robot, with no particular goal for the environment. $\phi_t^e$ specifies the environmental constraints that determine which sensor states are true in each mode, as well as the fact that the system can only be in one mode at any times. $\phi_t^s$ specifies all valid transitions for each mode.
\begin{align*}
    \phi_i^e = &\ \lnot r\land \lnot s \land \lnot t; \quad \phi_g^e = True;\\
    \phi_t^e = &\ \textbf{G}((a \leftrightarrow (\lnot r\land \lnot s \land \lnot t)
                \land (b \leftrightarrow (r\land \lnot s \land \lnot t)) \\
                &\quad \land (c \leftrightarrow (\lnot r \land s \land \lnot t)
                \land (d \leftrightarrow (\lnot r \land \lnot s \land t)) \\
                & \land \textbf{G}((a\land \lnot b \land \lnot c \land \lnot d) \lor               (\lnot a\land b \land \lnot c \land \lnot d)\\
               & \quad \lor (\lnot a\land \lnot b \land c \land \lnot d)
               \lor (\lnot a\land \lnot b \land \lnot c \land d));\\
    \phi_i^s = &\ a; \quad \phi_g^s = \textbf{GF}d;\\
    \phi_t^s = &\ \textbf{G}((a \rightarrow (\textbf{X}a \lor \textbf{X}b))
               \land (b \rightarrow (\textbf{X}a \lor \textbf{X}b \lor \textbf{X}c))\\
               & \quad \land (c \rightarrow (\textbf{X}a \lor \textbf{X}b \lor \textbf{X}c \lor \textbf{X}d))
               \land (d \rightarrow \textbf{X}d)); 
\end{align*}
 \textbf{Automatic construction of GR(1) LTL formulas} One benefit of using the GR(1) fragment of LTL is that it provides a well-defined template for defining a system's reactivity \cite{kress2009temporal}\footnote{The main benefit is that GR(1) formulas allow fast synthesis of their equivalent automatons \cite{piterman2006synthesis}.}. While in this work we follow the TLP convention that assumes the full GR(1) formulas are given, the majority of these formulas can actually be automatically generated if Asm. \ref{eq:tli_non-oracle} holds true. Specifically, once the abstraction, $r, s, t, a, b, c, d$ is defined, formulas $\phi_t^e, \phi_g^e$ are correspondingly defined as shown above, and they remain the same for different demonstrations. If a demonstration displaying $a\Rightarrow b\Rightarrow c\Rightarrow d$ is subsequently recorded, formulas $\phi_i^e, \phi_i^s, \phi_g^s$ as shown above can then be inferred. Additionally a partial formula $\phi_t^s=\textbf{G}((a \rightarrow (\textbf{X}a \lor \textbf{X}b)) \land (b \rightarrow (\textbf{X}b \lor \textbf{X}c) \land (c \rightarrow (\textbf{X}c \lor \textbf{X}d)) \land (d \rightarrow \textbf{X}d))$ can be inferred. Synthesis from this partial $\phi = ((\phi^e_i\land \phi^e_t\land \phi^e_g) \rightarrow (\phi^s_i\land \phi^s_t\land \phi^s_g))$ results in a partial automaton in Fig. \ref{fig:tli_method} with only black edges. During online iteration, if perturbations cause unexpected transitions, $b\Rightarrow a$ and/or $c\Rightarrow a$ and/or $c\Rightarrow b$, which are previously not observed in the demonstration, $\phi_t^s$ will be modified to incorporate those newly observed transitions as valid  mode switches, and a new automaton will be re-synthesized from the updated formula $\phi$. The gray edges in Fig. \ref{fig:tli_method} reflect those updates after invariance failures are experienced. Asm. \ref{eq:tli_non-oracle} ensures the completeness of the demonstrations with respect to modes, i.e., the initially synthesized automaton might be missing edges but not nodes compared to an automaton synthesized from the ground-truth full formula. For general ground-truth LTL formulas not part of the GR(1) fragment or demonstrations not necessarily satisfying Asm. \ref{eq:tli_non-oracle}, we cannot construct the formulas using the procedure outlined above. In that case, we can learn the formulas from demonstrations in a separate stage \cite{shah2018bayesian, chou2021learning}. 
 
 In this work, we assume full LTL formulas are provided by domain experts. Since they are full specifications of tasks, the resulting automatons will be complete w.r.t. all valid mode transitions (e.g., including both the black and gray edges in Fig. \ref{fig:tli_method}), and will only need to be synthesized once. Given the soup-scooping LTL defined above, we ran $10$ experiments, generating $1-3$ demonstrations for each, and learning a DS per mode. We applied perturbations uniformly sampled in all directions of any magnitudes up to the dimension of the entire workspace in order to empirically verify the task-success guarantee. We follow the QCQP optimization defined in Appendix B to find cuts to modulate the DS. Simulation videos can be found on the project page.

\section{Generalization Results} \label{sec:tli_generalization}
\begin{figure}[h] 
  \centering
  \includegraphics[width=1\linewidth]{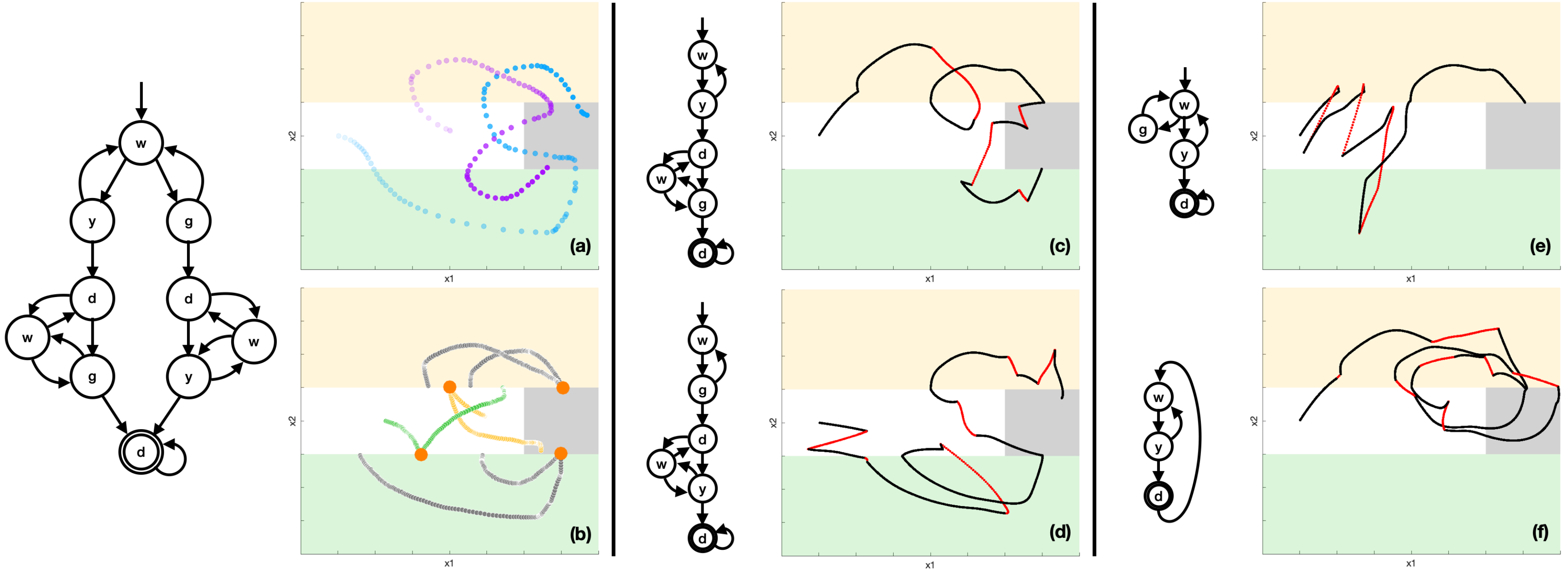}
  \caption{Reuse learned DS for new tasks by editing their task automaton via LTL. \textbf{(a)} Two demonstrations (purple and blue) solve a task of adding both chicken (yellow region) and broccoli (green region) to a pot (dark region) in different order as visualized by the automaton on the left. \textbf{(b)} Demonstrations are segmented into parts by mode region, which are then shifted to their average attractor locations (orange circles) for DS learning. Since a mode can proceed to different next modes in this case, we adapt the mode-based DS policy to be associated with each unique mode transition instead. The learned DS can generalize to new tasks given new automatons specifying \textbf{(c)} the order: white (w)  $\Rightarrow$ yellow (y) $\Rightarrow$ dark (d) $\Rightarrow$ green (g) $\Rightarrow$ dark (d), \textbf{(d)} the order: white $\Rightarrow$ green $\Rightarrow$ dark $\Rightarrow$ yellow $\Rightarrow$ dark, \textbf{(e)} the goal: getting only chicken, or \textbf{(f)} the goal: getting chicken continuously. The trajectory in red shows perturbations and the trajectory in black shows recovery according to the automaton structure. 
  }
  \label{fig:tli_generalization}
\end{figure}

LTL-DS can generalize to a new task by reusing learned DS if the new LTL shares the same set of modes. Consider another multi-step task of adding chicken and broccoli to a pot. Different humans might give demonstrations with different modal structures (e.g., adding chicken vs adding broccoli first) as seen in Fig. \ref{fig:tli_generalization} \textbf{(a)}. LTL-DS learns individual DS which can be flexibly combined to solve new tasks with new task automatons, as illustrated in Fig. \ref{fig:tli_generalization} \textbf{(c-f)}. To get these different task automatons, a human just needs to edit the $\phi_t^s$ portion of the LTL formulas differently.

We describe LTL formulas for variants of the cooking task of adding chicken and broccoli to a pot as visualized in Fig. \ref{fig:tli_generalization}. We use mode AP $w$, $y$, $g$, $d$ to define configurations of empty spoon (white region), transferring chicken (yellow region), transferring broccoli (green region), and finally dropping food in the pot (dark region) . We follow the description of scooping task LTL to define $\phi^e_i, \phi^e_t, \phi^e_g, \phi^s_i, \phi^s_g$ for the cooking tasks, which are shared by them all. We focus on $\phi^s_t$ here as it captures mode transitions and is different for a different task. We denote the $\phi^s_t$ portion of LTL for the new task of adding chicken first, adding broccoli first, adding chicken only, and adding chicken continuously as $\phi_{cb}$, $\phi_{bc}$, $\phi_{c}$, and $\phi_{cc}$ respectively.
\begin{align*}
        \phi_{cb} = &\ \textbf{G}((w_1 \rightarrow \textbf{X}y) \land (y \rightarrow (\textbf{X}w_1 \lor \textbf{X}d_1)) \\ 
        & \land (d_1 \rightarrow (\textbf{X}w_2 \lor \textbf{X}g)) \land (w_2 \rightarrow \textbf{X}g) \\
        & \land (g \rightarrow (\textbf{X}w_2 \lor \textbf{X}d_2)) \land (d_2 \rightarrow \textbf{X}d_2)); \\
        \phi_{bc} = &\ \textbf{G}((w_1 \rightarrow \textbf{X}g) \land (g \rightarrow (\textbf{X}w_1 \lor \textbf{X}d_1)) \\ 
        & \land (d_1 \rightarrow (\textbf{X}w_2 \lor \textbf{X}y)) \land (w_2 \rightarrow \textbf{X}y) \\
        & \land (y \rightarrow (\textbf{X}w_2 \lor \textbf{X}d_2)) \land (d_2 \rightarrow \textbf{X}d_2)); \\
        \phi_{c} = &\ \textbf{G}((w \rightarrow (\textbf{X}y \lor \textbf{X}g)) \land (y \rightarrow (\textbf{X}w \lor \textbf{X}d)) \\ 
        & \land (g \rightarrow \textbf{X}w) \land (d \rightarrow \textbf{X}d)); \\
        \phi_{cc} = &\ \textbf{G}((w \rightarrow \textbf{X}y) \land (y \rightarrow (\textbf{X}w \lor \textbf{X}d)) \land (d \rightarrow \textbf{X}w)); 
\end{align*}
Note mode $w_1$ and $w_2$ denote visiting the white region before and after some food has been added to the pot and they share the same motion policy. The same goes for mode $d_1$ and $d_2$. These formulas can be converted to task automatons in Fig. \ref{fig:tli_generalization}. We show animations of these tasks on the project page.  

\section{Robot Experiment 1: Soup-Scooping} \label{sec:tli_robot_scoop}
\begin{figure}[!h]
  \centering
  \includegraphics[width=0.8\linewidth]{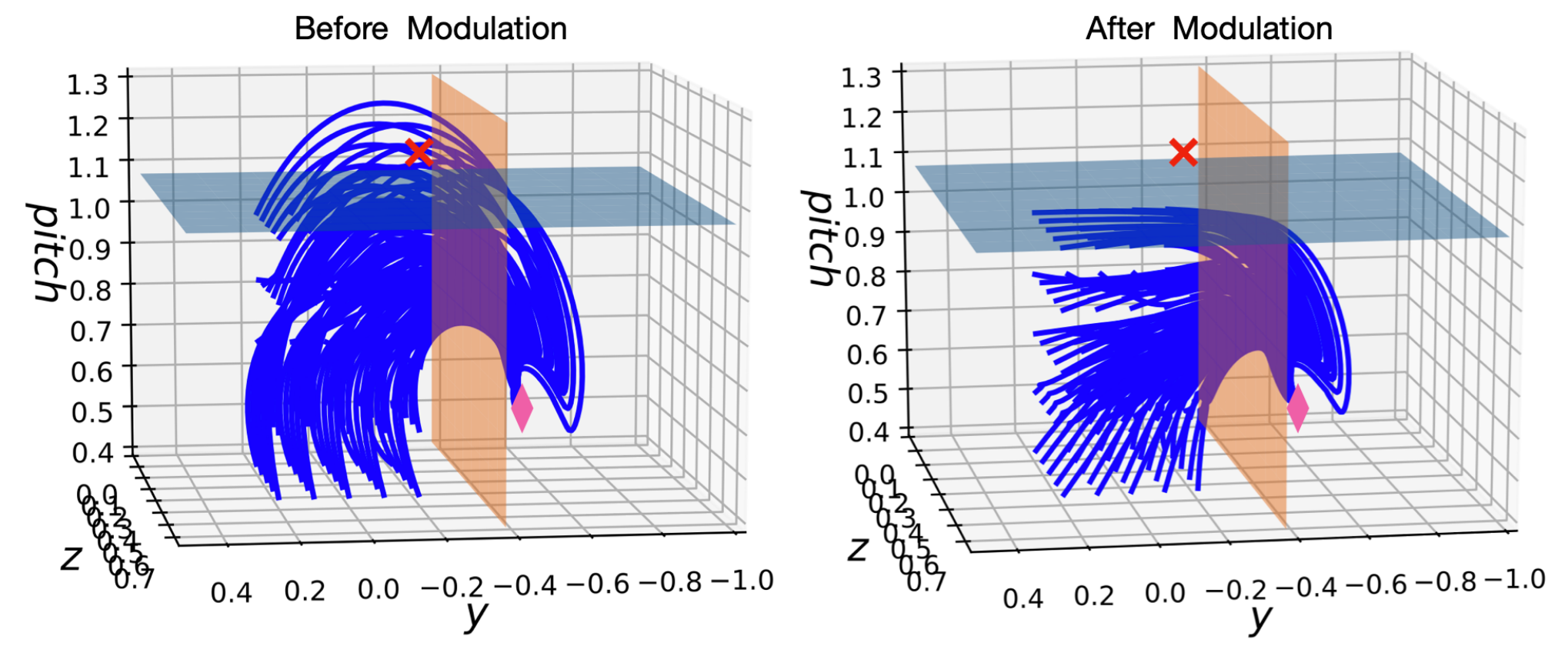}
  \caption{Modulation of sampled robot trajectories. The orange plane represents a guard surface between the transporting mode and the done mode, and the blue plane represents the mode boundary for the transporting mode. The red crosses denote an invariance failure, and the pink diamonds denote the attractor. Before modulation, there are trajectories prematurely exiting the mode; after modulation, all trajectories are bounded inside the mode.
  }
  \label{fig:tli_3d_mod}
\end{figure}

We implemented the soup-scooping task on a Franka Emika robot arm. As depicted in Fig.~1, the task was to transport the soup (represented by the red beads) from one bowl to the other. Two demonstration trajectories were provided to the robot via kinesthetic teaching, from which we learned a DS to represent the desired evolution of the robot end-effector for each mode. The target velocity, $\dot{x}$, predicted by the DS was integrated to generate the target pose, which was then tracked by a Cartesian pose impedance controller. The robot state, $x$, was provided by the control interface. Sensor AP $r$ tracked the mode transition when the spoon made contact with the soup, and sensor AP $t$ tracked the mode transition when the spoon reached the region above the target bowl. $r$ and $t$ became true when the distance between the spoon and the centers of the soup bowl and the target bowl (respectively) were below a hand-tuned threshold. Sensor AP $s$ became true when red beads were detected either from a wrist camera via color detection or through real-time human annotation. We visualize the modulation of robot DS in three dimensions---$y$, $z$, and pitch---in Fig. \ref{fig:tli_3d_mod}.

\begin{sidewaysfigure}
  \centering
  \includegraphics[width=\linewidth]{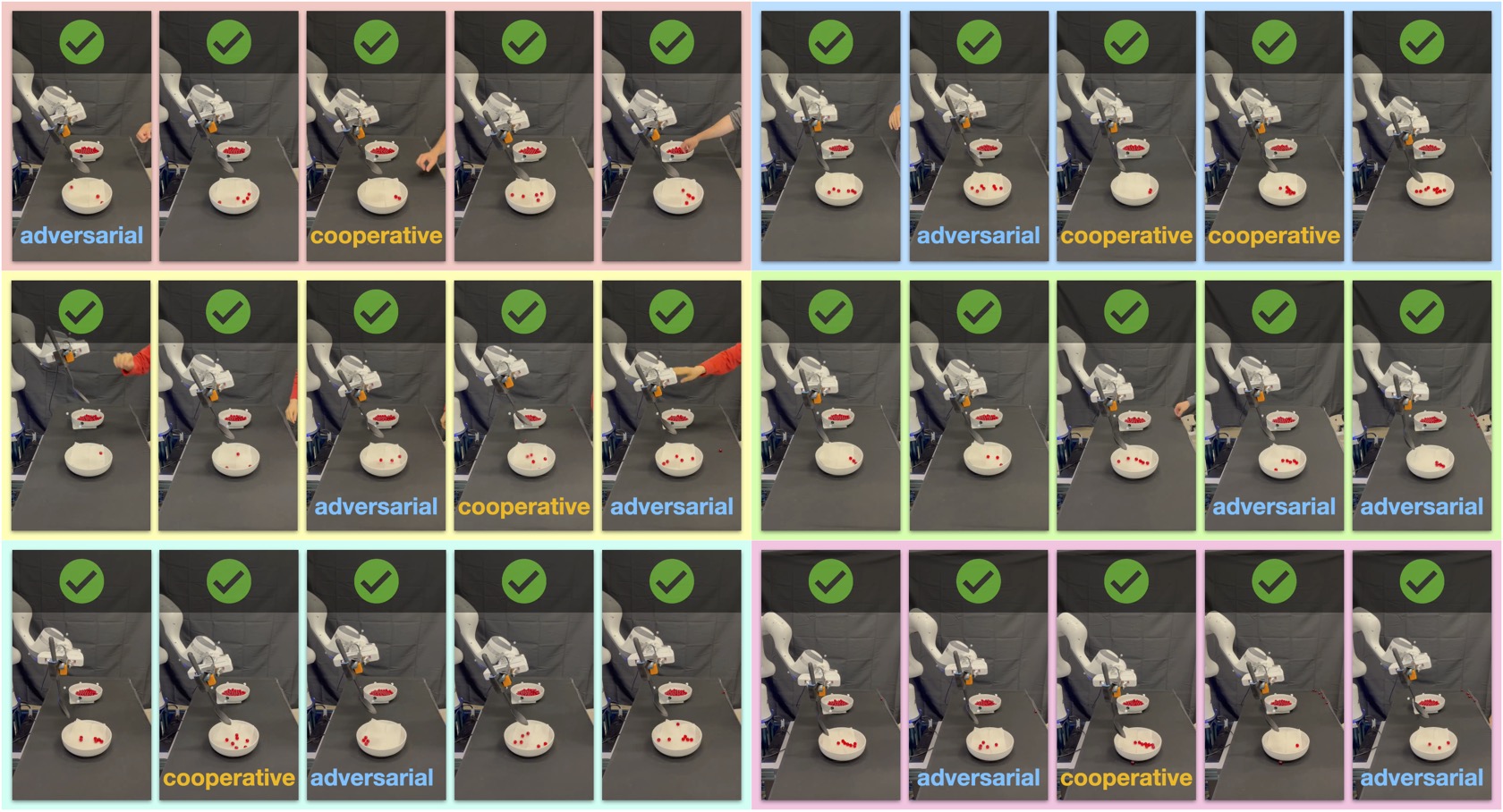}
  \caption{Ending snapshots (100\% success rate, see videos for action) of six randomly recruited human subjects performing unbiased perturbations in a total of 30 trials without cherry-picking. Common perturbation patterns (we annotate with the same colored text) emerge from different participants. Specifically, we see \textit{adversarial perturbations} where humans fight against the robot and \textit{cooperative perturbations} where humans help the robot to achieve the goal of transferring at least one bead from one bowl to the other.
  }
  \label{fig:tli_unbiased}
\end{sidewaysfigure}

\textbf{Unbiased human perturbations.} Since external perturbations are an integral part of our task complexity, we recruited six human subjects without prior knowledge of our LTL-DS system to perturb the robot scooping setup. Each subject is given five trials of perturbations. In total, we collected 30 trials as seen in Fig. \ref{fig:tli_unbiased}, each of which is seen as an unbiased i.i.d. source of perturbations. On our project page, we show all 30 trials succeed eventually in videos. We did not cherry-pick these results, and the empirical 100\% success rate further corroborates our theoretic success guarantee. Interestingly, common perturbation patterns (as seen in the videos) emerge from different participants. Specifically, we see \textbf{adversarial perturbations} where humans fight against the robot and \textbf{cooperative perturbations} where humans help the robot to achieve the goal of transferring at least one bead from one bowl to the other. In the case of adversarial perturbations, DS reacts and LTL replans. In the case of collaborative perturbations, DS is compliant and allows humans to also guide the motion. In the case where humans are not perturbing yet the robot makes a mistake (e.g. during scooping), LTL replans the scooping DS until the robot enters the transferring mode successfully. The fact that we do not need to hard code different rules to handle invariance failures caused by either perturbations or the robot's own execution failures in the absence of perturbations highlights the strength of our LTL-powered sensor-based task reactivity.

\section{Robot Experiment 2: Inspection Line} 

To further validate the LTL-DS approach we present a second experimental setup that emulates an inspection line, similar to the one used to validate the LPV-DS approach \cite{figueroa2018physically} -- which we refer to as the vanilla-DS and use to learn each of the mode motion policies.  In \cite{figueroa2018physically} this task was presented to validate the potential of the vanilla-DS approach to encode a highly-nonlinear trajectory going from \textcolor{red}{ (a) grasping region}, \textcolor{green}{ (b) passing through inspection entry,} (c) follow the inspection line and \textcolor{blue}{(d) finalizing at the release station} with a single DS. In this experiment we show that, even though it is impressive to encode all of these modes (and transitions) within a single continuous DS, if the sensor state or the LTL task-specification are not considered the vanilla-DS approach will fail to achieve the high-level goal of the task which is, to pass slide the object along the inspection line. To showcase this, in this work we focus only on \textcolor{green}{(b) $\rightarrow$}  \textcolor{blue}{(c) $\rightarrow$ (d)} with \textcolor{red}{ (a)} following a pre-defined motion and grasping policy for experimental completeness. 

\begin{figure}[!h]
  \centering
  \includegraphics[width=\linewidth]{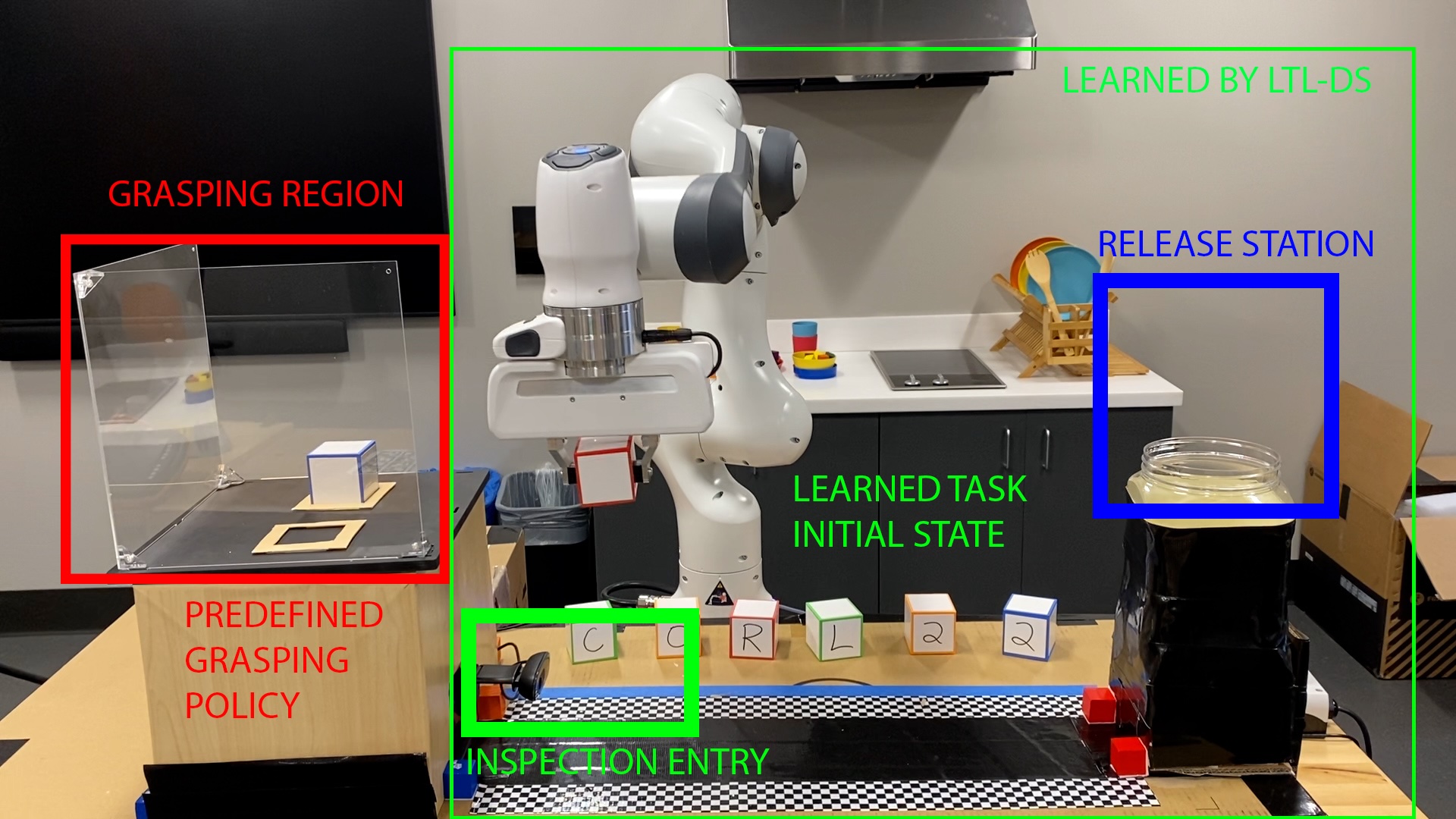}
  \caption{Robot Experiment 2: Inspection Line}
  \label{fig:tli_inspection}
\end{figure}

\textbf{Inspection Task Details.} \label{sec:tli_inspection}
The video of this experiment can be found on our website.
\begin{itemize}[leftmargin=*]
    \item \textit{Sensor model:} We implement the \textit{sensor model} of the inspection task as an object detector on the inspection track and distances to attractors (defined from AP region-based segmentation described in new Appendix I). As we created a black background for the inspection task and the camera is fixed, with a simple blob detector we can detect if the robot is inside or outside of the inspection line. Hence, the \textit{sensor state} is a binary variable analogous to that of the scooping task.
    \item \textit{Task specification:} The proposed inspection task can be represented with 2 modes \textcolor{green}{(a) Go to inspection entry} $\rightarrow$ \textcolor{blue}{(b) follow inspection line and release}. The AP regions are the bounding boxes around the \textit{inspection entry} and \textit{release station} shown in Fig. \ref{fig:tli_inspection} which correspond to the attractor regions for each mode. Mode (a) requires the robot to reach the mode attractor and detecting the presence of the cube once it has been reached. Mode (b) requires the cube to slide along the inspection track (reaching the end) and then lift the cube to drop it at the release station.
    
    \item \textit{Offline Learning:} We use two demonstrations of the inspection task, together with an LTL specification and run our offline learning algorithm used for the soup-scooping task (without any modifications), as shown in the supplementary video from \texttt{0:00-0:18s}. Without any adversarial perturbations or environmentally induced failures, the vanilla-DS approach is capable of accomplishing the defined inspection task without invariance failures as shown in \texttt{0:19-0:32s}.
    
    \item \textit{Invariance Failures of Vanilla-DS:} Even though the vanilla-DS approach is now used to learn a less complex trajectory (in terms of trajectory non-linearity), as we excluded the grasping region, we can see that it easily fails to achieve the inspection task when subject to large adversarial perturbations that lead the robot towards an \textit{out-of-distribution} state. This means that the robot was perturbed in such a way that it is now far from the region where the demonstrations were provided. Yet, it is robust to small adversarial perturbations that keep the robot \textit{in-distribution}, as shown in the supplementary video from \texttt{0:33-1:18min}. The latter is the strength of DS-based motion policies in general and these are the type of perturbations that are showcased in \cite{figueroa2018physically}. However, since the DS is only guaranteed to reach the target by solely imposing Lyapunov stability constraints it always reaches it after a large adversarial perturbation, with the caveat of not accomplishing the actual inspection task. Note that this limitation is not specific to the vanilla-DS approach \cite{figueroa2018physically}, it is a \textbf{general limitation} of goal-reaching LfD methods that only care about guaranteeing stability \textbf{at the motion-level} be it through Lyapunov or Contraction theory. \textit{Hence, by formulating the problem as TLI and introducing sensor states and LTL specification into the imitation policy we can achieve convergence at the motion-level and task-level.}\\
    
    
    \item \textit{Invariance Guarantee with LTL-DS:} As shown in the supplementary video from \texttt{1:19-1:43min} we collect a set of invariance failures to construct our mode boundary. Further, from \texttt{1:43-2:00min} we show the approximated mode boundary from 4 recorded failure states that approximate the vertical boundary and then from 10 recorded failure states which now approximate the horizontal boundary of the mode. The blue trajectories in those videos correspond to rollouts of the vanilla-DS learned from the demonstrations in that mode. 
    
    From \texttt{2:00-3:40min} we show two continuous runs of the inspection task, each performing two inspections. We stress test the learned boundary and LTL-DS approach by performing small and large adversarial perturbations. As shown in the video, when adversarial perturbations are small the DS motion policy is robust and still properly accomplishes the inspection task. When adversarial perturbations are large enough to push the robot outside of the learned boundary, the LTL-DS brings the robot back to the inspection entry mode and tries the inspection line again and again and again until the inspection task is achieved as defined by the LTL specification - \textbf{guaranteeing task completion.} \\
\end{itemize}

\textbf{Comment on Task Definition:} In order to learn and encode the entire task (from grasp to release) with LTL-DS we would need to include a grasping controller within our imitation policy. It is possible to extend the LTL-DS approach to consider grasping within the imitation policy, yet due to time limitations we focus solely on the parts of the task that can be learned by the current policy -- that requires only controlling for the motion of the end-effector. We are concurrently working on developing an approach to learn a grasping policy entirely through imitation, which to the best of our knowledge does not exist within the problem domains we target. In a near future, we plan to integrate these works in order to allow LTL-DS to solve problems that include actuating grippers in such a feedback control framework. Note that, the vanilla-DS approach does not consider the grasping problem either and the experimental setup presented in \cite{figueroa2018physically} uses a simple open-loop gripper controller that is triggered when the DS reaches the attractor, such triggering is hand-coded and not learned either in their setup.

\section{Robot Experiment 3: Color Tracing} \label{sec:tli_color}

\begin{figure}[!h]
  \centering
  \includegraphics[width=1\linewidth]{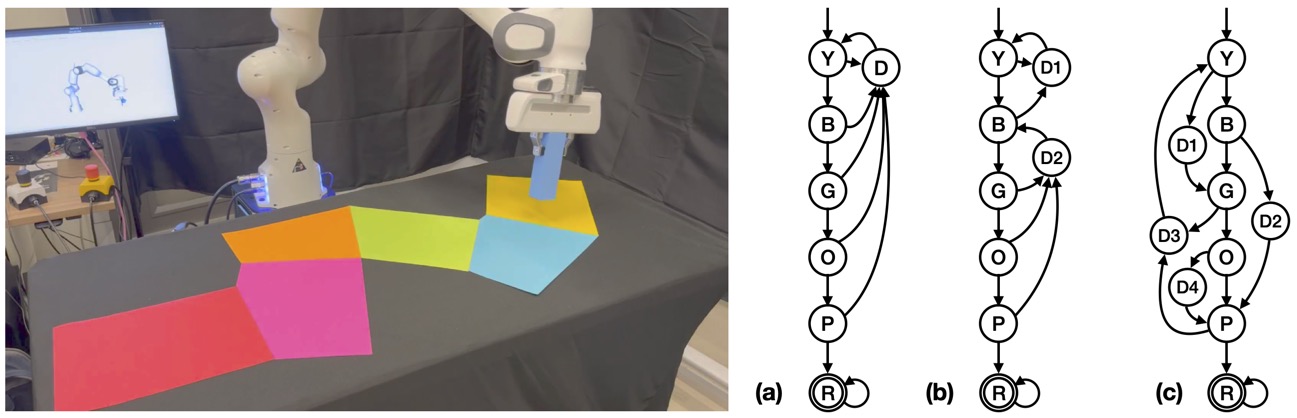}
  \caption{The goal is for the end-effector to move through (while staying within) (Y)ellow, (B)lue, (G)reen, (O)range, (P)ink, and eventually reach (R)ed. If a sensor detects the end-effector is perturbed into the (D)ark region, the system needs to replan according to a given task automaton such as (a), (b), (c). Note D1, D2, D3, D4 refer to different modes (entering the dark region from different colors) that appear visually the same.}
  \label{fig:tli_color}
\end{figure}

This experiment demonstrates LTL-DS' handling of long-horizon multi-step tasks with non-trivial task structures. Given a single demonstration of kinesthetically teaching the robot end-effector to move through the colored tiles, the system learns a DS for each of the colored mode. The learned DS can then be flexibly recombined according to different LTL-equivalent task automatons to react differently given invariance failures. Specifically, we show in the videos on our website  three different replanning: (a) mode exit at any colored tile transitions to re-entry at the yellow tile; (b) mode exit at any colored tile after the blue tile transitions to re-entry at the blue tile; and (c) mode exit at the yellow tile transitions to the blue tile, while mode exit at the blue tile transitions to the pink tile.

\chapter{Learning Grounding Classifiers for Task and Motion Imitation}

\section{More Mode Classification Results for 2D Navigation Tasks}
\label{app:glide_polygon}

\begin{figure}[!htb]
  \centering\small
  \includegraphics[width=\textwidth]{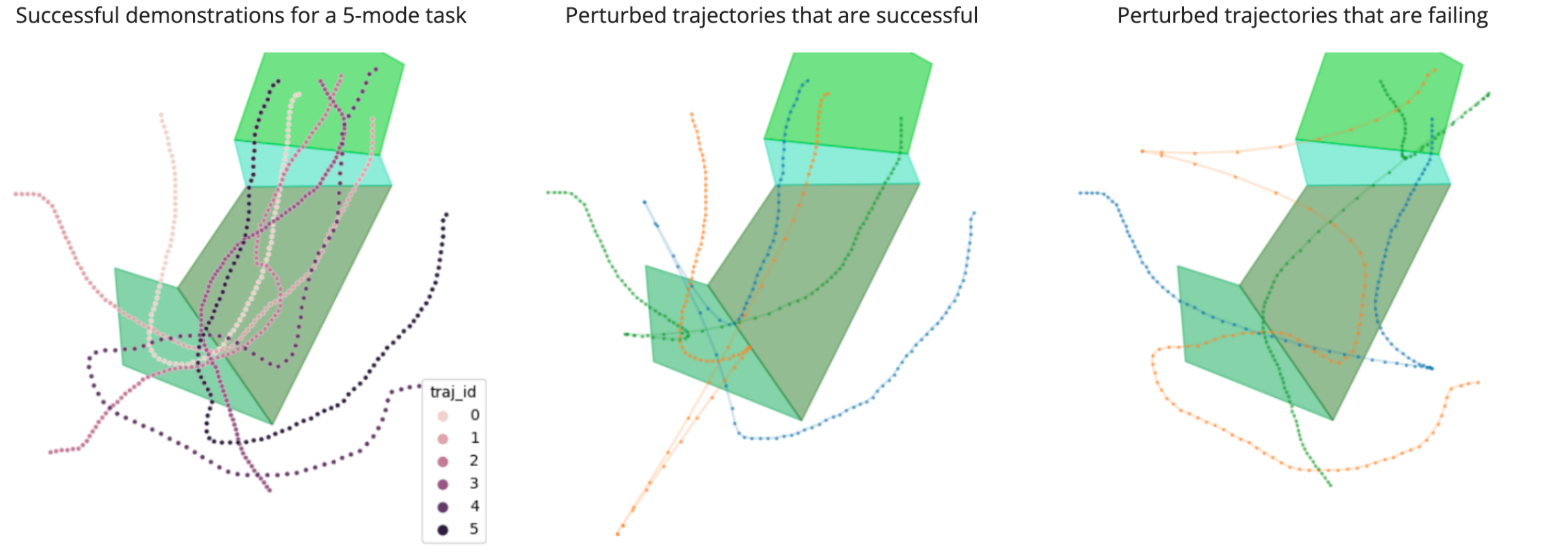}
  \caption{Additional 2D grounding examples. Column (a) shows the ground truth mode segmentation and successful demonstrations. Column (b) shows the learned grounding and its percentage overlap with the ground truth. Columns (c-e) visualize the grounding learned without counterfactual data, a correct feasibility matrix, and task prediction loss respectively.}
  \label{fig:glide_2d_data_gen}
\end{figure}

\begin{figure}[!htb]
  \centering\small
  \includegraphics[width=\textwidth]{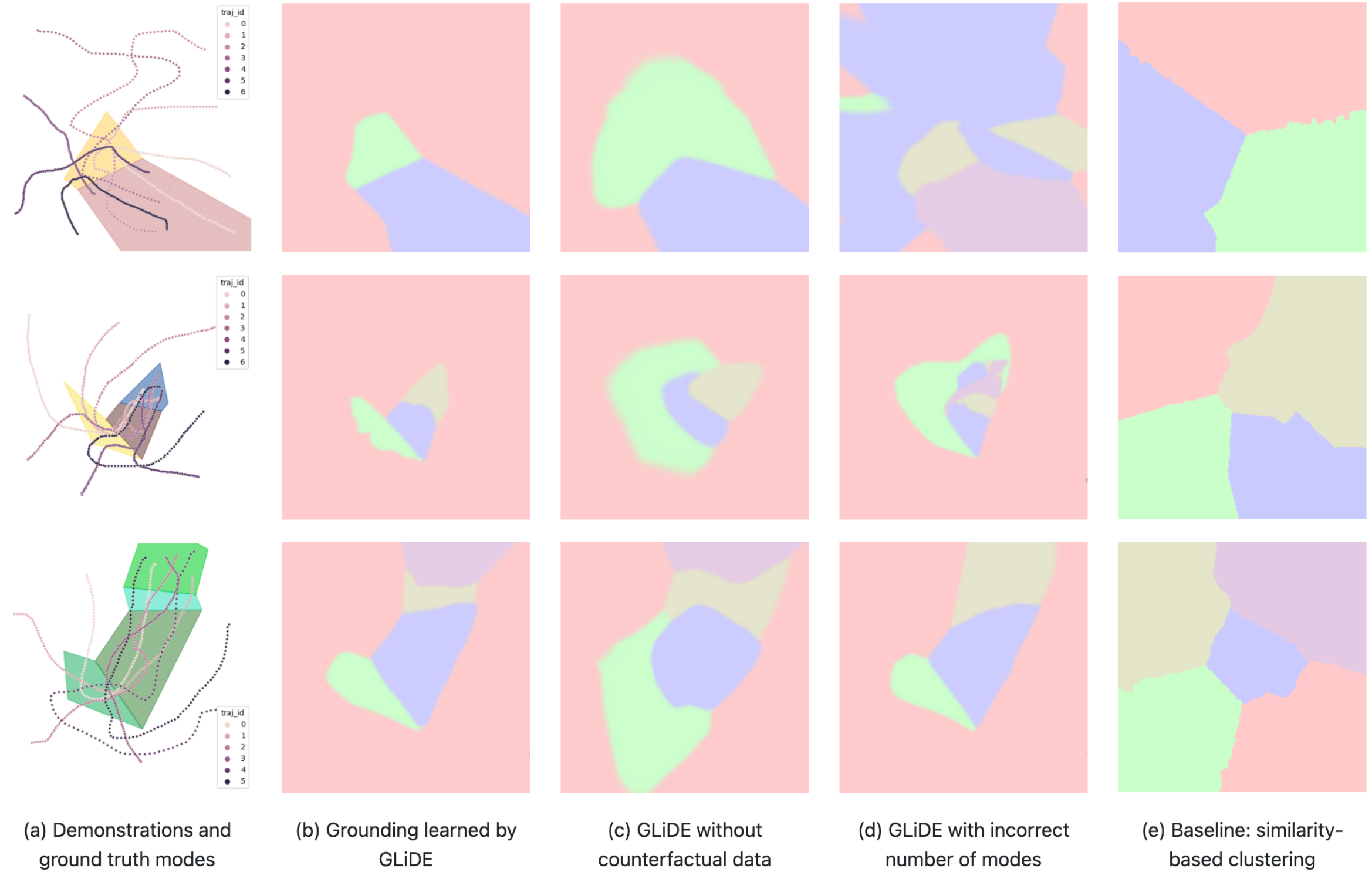}
  \caption{Additional 2D grounding examples. Column (a) shows the ground truth mode segmentation and successful demonstrations. Column (b) shows the learned grounding and its percentage overlap with the ground truth. Columns (c-e) visualize the grounding learned without counterfactual data, a correct feasibility matrix, and task prediction loss respectively.}
  \label{fig:glide_2d_nav_more}
\end{figure}

Figure \ref{fig:glide_2d_data_gen} visualizes data augmentation of generating additional successful trajectories (middle) and failing counterfactuals (right) from a few successful demonstrations (left). Figure \ref{fig:glide_2d_nav_more} visualizes learned grounding for additional randomly generated 2D navigation environments with a 3-, 4- and 5-mode task structure.

\section{Heuristic Rules for Mode Family Groundtruth in RoboSuite}
\label{app:glide_mode-definition-robosuite}

\begin{figure}[!htb]
  \centering\small
  \includegraphics[width=\textwidth]{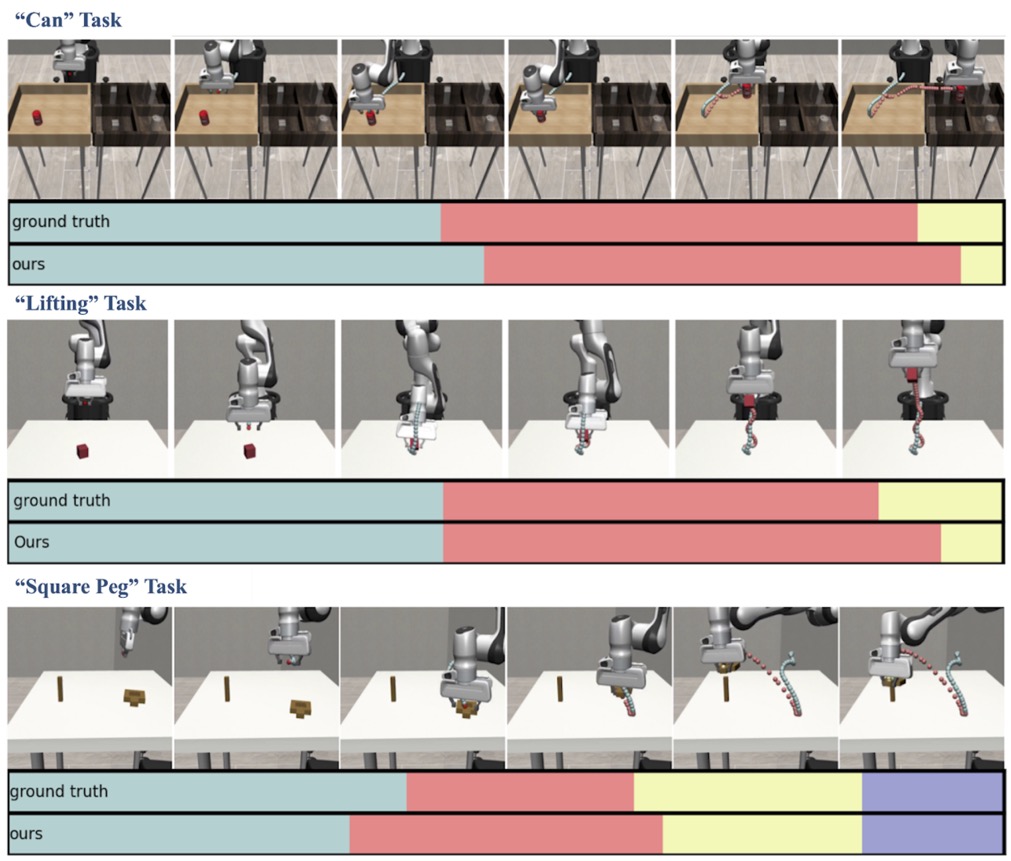}
  \caption{Comparison on the robosuite can task between our method's segmented modes and the ground truth modes. Generally, our system is able to accurately recover mode boundaries.}
  \label{fig:glide_robosuitecansegmentation}
\end{figure}

We use the following heuristic rules to define the ground truth mode families to evaluate the grounding learned by \model:
\begin{itemize}[leftmargin=2.5em,nosep]
\item \textit{Can (3 modes):} the ground truth modes are reaching for the can (until the end effector makes contact with the can), transporting the can to the bin, and finally hovering about the target bin.
\item \textit{Lift (3 modes):} the ground truth modes are reaching for the cube (until the end effector makes contact with the cube), lifting the cube off the table, and finally moving to a certain height above the table.
\item \textit{Square (4 modes):} the ground truth modes are reaching for the nut (until the end effector makes contact with the nut), transporting the nut to the peg, aligning the nut above the peg, and finally lowering the nut into the assembled position.
\end{itemize}

We assess the predicted and ground truth mode (based on the heuristics) for each sample in the robosuite demonstrations to compute accuracy (i.e., the percentage of samples where the predicted and ground-truth modes are the same).
\fig{fig:glide_robosuitecansegmentation} shows the visualization of the mode segmentation from \model, compared to the ground truth on the can placing task. Our model faithfully identifies all the modes and yields a high consistency with the modes defined by human-crafted rules.

\section{Prompt Engineering For RoboSuite Tasks}
\subsection{Prompt Example}

\lstdefinestyle{mystyle}{
    backgroundcolor=\color{CadetBlue!15!white},   
    commentstyle=\color{Red3},
    numberstyle=\tiny\color{gray},
    stringstyle=\color{Blue3},
    basicstyle=\small\ttfamily,
    breakatwhitespace=false,         
    breaklines=true,                 
    numbers=left,                    
    numbersep=5pt,                  
    showspaces=false,                
    showstringspaces=false,
    showtabs=false,                  
    tabsize=2
}%
\lstset{language=[5.3]Lua,style={mystyle}}%

\begin{lstlisting}[breaklines=true]
You are an expert in generating robot action plans and features.

Given a language description of a task, such as "clean the plate in a sink," you should first generate an abstract plan for the task. Put them in <plan></plan>. The plan is a list of steps. You should ignore object finding or localization actions. For example,
<plan>
steps = [{'id': 1, 'desc': 'Reach the plate'},
 {'id': 2, 'desc': 'Close gripper to grasp the plate'},
 {'id': 3, 'desc': 'Move to the sink'},
 {'id': 4, 'desc': 'Turn on the faucet'}]
</plan>

Next, you should generate a feasibility matrix between all the steps. Put them in <feasibility></feasibility>
<feasibility>
feasibility = {
  (1, 2): True,   # after reaching the plate, we can directly close the gripper
  (1, 3): False,  # after reaching the plate, we can't directly move to the sink
  (1, 4): False,  # after reaching the plate, we can't directly turn on the faucet
  (2, 3): True,   # after closing the gripper, we can directly move to the sink
  (2, 4): False,  # after closing the gripper, we can't directly turn on the faucet
  (3, 4): True,   # after moving to the sink, we can't directly turn on the faucet
}
</feasibility>

The user will also give you a list of available features, such as robot poses, object poses. An example is the following:
<available_features>
avaiable_features = ['plate_pos', 'plate_quat', 'plate_to_robot0_eef_pos', 'plate_to_robot0_eef_quat', 'sink_pos', 'sink_quat', 'sink_to_robot0_eef_pos', 'sink_to_robot0_eef_quat', 'faucet_pos', 'faucet_quat', 'faucet_to_robot0_eef_pos', 'faucet_to_robot0_eef_quat', 'robot0_eef_pos', 'robot0_eef_quat', 'gripper_state']
</available_features>
We will use the following convention: XXX_pos and XXX_quat denote the pose of the object. XXX_to_robot_eef denotes the relative pose between the object and the robot end-effector.

You should select a subset of features. Such as:
<selected_features>
features = [ 'plate_to_robot0_eef_pos', 'plate_to_robot0_eef_quat', 'sink_to_robot0_eef_pos', 'sink_to_robot0_eef_quat', 'faucet_to_robot0_eef_pos', 'faucet_to_robot0_eef_quat' ,
' 'gripper_state'']
</selected_features>

For each step, you also need to generate a "goal"  pose, which is either an absolute pose of the robot or a relative pose between the robot and another object. In this case,
<goal>
goals = [
  {'id': '1', 'feature': 'plate_to_robot0_eef_pos'},
  {'id': '2', 'feature': 'plate_to_robot0_eef_pos'},
  {'id': '3', 'feature': 'sink_to_robot0_eef_pos'},
  {'id': '4', 'feature': 'faucet_to_robot0_eef_pos'},
]
</goal>
\end{lstlisting}

\subsection{LLM Response for Can Task}

\begin{lstlisting}
Task: move the can from tray 1 to tray 2. Both trays are fixed. Available features: 
<available_features>
available_features= ['can_pos', 'can_quat', 'can_to_robot0_eef_pos', 'can_to_robot0_eef_quat', 'robot0_eef_pos', 'robot0_eef_quat',  'gripper_state'] 
</available_features>

<plan>
steps = [{'id': 1, 'desc': 'Reach the can in tray 1'},
 {'id': 2, 'desc': 'Close gripper to grasp the can'},
 {'id': 3, 'desc': 'Move to tray 2'},
 {'id': 4, 'desc': 'Open gripper to release the can'}]
</plan>

<feasibility>
feasibility = {
  (1, 2): True,   # after reaching the can in tray 1, we can directly close the gripper
  (1, 3): False,  # after reaching the can in tray 1, we can't directly move to tray 2
  (1, 4): False,  # after reaching the can in tray 1, we can't directly open the gripper
  (2, 3): True,   # after closing the gripper, we can directly move to tray 2
  (2, 4): True,  # after closing the gripper, we can directly open the gripper
  (3, 4): False,   # after moving to tray 2, we can't directly open the gripper
}
</feasibility>

<selected_features>
features = ['can_to_robot0_eef_pos', 'can_to_robot0_eef_quat', 'robot0_eef_pos', 'robot0_eef_quat', 'gripper_state']
</selected_features>

<goal>
goals = [
  {'id': '1', 'feature': 'can_to_robot0_eef_pos'},
  {'id': '2', 'feature': 'can_to_robot0_eef_pos'},
  {'id': '3', 'feature': 'robot0_eef_pos'},
  {'id': '4', 'feature': 'robot0_eef_pos'},
]
</goal>
\end{lstlisting}

\subsection{LLM Response for Lifting Task}

\begin{lstlisting}
Task: lift a block from the table. Available features:
<available_features> available_features= ['block_pos', 'block_quat', 'block_to_robot0_eef_pos', 'block_to_robot0_eef_quat', 'robot0_eef_pos', 'robot0_eef_quat',  'gripper_state']
</available_features>

<plan>
steps = [{'id': 1, 'desc': 'Move to the block'},
         {'id': 2, 'desc': 'Close gripper to grasp the block'},
         {'id': 3, 'desc': 'Lift the block'}]
</plan>

<feasibility>
feasibility = {
  (1, 2): True,  # after moving to the block, we can directly close the gripper
  (1, 3): False,  # after moving to the block, we can't directly lift the block
  (2, 3): True,  # after closing the gripper, we can directly lift the block
}
</feasibility>

<selected_features>
features = ['block_to_robot0_eef_pos', 'block_to_robot0_eef_quat', 'robot0_eef_pos', 'robot0_eef_quat', 'gripper_state']
</selected_features>

<goal>
goals = [
  {'id': '1', 'feature': 'block_to_robot0_eef_pos'},
  {'id': '2', 'feature': 'block_to_robot0_eef_pos'},
  {'id': '3', 'feature': 'robot0_eef_pos'},
]
</goal>
\end{lstlisting}

\subsection{LLM Response For Square Peg Task}
\begin{lstlisting}
Task: insert a nut with a handle to a peg. Available features:
<available_features> available_features= ['nut_pos', 'nut_quat', 'nut_to_robot0_eef_pos', 'nut_to_robot0_eef_quat', 'handle_pos', 'handle_quat', 'handle_to_robot0_eef_pos', 'handle_to_robot0_eef_quat', 'peg_pos', 'peg_quat', 'peg_to_robot0_eef_pos', 'peg_to_robot0_eef_quat', 'robot0_eef_pos', 'robot0_eef_quat',  'gripper_state']
</available_features>

<plan>
steps = [{'id': 1, 'desc': 'Reach the nut'},
 {'id': 2, 'desc': 'Close the gripper to grasp the nut'},
 {'id': 3, 'desc': 'Move to the peg'},
 {'id': 4, 'desc': 'Insert the nut into the peg'}]
</plan>

<feasibility>
feasibility = {
  (1, 2): True,
  (1, 3): False,
  (1, 4): False,
  (2, 3): True,
  (2, 4): False,
  (3, 4): True,
}
</feasibility>

<selected_features>
features = ['nut_to_robot0_eef_pos', 'nut_to_robot0_eef_quat', 'peg_to_robot0_eef_pos', 'peg_to_robot0_eef_quat', 'gripper_state']
</selected_features>

<goals>
goals = [
  {'id': '1', 'feature': 'nut_to_robot0_eef_pos'},
  {'id': '2', 'feature': 'nut_to_robot0_eef_pos'},
  {'id': '3', 'feature': 'peg_to_robot0_eef_pos'},
  {'id': '4', 'feature': 'peg_to_robot0_eef_pos'},
]
</goals>
\end{lstlisting}

\section{Additional Information for Marble Scooping Task}
\label{app:glide_scooping}

\begin{figure}[!htb]
  \centering\small
  \includegraphics[width=\textwidth]{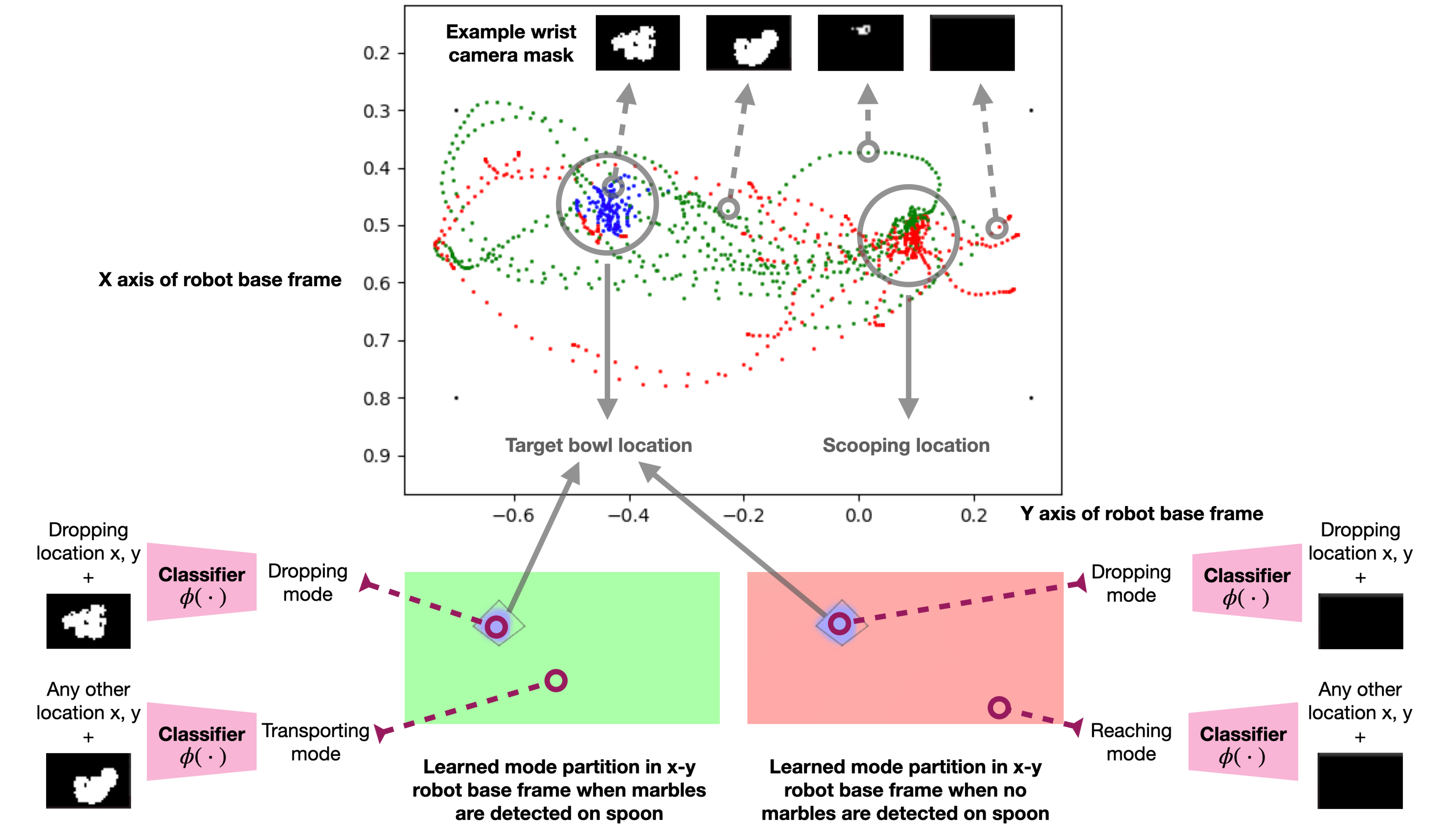}
  \caption{Visualizing the state representation of the scooping task demonstrations and the learned mode partitions.}
  \label{fig:glide_scoop_mode_vis}
\end{figure}

We prompt LLM to generate a subset of features relevant to predicting task success: X, and Y locations of the robot end-effector in the robot base frame as well as the wrist camera mask. Due to a lack of contact sensors, we omit the scooping mode, and prompt the LLM to generate a plan: Reaching $\rightarrow$ Transporting $\rightarrow$ Dropping (assuming scooping is always successful when transitioning from the reaching mode to the transporting mode). The corresponding feasibility matrix is $F^3$. In Fig. \ref{fig:glide_scoop_mode_vis} (top), we plot demonstrations in X and Y and use the color of the scattered plot to indicate ground truth modes (reaching is red,  transporting is green, and dropping is blue). Examples of logged spoon masks along these trajectories are shown at the top. At the bottom, we visualize the learned classifier, which has correctly learned three modes (indicated by three distinct colors) by partitioning the space according to X and Y locations and the masks. Note that the location of the learned blue mode matches the dropping bowl location. 

\defbibheading{bibintoc}{\chapter*{#1}\addcontentsline{toc}{backmatter}{\refname}} 
\printbibliography[title={\refname},heading=bibintoc]

\end{document}